\documentclass[10pt,twocolumn,letterpaper]{article}

\usepackage{cvpr}              

\usepackage[OT1]{fontenc}

\usepackage{multirow}

\usepackage{fvextra}
\usepackage{cprotect}

\usepackage[export]{adjustbox}

\usepackage{pgfplots}
\usetikzlibrary{calc, positioning, shapes.geometric}
\pgfplotsset{compat=newest}

\tikzset{
  between/.style args={#1 and #2}{
    at = ($(#1)!0.5!(#2)$)
  }
}

\definecolor{color_hl}{RGB}{221,170,51}

\definecolor{color_l}{RGB}{204,187,68}
\definecolor{color_e}{RGB}{68,119,170}
\definecolor{color_d}{RGB}{34,136,51}
\definecolor{color_ca}{RGB}{238,102,119}
\definecolor{color_sa}{RGB}{102,204,238}
\definecolor{color_gru}{RGB}{187,187,187}
\definecolor{color_sk}{RGB}{140,140,140}
\definecolor{color_skd}{RGB}{170,51,119}

\newcommand{\smyellowsquare}{\includegraphics{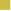}}

\newcommand{\smbluesquare}{\includegraphics{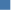}}
\newcommand{\smgreensquare}{\includegraphics{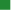}}

\newcommand{\rpm}{\raisebox{.2ex}{\(\scriptstyle\pm\)}}

\definecolor{cvprblue}{rgb}{0.21,0.49,0.74}
\usepackage[pagebackref,breaklinks,colorlinks,allcolors=cvprblue]{hyperref}

\def\paperID{9} 
\def\confName{CVPR}
\def\confYear{2025}

\title{DELTA: Dense Depth from Events and LiDAR using Transformer's Attention}

\author{Vincent Brebion\textsuperscript{1,2,3} \hspace{5mm} Julien Moreau\textsuperscript{2} \hspace{5mm} Franck Davoine\textsuperscript{3}\\
\textsuperscript{1}Centre for Environmental and Climate Science, Lund University, Sweden -- {\tt\small vincent.brebion@cec.lu.se}\\
\textsuperscript{2}Universit\'e de technologie de Compi\`egne, CNRS, Heudiasyc, France -- {\tt\small julien.moreau@hds.utc.fr}\\
\textsuperscript{3}CNRS, INSA Lyon, UCBL, LIRIS, UMR5205, France -- {\tt\small franck.davoine@cnrs.fr}
}

\begin{document}
\maketitle
\begin{abstract}
  Event cameras and LiDARs provide complementary yet distinct data: respectively, asynchronous detections of changes in lighting versus sparse but accurate depth information at a fixed rate. To this day, few works have explored the combination of these two modalities. In this article, we propose a novel neural-network-based method for fusing event and LiDAR data in order to estimate dense depth maps. Our architecture, DELTA, exploits the concepts of self- and cross-attention to model the spatial and temporal relations within and between the event and LiDAR data. Following a thorough evaluation, we demonstrate that DELTA sets a new state of the art in the event-based depth estimation problem, and that it is able to reduce the errors up to four times for close ranges compared to the previous SOTA.
\end{abstract}

\section{Introduction}\label{sec:intro}
Due to their unique nature, event cameras offer a large paradigm change in the field of computer vision. By capturing changes of lighting for each pixel independently and asynchronously, and by transmitting this information as spikes of data (\textit{events}), these cameras provide ultra-low latency perception (in the order of the microsecond). They allow for novel applications and present an invaluable potential, especially under challenging lighting or motion conditions where traditional frame-based cameras would fail.

\begin{figure}
  \centering
  \setlength\tabcolsep{0.9pt}
  \begin{tabular}{@{}ccccc@{}}
    \includegraphics[width=0.295\linewidth,cfbox=color_l 1.5pt 0pt]{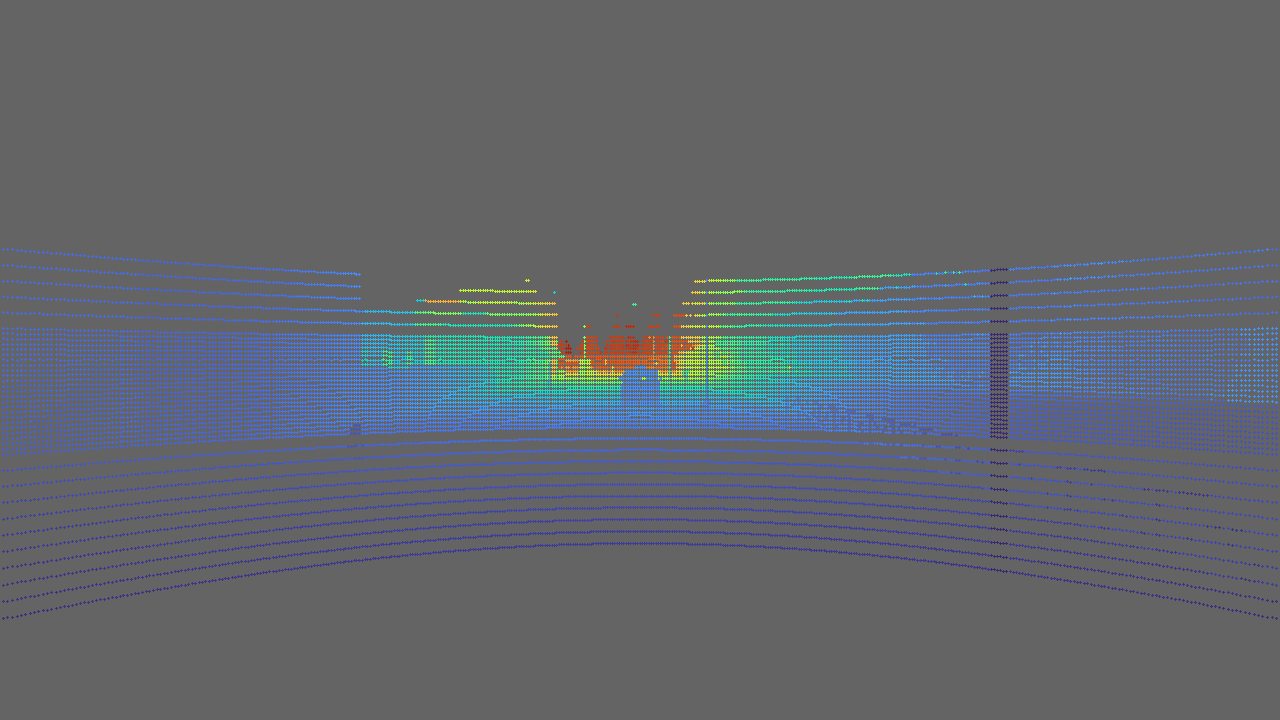} &
    \multirow{-2}[7]{*}{\textbf{+}} &
    \includegraphics[width=0.295\linewidth,cfbox=color_e 1.5pt 0pt]{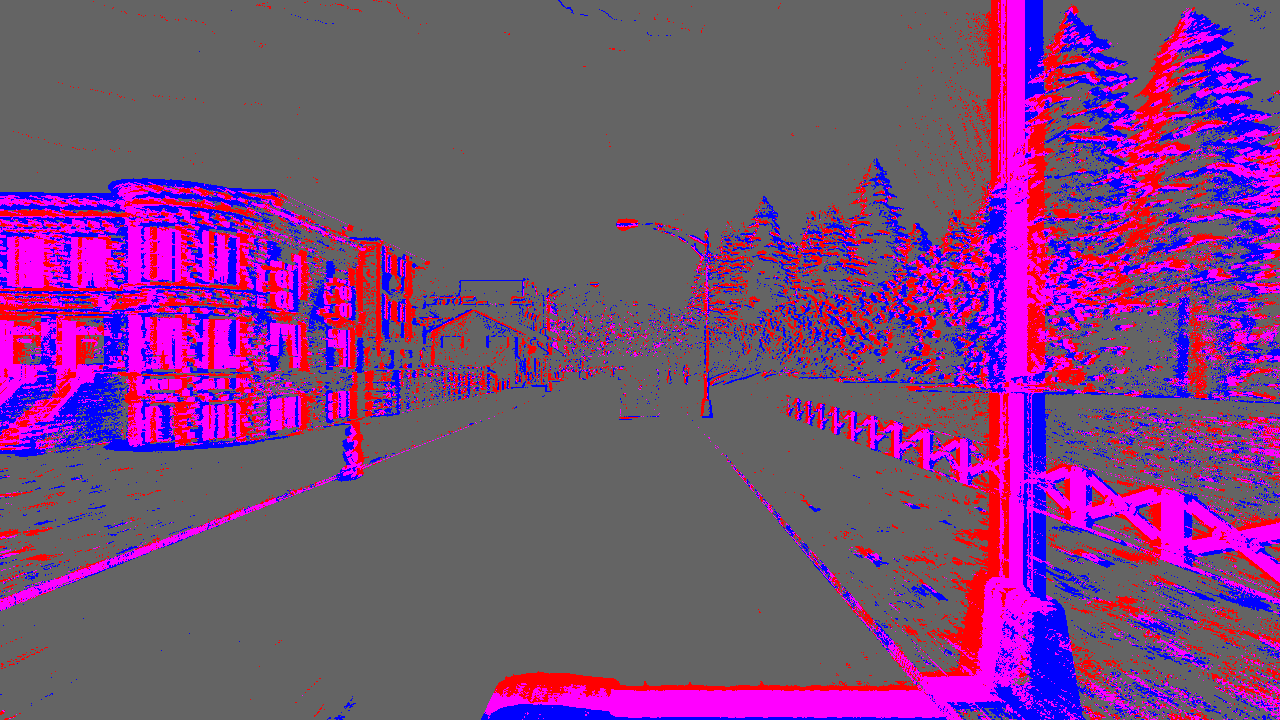} &
    \multirow{-2}[7]{*}{\textbf{=}} &
    \includegraphics[width=0.295\linewidth,cfbox=color_d 1.5pt 0pt]{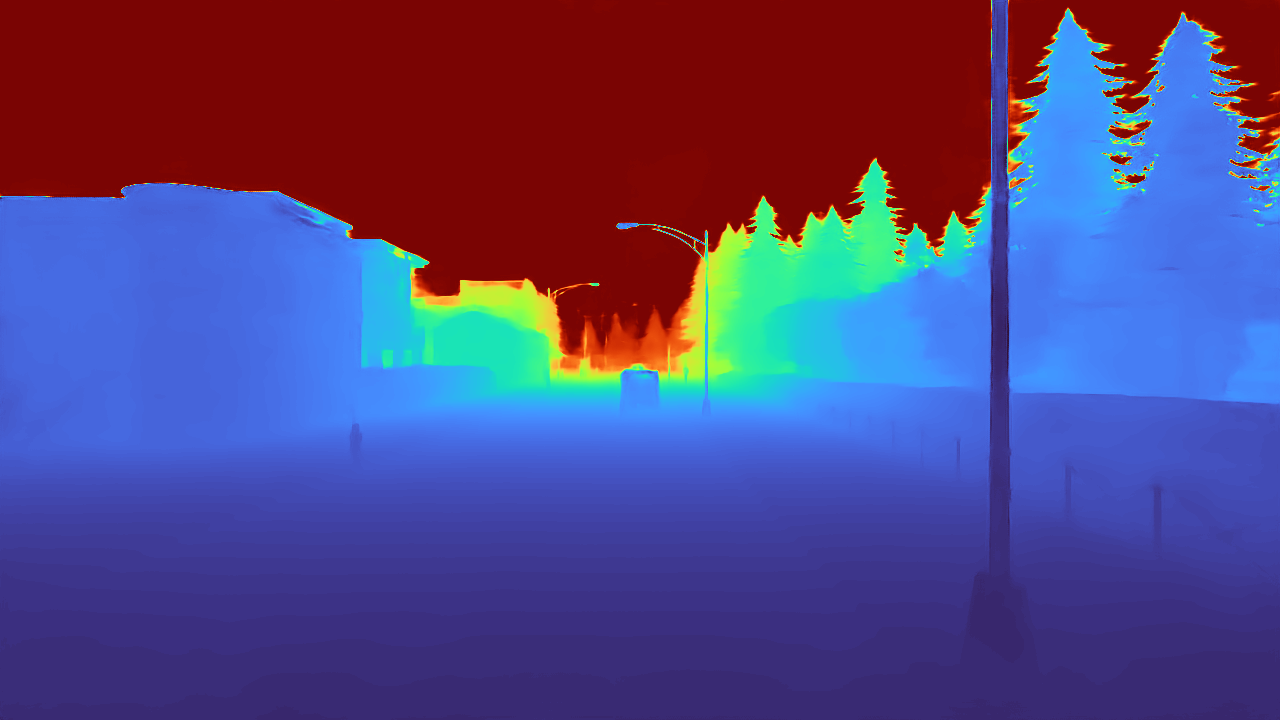} \\
    \begin{tikzpicture}
      \foreach \t in {0.05,0.25,...,1.45}
        \draw[line width=0.3mm,color_l] (0+\t, 0) -- (0+\t, 0.3);
      \draw[->,line width=0.4mm] (0, 0) -- (1.6, 0);
    \end{tikzpicture} &
    &
    \begin{tikzpicture}
      \foreach \t in {0.05,0.15,...,1.45}
        \draw[line width=0.3mm,color_e] (0+\t, 0) -- (0+\t, 0.3);
      \draw[->,line width=0.4mm] (0, 0) -- (1.6, 0);
    \end{tikzpicture} &
    &
    \begin{tikzpicture}
      \foreach \t in {0.05,0.15,...,1.45}
        \draw[line width=0.3mm,color_d] (0+\t, 0) -- (0+\t, 0.3);
      \draw[->,line width=0.4mm] (0, 0) -- (1.6, 0);
    \end{tikzpicture}
  \end{tabular}
  \caption{Overall principle of DELTA. Sparse and low-rate projected LiDAR data (\smyellowsquare{}) is densified spatially and temporally using higher-rate temporal windows of event data (\smbluesquare{}), resulting in dense high-rate depth maps (\smgreensquare{}). Displayed here is an example of the high-quality depth maps produced by DELTA.}\label{fig:teaser}
\end{figure}
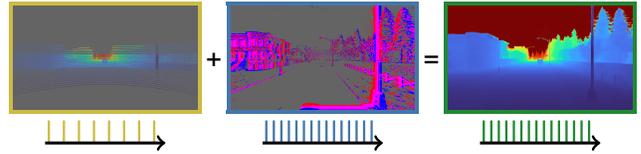

Their combination with other modalities like RGB or grayscale cameras is a topic that has been deeply explored over the past few years. Yet, the use of event cameras with LiDARs remains relatively untouched, despite the valuable information a depth sensor can bring for scene analysis. As displayed in \cref{fig:teaser}, we propose here to exploit both the precise edge detection and the high temporal resolution of the event data to densify both spatially and temporally the LiDAR data, resulting in dense high-rate depth maps.

This article offers four key contributions. We propose
\begin{enumerate*}[label=\textbf{(\textcolor{cvprblue}{\arabic*})}]
  \item a novel attention-based network, which we call DELTA (for ``Dense depth from Events and LiDAR using Transformer's Attention''). DELTA combines information from low-rate projected LiDAR point clouds with higher-rate small temporal windows of events, in order to derive accurate dense depth maps. Thanks to its attention- and recurrence-based design, unlike traditional CNNs, our network is able to extract the most relevant spatial and temporal features within and between the event and LiDAR data. The introduction of
  \item a propagation memory for a fusion at the highest input rate and of
  \item a central memory acting as a main recurrence both allow us to outperform the state of the art, and are critical contributions as shown through ablation studies. An extensive evaluation of DELTA is conducted on multiple automotive datasets, where LiDAR and event sensors are most commonly used together. We demonstrate that in most cases our proposed network is able to offer
  \item a clear improvement over the state of the art, especially for close ranges (often the most critical in dynamic scenes, \eg, close pedestrians in automotive scenarios), where the mean error in depth estimation is reduced up to four times.
\end{enumerate*}

Additional analysis and results are available as supplementary material. The source code and trained models are available at {\small\url{https://vbrebion.github.io/DELTA}}.

\section{Related Work}

\subsection{Transformers for Event-based Data}
The Transformer~\cite{Vaswani2017AttentionIA} has become the state-of-the-art architecture in numerous domains. Its attention mechanism models explicitly the relations between relevant elements in a sequence, making it able to understand underlying structures. For computer vision, the arrival of the Vision Transformer (ViT)~\cite{Dosovitskiy2020AnII} has been a notable landmark, outperforming more traditional convolution-based networks. As such, researchers have started investigating how the Transformer architecture could be adapted to event-based cameras. Two philosophies have emerged over the years.
\begin{enumerate*}[label=\textbf{(\textcolor{cvprblue}{\arabic*})}]
  \item Some authors use directly the raw stream of events (without any pre-processing) as the input sequence to their network, and use the Transformer architecture to process it. This approach is particularly complex, as each event contains little information, making the modeling of their relations difficult. To contain enough context, sequences of events should also be of consequent size. As of today, this method has only been applied to the tasks of object detection and classification~\cite{Li2022EventT,Kamal2023AssociativeMA,Peng2023GetGE}, where the event data can be highly compressed.
  \item To circumvent these issues, most authors instead accumulate events in a frame-like representation, and process it using the standard patch-based format proposed with ViT. Investigated tasks include video reconstruction~\cite{Weng2021EventbasedVR}, object detection~\cite{Gehrig2022RecurrentVT}, classification~\cite{Wang2022ExploitingSS,Sabater2022EventTA,Sabater2022EventTA+}, depth estimation~\cite{Sabater2022EventTA+}, and optical flow~\cite{Tian2022EventTF}.
\end{enumerate*}
In this work, we follow principle \textbf{(\textcolor{cvprblue}{2})}, as it allows the network to extract meaningful spatial and temporal features from the event and LiDAR data for the depth densification task.

\subsection{Fusion of Event and LiDAR Data}
To this day, most research works using both the LiDAR and event-based modalities address the problem of extrinsic calibration~\cite{Cocheteux2024MULiEvMU,Jiao2023LCECalibAL,Song2018CalibrationOE,Ta2022L2ELT}, or use them as part of the construction of a dataset~\cite{Brebion2023LearningTE,Chaney2023M3EDMM,Gehrig2021DSECAS,Zhu2018TheMS}. Recently, authors have started investigating the issues of enhancing point clouds with event-based data~\cite{Li2021Enhancing3L}, of estimating dense depth maps from event and LiDAR data~\cite{Brebion2023LearningTE,Cui2022DenseDE}, or of tracking humans in adversarial lighting conditions~\cite{Saucedo2023EventCA}.

\subsection{Event-Based Depth Estimation}
The idea of estimating sparse or dense depth maps from events has been actively explored over the past decade. Three main approaches can be distinguished.
\begin{enumerate*}[label=\textbf{(\textcolor{cvprblue}{\arabic*})}]
  \item Some authors estimate depths in a monocular fashion, using only events from a single event camera~\cite{Chiavazza2023LowlatencyMD,HidalgoCarrio2020LearningMD,Kim2016RealTime3R,Liu2022EventbasedMD,Nunes2023TimetocontactMB,Ranon2021StereoSpikeDL,Zhu2019UnsupervisedEL}, or using events and frames~\cite{Gehrig2021CombiningEA,Sabater2022EventTA+,Hamaguchi2023HierarchicalNM,Liu2024PCDepthPC} from a DAVIS camera~\cite{Brandli2014A2}. These approaches are particularly challenging, as they lack any three-dimensional information.
  \item Some authors have tried to estimate depth in a stereo fashion, by using a pair of event cameras~\cite{Cho2023LearningAD,Ghosh2022MultiEventCameraDE,Nam2022StereoDF,Ranon2021StereoSpikeDL,Schraml2010DynamicSV,Schraml2016AnES}, with~\cite{Cho2023LearningAD} and~\cite{Ranon2021StereoSpikeDL} achieving notably good results.
  \item Finally, some authors prefer to use directly a depth sensor, and use the stream of event as a mean to densify and/or to temporally upsample the depth data. This depth sensor can either be an RGB-D camera~\cite{Weikersdorfer2014Eventbased3S} or a LiDAR~\cite{Brebion2023LearningTE,Cui2022DenseDE,Li2021Enhancing3L}, with~\cite{Cui2022DenseDE,Li2021Enhancing3L} being based respectively on 3D- and 2D-geometry methods, and with~\cite{Brebion2023LearningTE} (our previous work) being based on a CNN.
\end{enumerate*}

\subsection{Positioning of our Work Compared to the State of the Art}
Like~\cite{Brebion2023LearningTE,Cui2022DenseDE,Li2021Enhancing3L}, we exploit LiDAR data to solve the problem of event-based depth estimation. Like these works, we argue (and we show in \cref{sec:eval}) that having reference depth points (even if sparse both spatially and temporally) is of great help for solving the event-based depth estimation problem. But unlike~\cite{Cui2022DenseDE,Li2021Enhancing3L}, we do not use a geometry-based approach, due to limitations highlighted by these two works: depth estimation is only possible for areas with close LiDAR points, output depth maps frequency is restricted to the one of the LiDAR data, and these methods are sensitive to noise. Our approach is more similar to our previous work~\cite{Brebion2023LearningTE}, as we use a learning-based approach to solve these limitations. But while we used a convolution-based network in~\cite{Brebion2023LearningTE}, we propose here instead a novel attention-based network, which can better capture the spatial and temporal relations within and between event and LiDAR data, without being restricted by the limited reception field of convolutions. As shown in \cref{sec:eval}, this novel architecture greatly outperforms the state of the art, especially for short ranges.

\section{Method}

\begin{figure*}
  \centering
  \resizebox{0.91\linewidth}{!}{
    \begin{tikzpicture}[every node/.style={inner sep=0.001,outer sep=0.001,font=\normalsize}]
      \fontfamily{cmss}\selectfont


      \node[] (L) at (0,0) {\includegraphics[height=2cm,cfbox=color_l 0.05cm 0pt]{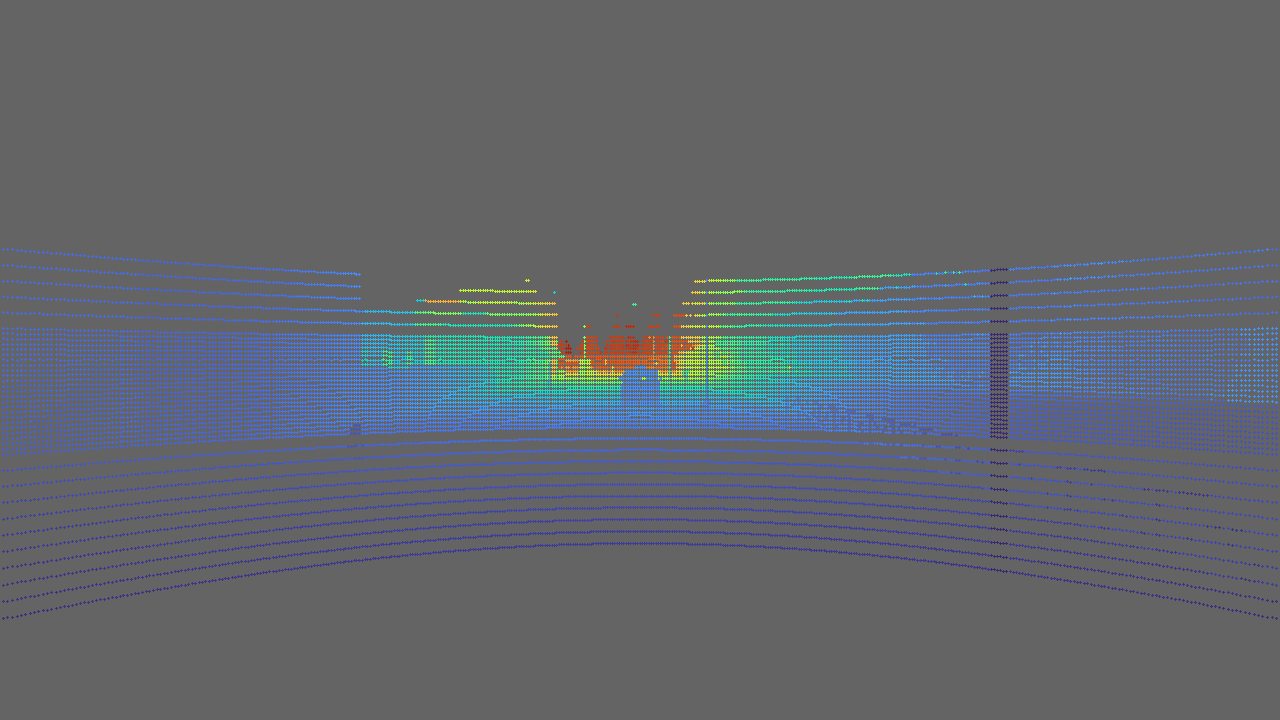}};
      \node[above=0.1cm of L] (Ll) {\normalsize Input Projected LiDAR};

      \foreach \t in {0.05,0.25,...,0.65}
        \draw[line width=0.4mm,color_l] ($(L.west)+(-1.3+\t, 0)$) -- ($(L.west)+(-1.3+\t, 0.375)$);
      \draw[->,line width=0.6mm] ($(L.west)+(-1.3, 0)$) -- ($(L.west)+(-0.5, 0)$);

      \node[draw, trapezium, minimum width=1.25cm, on grid, right=3.75cm of L, opacity=0.0] (tmplh) {};
      \node[draw, trapezium, fill=color_l, minimum width=1.25cm, rotate around={-90:(tmplh.center)}] (LH) at (tmplh) {};

      \node[draw, circle, minimum width=0.625cm, right=0.75cm of tmplh] (LP) {\Large +};

      \node[draw, fill=color_ca, rounded corners, minimum width=1cm, minimum height=1cm, right=1.5cm of LP] (LCA) {CA\textsubscript{P2}};

      \foreach \t in {0.15,0.35,...,0.65}
        \draw[line width=0.4mm,color_e] ($(LCA.east)+(0.2+\t, 0.2)$) -- ($(LCA.east)+(0.2+\t, 0.575)$);
      \foreach \t in {0.05,0.25,...,0.65}
        \draw[line width=0.4mm,color_l] ($(LCA.east)+(0.2+\t, 0.2)$) -- ($(LCA.east)+(0.2+\t, 0.575)$);
      \draw[->,line width=0.6mm] ($(LCA.east)+(0.2, 0.2)$) -- ($(LCA.east)+(1.0, 0.2)$);

      \node[draw, fill=color_sa, rounded corners, minimum width=1cm, minimum height=1cm, right=2.5cm of LCA] (LSA1) {SA};

      \node[draw, fill=color_sa, rounded corners, minimum width=1cm, minimum height=1cm, right=1.5cm of LSA1] (LSA2) {SA};

      \node[below=2.9cm of L] (E) {\includegraphics[height=2cm,cfbox=color_e 0.05cm 0pt]{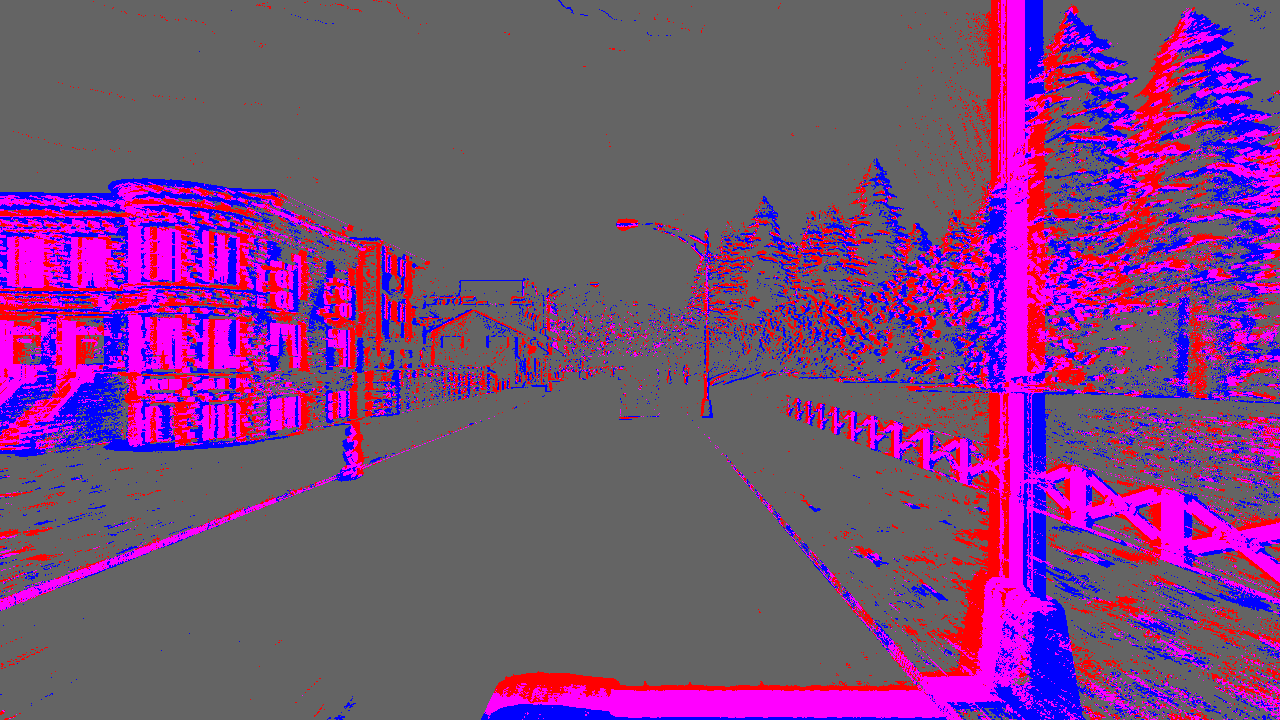}};
      \node[above=0.1cm of E] (El) {\normalsize Input Event Volume};

      \foreach \t in {0.05,0.15,...,0.65}
        \draw[line width=0.4mm,color_e] ($(E.west)+(-1.3+\t, 0)$) -- ($(E.west)+(-1.3+\t, 0.375)$);
      \draw[->,line width=0.6mm] ($(E.west)+(-1.3, 0)$) -- ($(E.west)+(-0.5, 0)$);

      \node[draw, trapezium, minimum width=1.25cm, on grid, right=3.75cm of E, opacity=0.0] (tmpeh) {};
      \node[draw, trapezium, fill=color_e, minimum width=1.25cm, rotate around={-90:(tmpeh.center)}] (EH) at (tmpeh) {};

      \node[draw, circle, minimum width=0.625cm, right=0.75cm of tmpeh] (EP) {\Large +};

      \node[draw, fill=color_ca, rounded corners, minimum width=1cm, minimum height=1cm, below=2.6667cm of LCA] (PCA) {CA\textsubscript{P1}};

      \node[draw, minimum width=1cm, minimum height=1cm, above=0.8cm of PCA, align=center] (P) {Prop.\\mem.};

      \node[draw, fill=color_sa, rounded corners, minimum width=1cm, minimum height=1cm, below right=0.3333cm and 2.5cm of PCA] (ESA1) {SA};

      \node[draw, fill=color_sa, rounded corners, minimum width=1cm, minimum height=1cm, right=1.5cm of ESA1] (ESA2) {SA};

      \node[draw, circle, minimum width=1cm, minimum height=1cm, between=LP and EP] (PE) {};
      \draw[thick] ($(PE)+(-0.5cm,0)$) sin ($(PE)+(-0.25cm,0.3cm)$) cos (PE) sin ($(PE)+(0.25cm,-0.3cm)$) cos ($(PE)+(0.5cm,0)$);

      \node[draw, fill=color_ca, rounded corners, minimum width=1cm, minimum height=1cm, below right=1.5cm and 0.75cm of LSA2] (MCA) {CA\textsubscript{F}};

      \node[draw, fill=color_gru, rounded corners, minimum width=1cm, minimum height=1cm, right=0.75cm of MCA] (MGRU) {GRU};

      \node[draw, minimum width=1cm, minimum height=1cm, right=0.75cm of MGRU, align=center] (M) {Cent.\\mem.};

      \node[draw, fill=color_sa, rounded corners, minimum width=1cm, minimum height=1cm, on grid, below=3.5cm of ESA2] (DSA2) {SA};

      \node[draw, fill=color_sa, rounded corners, minimum width=1cm, minimum height=1cm, left=1.5cm of DSA2] (DSA1) {SA};

      \node[draw, trapezium, minimum width=1.25cm, on grid, below=3.5cm of tmpeh, opacity=0.0] (tmpdh) {};
      \node[draw, trapezium, fill=color_d, minimum width=1.25cm, rotate around={-90:(tmpdh.center)}] (DH) at (tmpdh) {};

      \node[on grid, left=3.75cm of tmpdh] (D) {\includegraphics[height=2cm,cfbox=color_d 0.05cm 0pt]{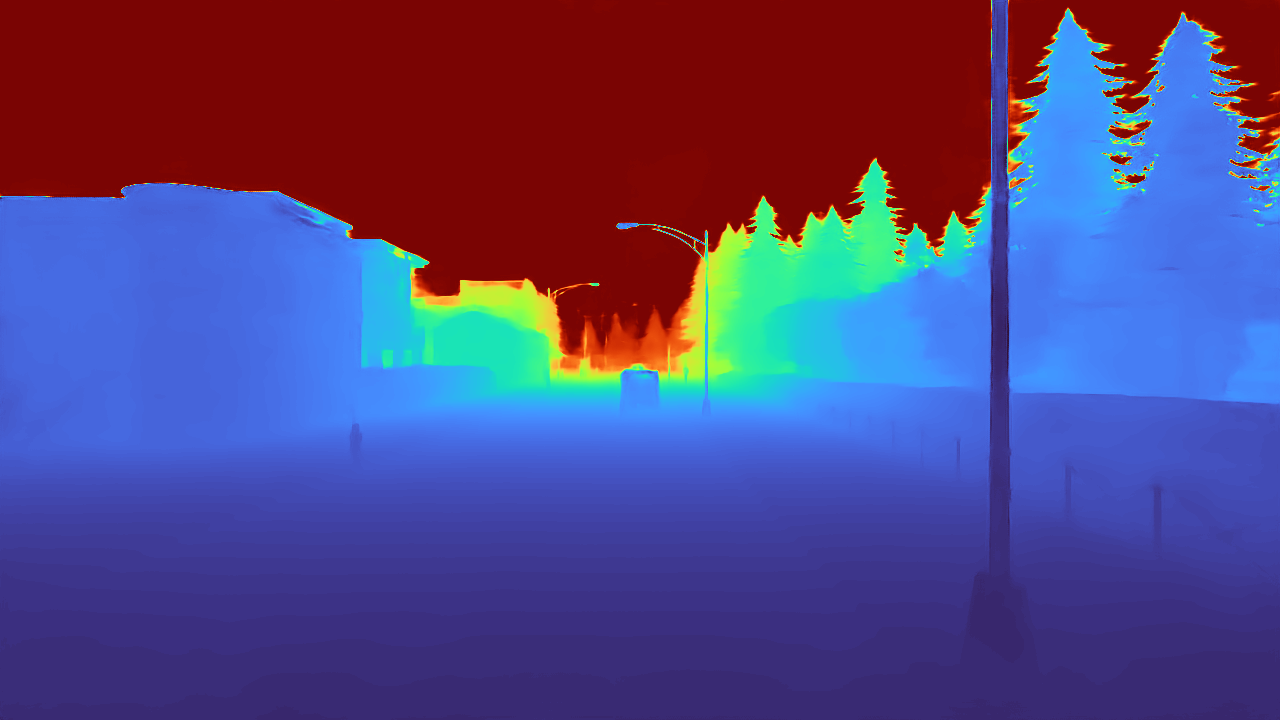}};
      \node[above=0.1cm of D] (Dl) {\normalsize Output Depth Map};

      \foreach \t in {0.05,0.15,...,0.65}
        \draw[line width=0.4mm,color_d] ($(D.west)+(-1.3+\t, 0)$) -- ($(D.west)+(-1.3+\t, 0.375)$);
      \draw[->,line width=0.6mm] ($(D.west)+(-1.3, 0)$) -- ($(D.west)+(-0.5, 0)$);

      \node[draw, circle, color_sk, minimum width=0.625cm, below left=1.6875cm and 0.75cm of LSA1] (SP1a) {\Large +};
      \node[draw, circle, color_sk, minimum width=0.625cm, below left=1.6875cm and 0.75cm of LSA2] (SP2a) {\Large +};
      \node[draw, circle, color_sk, minimum width=0.625cm, left=0.25cm of DSA1] (SP1b) {\Large +};
      \node[draw, circle, color_sk, minimum width=0.625cm, left=0.25cm of DSA2] (SP2b) {\Large +};

      \draw[dashed, color_sk, ->, >=latex, line width=0.5mm] (ESA1.west -| SP1a) -- (SP1a);
      \draw[dashed, color_sk, ->, >=latex, line width=0.5mm] (ESA2.west -| SP2a) -- (SP2a);
      \draw[dashed, color_sk, ->, >=latex, line width=0.5mm] (LSA1.west -| SP1a) -- (SP1a);
      \draw[dashed, color_sk, ->, >=latex, line width=0.5mm] (LSA1.west -| SP2a) -- (SP2a);
      \draw[dashed, color_sk, ->, >=latex, line width=0.5mm] (SP1a) -| (SP1b);
      \draw[dashed, color_sk, ->, >=latex, line width=0.5mm] (SP2a) -| (SP2b);
      \draw[dashdotted, color_skd, ->, >=latex, line width=0.5mm] (EH) -- (DH);

      \draw[->,>=latex, line width=0.5mm, color_l] (L) -- node [above=0.1cm, midway] (al) {(H, W, 1)} (LH);
      \draw[->,>=latex, line width=0.5mm, color_l] (LH) -- (LP);
      \draw[->,>=latex, line width=0.5mm, color_l] (LP) -- node [above=0.1cm, pos=0.85] (aplq) {\color{color_ca} \textbf{Q}} (LCA);
      \draw[->,>=latex, line width=0.5mm, color_e] (LCA) -- (LSA1);
      \draw[->,>=latex, line width=0.5mm, dashed, color_l] (LCA) -- (LSA1);
      \draw[->,>=latex, line width=0.5mm, color_e] (LSA1) -- (LSA2);
      \draw[->,>=latex, line width=0.5mm, dashed, color_l] (LSA1) -- (LSA2);
      \draw[->,>=latex, line width=0.5mm, color_e] (LSA2) -| node [left=0.1cm, pos=0.95] (amkv) {\color{color_ca} \textbf{K/V}} (MCA);
      \draw[->,>=latex, line width=0.5mm, dashed, color_l] (LSA2) -| (MCA);

      \draw[->,>=latex, line width=0.5mm, color_e] (E) -- node [above=0.1cm, midway] (ae) {(H, W, 4)} (EH);
      \draw[->,>=latex, line width=0.5mm, color_e] (EH) -- (EP);
      \draw[->,>=latex, line width=0.5mm, color_e] (EP -| PCA) -- node [left=0.1cm, pos=0.7] (apkv) {\color{color_ca} \textbf{K/V}} (PCA);
      \draw[->,>=latex, line width=0.5mm, color_e] (EP) -- (ESA1);
      \draw[->,>=latex, line width=0.5mm, color_e] (ESA1) -- (ESA2);
      \draw[->,>=latex, line width=0.5mm, color_e] (ESA2) -| node [left=0.1cm, pos=0.935] (amq) {\color{color_ca} \textbf{Q}} (MCA);

      \draw[->,>=latex, line width=0.5mm] (PE) -- (LP);
      \draw[->,>=latex, line width=0.5mm] (PE) -- (EP);

      \draw[->,>=latex, line width=0.5mm, color_e] ([xshift=-0.2cm] P.south) -- node [left=0.1cm, pos=0.75] (apq) {\color{color_ca} \textbf{Q}} node [left=0.1cm, pos=0.3] (apq2) {(128, D)} ([xshift=-0.2cm] PCA.north);
      \draw[->,>=latex, line width=0.5mm, color_e] ([xshift=0.2cm] PCA.north) -- node [right=0.1cm] (apo) {(128, D)} ([xshift=0.2cm] P.south);
      \draw[->,>=latex, line width=0.5mm, color_e] (P) -- node [left=0.1cm, pos=0.7] (aplkv) {\color{color_ca} \textbf{K/V}} node [right=0.1cm] (aplkv2) {(128, D)} (LCA);

      \draw[->,>=latex, line width=0.5mm] (MCA) -- (MGRU);
      \draw[->,>=latex, line width=0.5mm] ([yshift=0.2cm] M.west) -- ([yshift=0.2cm] MGRU.east);
      \draw[->,>=latex, line width=0.5mm] ([yshift=-0.2cm] MGRU.east) -- ([yshift=-0.2cm] M.west);

      \draw[->,>=latex, line width=0.5mm, color_d] (M) |- (DSA2);
      \draw[->,>=latex, line width=0.5mm, color_d] (DSA2) -- (SP2b);
      \draw[->,>=latex, line width=0.5mm, color_d] (SP2b) -- (DSA1);
      \draw[->,>=latex, line width=0.5mm, color_d] (DSA1) -- (SP1b);
      \draw[->,>=latex, line width=0.5mm, color_d] (SP1b) -- (DH);
      \draw[->,>=latex, line width=0.5mm, color_d] (DH) -- node [above=0.1cm, midway] (ad) {(H, W, 1)} (D);

      \draw[line width=0.6mm] ($(D.south west)+(-1.5cm,-1cm)$) -- ($(D.south west)+(21.75cm,-1cm)$);

      \node[align=center, below right=1.75cm and 9cm of D.south west] (Title) {\Large\textbf{Legend}};

      \node[draw, trapezium, minimum width=1cm, below left=1.25cm and 10.5cm of Title, opacity=0.0] (tmplh) {};
      \node[draw, trapezium, fill=color_l, minimum width=1cm, rotate around={-90:(tmplh.center)}] (LH) at (tmplh) {};

      \node[draw, trapezium, minimum width=1cm, right=0cm of tmplh, opacity=0.0] (tmpeh) {};
      \node[draw, trapezium, fill=color_e, minimum width=1cm, rotate around={-90:(tmpeh.center)}] (EH) at (tmpeh) {};

      \node[align=center, right=0.3cm of tmpeh] (HEl) {\normalsize Convolutional\\\normalsize encoding heads};

      \node[draw, trapezium, minimum width=1cm, on grid, below left=1.25cm and -0.075cm of tmpeh.south west, opacity=0.0] (tmpdh) {};
      \node[draw, trapezium, fill=color_d, minimum width=1cm, rotate around={-90:(tmpdh.center)}] (DH) at (tmpdh) {};

      \node[align=center, right=0.775cm of tmpdh] (HDl) {\normalsize Convolutional\\\normalsize decoding head};

      \node[draw, fill=color_ca, rounded corners, minimum width=1cm, minimum height=1cm, right=1.5cm of HEl] (CA) {CA};

      \node[align=center, right=0.5cm of CA] (CAl) {\normalsize Cross-attention\\\normalsize module};

      \node[draw, fill=color_sa, rounded corners, minimum width=1cm, minimum height=1cm, on grid, below=0.75cm of CA.south] (SA) {SA};

      \node[align=center, right=0.7cm of SA] (SAl) {\normalsize Self-attention\\\normalsize module};

      \node[draw, circle, minimum width=1cm, minimum height=1cm, right=1.5cm of CAl] (PE) {};
      \draw[thick] ($(PE)+(-0.5cm,0)$) sin ($(PE)+(-0.25cm,0.3cm)$) cos (PE) sin ($(PE)+(0.25cm,-0.3cm)$) cos ($(PE)+(0.5cm,0)$);

      \node[align=center, right=0.8cm of PE] (PEl) {\normalsize 2D positional\\\normalsize embedding};

      \node[draw, fill=color_gru, rounded corners, minimum width=1cm, minimum height=1cm, on grid, below=0.75cm of PE.south] (GRU) {GRU};

      \node[align=center, right=0.5cm of GRU] (GRUl) {\normalsize Gated Recurrent\\\normalsize Unit module};

      \node[draw, align=center, minimum width=1cm, minimum height=1cm, right=1.5cm of PEl] (P) {Prop.\\mem.};

      \node[align=center, right=0.75cm of P] (Pl) {\normalsize Propagation\\\normalsize memory};

      \node[draw, align=center, minimum width=1cm, minimum height=1cm, on grid, below=0.75cm of P.south] (M) {Cent.\\mem.};

      \node[align=center, right=1.1cm of M] (Ml) {\normalsize Central\\\normalsize memory};

      \node[minimum width=1cm, minimum height=1cm, right=1.5cm of Pl] (S) {};
      \draw[dashed, color_sk, ->, >=latex, line width=0.5mm] (S.west) -- (S.east);

      \node[align=center, right=0.8cm of S] (Sl) {\normalsize Skip\\\normalsize connections};

      \node[minimum width=1cm, minimum height=1cm, on grid, below=0.75cm of S.south] (G) {};
      \draw[dashdotted, color_skd, ->, >=latex, line width=0.5mm] (G.west) -- (G.east);

      \node[align=center, right=0.95cm of G] (Gl) {\normalsize Decoding\\\normalsize guide};
    \end{tikzpicture}
  }

  \caption{The complete architecture of our DELTA network. Unless noted, data is of shape \((N, D)\), where \(N\) is the number of patches, and \(D\) their dimensionality (please refer to \cref{sec:method:encod_heads,sec:eval:impl_detail:data_size} for more details).}\label{fig:network}
\end{figure*}

\subsection{A Single Depth Map}
In~\cite{Brebion2023LearningTE}, we argued (and still argue) that, as an event describes a change between two illumination values, it may also be linked to a change between two depth values, hence the need of estimating a depth \textit{before} the event happened (\(\mathrm{d}_\text{bf}\)) and a depth \textit{after} the event happened (\(\mathrm{d}_\text{af}\)). However, in~\cite{Brebion2023LearningTE}, the computation of the depths is not done at the event level, but at the temporal window level. Therefore, we proposed to estimate two depth maps for each temporal window of events (\(\mathrm{D}_\text{bf}\) and \(\mathrm{D}_\text{af}\)), and to then use the events as a mask to assign to each of them their two depths \(\mathrm{d}_\text{bf}\) and \(\mathrm{d}_\text{af}\).

We argue here that this solution was ill-posed: multiple events can be produced by a single pixel during a temporal window, with each of these events requiring their own individual depths \(\mathrm{d}_\text{bf}\) and \(\mathrm{d}_\text{af}\). Furthermore, at the temporal window level, estimating two depth maps is redundant, as the \(\mathrm{D}_\text{af}\) depth map of a temporal window can be recovered from the \(\mathrm{D}_\text{bf}\) depth map of the following temporal window.

Since our objective here is not to estimate depths at the event level but at the temporal window level, we follow these conclusions and only estimate a single depth map for each temporal window of events, \ie, the \(\mathrm{D}_\text{bf}\) map.

\subsection{Architecture}\label{sec:method:architecture}
We propose a novel attention-based recurrent network to estimate depth maps from LiDAR and event data, resulting from an iterative refinement. As illustrated in \cref{fig:network}, our DELTA network is based on a U-Net architecture~\cite{Ronneberger2015UNetCN}, with two input branches for frame-like representations of the LiDAR and event data, a propagation memory, a central memory state, and a decoding branch.

\paragraph{Encoding heads}\label{sec:method:encod_heads}
To be able to apply attention on them, the event volumes and the projected LiDAR data (both of shape \((H, W, C)\), where \(C\) is the number of channels) are first split into \(N\) small patches of size \(P\times{}P\), as originally proposed by Dosovitskiy \etal~\cite{Dosovitskiy2020AnII}. This splitting is performed through stacked convolutional layers, and results in data of shape \((N, D)\), where \(N\) is the number of patches, and \(D\) is the dimensionality of each patch. As attention is an order-independent operation, these encoded patches are summed with a fixed 2-dimensional positional embedding, following the formulation of Carion \etal~\cite{Carion2020EndtoEndOD}. This way, each patch has its own unique signature, making the network able to distinguish them.

\paragraph{Data encoding and fusion}
Event and LiDAR patches each go through two self-attention modules, to encode their own internal relations. A cross-attention module (CA\textsubscript{F} in \cref{fig:network}) is then used to encode the cross-relations between the event and LiDAR patches, resulting in a single fused representation. This step is crucial for ensuring the accurate fusion of the two modalities, as shown in \cref{sec:eval:ablation}.

\paragraph{LiDAR propagation}
Due to the use of the cross-attention module CA\textsubscript{F}, contrary to the models of~\cite{Brebion2023LearningTE,Gehrig2021CombiningEA}, the LiDAR and event branches can not be totally decorrelated, as both LiDAR and event data are necessary at each time step. On the basis that event volumes
\begin{tikzpicture}[scale=0.7, every node/.style={transform shape}]
  \foreach \t in {1.95,2.05,...,2.65}
    \draw[line width=0.3mm,color_e] (\t,0.8) -- (\t,1.175);
  \draw[->,line width=0.4mm] (1.9,0.8) -- (2.7,0.8);
\end{tikzpicture}
are more frequently available than LiDAR point clouds
\begin{tikzpicture}[scale=0.7, every node/.style={transform shape}]
  \foreach \t in {1.95,2.15,...,2.65}
    \draw[line width=0.3mm,color_l] (\t,0.8) -- (\t,1.175);
  \draw[->,line width=0.4mm] (1.9,0.8) -- (2.7,0.8);
\end{tikzpicture}%
, unless new LiDAR data is available, we propagate at each time step the previous LiDAR data using the incoming event data
\begin{tikzpicture}[scale=0.7, every node/.style={transform shape}]
  \foreach \t in {2.05,2.25,...,2.45}
    \draw[line width=0.3mm,color_e] (\t,0.8) -- (\t,1.175);
  \foreach \t in {1.95,2.15,...,2.65}
    \draw[line width=0.3mm,color_l] (\t,0.8) -- (\t,1.175);
  \draw[->,line width=0.4mm] (1.9,0.8) -- (2.7,0.8);
\end{tikzpicture}%
. To do so, the input events update a small (\cref{sec:eval:impl_detail:memory_size}) propagation memory via a cross-attention module (CA\textsubscript{P1} in \cref{fig:network}). This propagation memory is then used in a second cross-attention module (CA\textsubscript{P2}), where the previous LiDAR data queries an update from the propagation memory in order to produce an updated LiDAR representation.

\paragraph{Memory update}\label{sec:method:memory}
Considering the case where the event camera and the LiDAR could be placed on a dynamic platform (\eg, a road vehicle), then if that platform was to come to a halt (\eg, at an intersection or in a traffic jam), few to no event would be produced by the camera. As such, densifying the LiDAR would become a difficult task, as the events would not be able to provide any guiding information. To solve this issue, given the sequential nature of the inputs, we add a central memory. This way, even if the event camera and LiDAR become static, the network can still exploit the memory of the previous information from these sensors to derive accurate predictions. This memory state also allows for temporal smoothness of the output of the network (as shown in \cref{sec:eval:ablation}), which is crucial given the high noise in the event data. Regarding the implementation, the fused LiDAR and event data are given to a Gated Recurrent Unit (GRU) module~\cite{Cho2014LearningPR}, which updates this central memory.

\paragraph{Decoding}
To obtain the final depth map, the data from the central memory passes through two self-attention modules, each followed by a skip connection with the corresponding summed and normalized LiDAR and event data from the encoding branch. To obtain an image-like output, a final decoding head regroups the decoded patches and reshapes the data to its original size. This decoding head is composed of stacked convex upsampling modules~\cite{Teed2020RAFTRA}, where the upsampling is guided by the corresponding data from the events encoding head. Output is of shape \((H, W, 1)\), \ie{}, the same spatial resolution as the inputs.

\subsection{Loss Functions}
To train DELTA, two complementary losses are used: a pixel-wise \(\ell_1\) loss \(\mathcal{L}_{\ell_1}\), and a multiscale gradient-matching loss \(\mathcal{L}_\text{msg}\)~\cite{Ummenhofer2016DeMoNDA}. The role of \(\mathcal{L}_{\ell_1}\) is to ensure the correctness of the depth prediction of all pixels compared to the ground truth. However, an \(\ell_1\) loss alone tends to produce blurry predictions, so a second regulatory loss is required to produce sharper depth maps. This is the role of \(\mathcal{L}_\text{msg}\), which ensures that the gradients of the predicted depth maps at different scales are consistent with the ones of the ground truth depth maps. We follow the formulation of~\cite{Brebion2023LearningTE} for \(\mathcal{L}_\text{msg}\), with the number of scales set to 5. Our final loss \(\mathcal{L}\) is a sum of these two losses over a sequence of length \(T\):
\begin{equation}
  \mathcal{L} = \sum_{t=0}^{T-1} (\mathcal{L}_{\ell_1}^t + \mathcal{L}_{\text{msg}}^t).
\end{equation}

\section{Evaluation}\label{sec:eval}
\textit{As a complement to this section, an additional ablation study, an analysis on computational complexity, and additional visual results are all available as Suppl.\ Material.}

\subsection{Datasets}\label{sec:eval:datasets}
To conduct our evaluation, we use in this work three datasets of the state of the art: the SLED dataset~\cite{Brebion2023LearningTE}, the MVSEC dataset~\cite{Zhu2018TheMS}, and the M3ED dataset~\cite{Chaney2023M3EDMM}.

\paragraph{SLED}
The SLED dataset~\cite{Brebion2023LearningTE} is a synthetic dataset recorded in 2023 using the CARLA simulator~\cite{Dosovitskiy2017CARLAAO}. It contains 20 minutes of perfectly synchronized and calibrated driving data, composed of high spatial definition (1280\texttimes{}720) events and images, point clouds from a 40-channel LiDAR at 10Hz with a maximum range of 200m, and dense ground truth depth maps.

\paragraph{MVSEC}
The MVSEC dataset~\cite{Zhu2018TheMS} was recorded in 2018. It contains 30 minutes of long outdoor driving sequences, with low spatial definition (346\texttimes{}260) events and images, point clouds from a 16-channel LiDAR at 20Hz with a maximum range of 100m, and semi-dense ground truth depth maps. However, the data is loosely synchronized, the calibration is approximate, and the ground truth depth maps are erroneous when there are moving objects in the scene. Yet, it remains the most popular dataset for depth estimation in the event community, making it an interesting benchmark.

\paragraph{M3ED}
The M3ED dataset~\cite{Chaney2023M3EDMM} was recorded in 2023, and acts as an informal successor to the MVSEC dataset. It is composed (among others) of 110 minutes of outdoor driving sequences, with high spatial definition stereo events (1280\texttimes{}720) and images (1280\texttimes{}800), point clouds from a 64-channel LiDAR at 10Hz with a maximum range of 120m, and sparse ground truth depth maps. However, given the size of the dataset, and since LiDAR data is not provided for the test set, we can not use it as is. To still be able to provide insightful results, we subsampled the dataset and redefined the train/val/test sets as described in \cref{tab:split_m3ed}.

\begin{table}
  \centering
  \resizebox{\linewidth}{!}{
    \begin{tabular}{@{}lcr@{}}
      \toprule
      Redefined set & Recordings & Total length \\
      \midrule
      \multirow{2}{*}{Train} & \Verb|penno_small_loop_day|; \Verb|rittenhouse_day|; & \multirow{2}{*}{8m02s} \\
      & \Verb|penno_small_loop_night| \\
      \midrule
      Val & \Verb|horse_day|; \Verb|ucity_small_loop_night| & 8m59s \\
      \midrule
      Test & \Verb|city_hall_day|; \Verb|city_hall_night| & 9m43s \\
      \bottomrule
    \end{tabular}
  }
  \caption{M3ED sets used within this work.}\label{tab:split_m3ed}
\end{table}

\paragraph{DSEC}
We also tried initially to use the popular DSEC dataset~\cite{Gehrig2021DSECAS}, but its ground truth depth maps are given at the timestamps of the RGB frames (and are thus not synchronized with the LiDAR point clouds)\footnote{\resizebox{0.94\linewidth}{!}{\tiny\url{https://github.com/uzh-rpg/DSEC/issues/7\#issuecomment-1416776152}}}, unfortunately making them incompatible with the training and evaluation of a LiDAR-and-event fusion method like ours.

\subsection{Implementation Details}
\paragraph{Data representation}\label{sec:eval:impl_detail:data_repres}
The event data is split in temporal windows of fixed size \(\Delta t = 50\text{ms}\), based on the rate of the ground truth of the three datasets in use. The events in each window are accumulated into an Event Volume of shape \((H, W, 4)\), following the formulation of Zhu \etal~\cite{Zhu2019UnsupervisedEL}. The LiDAR point clouds are represented as their projection on the event camera's image plane. Pixels without any depth value are set to 0. Both the LiDAR projections and ground truth depth maps are normalized between 0.0 and 1.0, where 1.0 is the maximum LiDAR range in the dataset in use.

\paragraph{Data size}\label{sec:eval:impl_detail:data_size}
The patch size \(P\) is set to 16 pixels for high-resolution data (SLED and M3ED), and 12 pixels for low-resolution (MVSEC) to better capture details. We use a standard dimensionality of \(D=1024\). During training, data is randomly cropped to a size of \(512 \times 512\) pixels for the high-resolution SLED and M3ED datasets, and to \(252 \times 252\) pixels for the low-resolution MVSEC dataset.

\paragraph{Memories size and initialization}\label{sec:eval:impl_detail:memory_size}
Since the role of the central memory is to condense past data, its shape must be the same as the data itself, \ie, \((N, D)\). On the contrary, the propagation memory is only a parametric representation of how the LiDAR data should be propagated to match the current events. While its dimensionality is still constrained to \(D\), its number of elements \(N\) can be tuned: we empirically chose a size of \(128\) in this work. As for their initialization, the central memory is initialized with a copy of the two-dimensional positional embedding, while the initial state of the propagation memory is learned.

\paragraph{Training details}
For training on the three datasets, we use the Adam optimizer~\cite{Kingma2015AdamAM} with batch size \(B = 4\). When training from scratch on the SLED and the M3ED datasets, 100 and 50 epochs are used respectively, the initial learning rate is set to \(10^{-4}\), and it is decayed by \(0.01^{1/99}\) and \(0.01^{1/49}\) respectively after each epoch (in order to reach a learning rate of \(10^{-6}\) at the last epoch). When training from scratch on the MVSEC dataset, 20 epochs are used, and the learning rate is set to a constant value of \(10^{-4}\). When finetuning on MVSEC or M3ED, 5 epochs are used, and the learning rate is set to a constant value of \(10^{-5}\).

\paragraph{Evaluation metrics}
For all datasets, we use the mean absolute depth error metric~\cite{Brebion2023LearningTE,Cui2022DenseDE,Gehrig2021CombiningEA,Ranon2021StereoSpikeDL,Sabater2022EventTA+}, but also the standard depth estimation metrics of~\cite{Eigen2014DepthMP} (AbsRel, RMS, RMSlog, \(\delta^\text{1,2,3}\)). Following the convention on the MVSEC dataset~\cite{Zhu2018TheMS}, results are presented with five cutoff distances (10m, 20m, 30m, half the maximum range, and the maximum range). DELTA is trained on the full depth maps, \ie, at the maximum range, and these cutoffs are only applied at test time. For fairness of evaluation, comparisons with ALED~\cite{Brebion2023LearningTE} are only made on its \(\mathrm{D}_\text{bf}\) depth maps.

\subsection{Results on the SLED Dataset}\label{sec:eval:results_sled}
We begin by training DELTA solely on the SLED dataset, and denote this version DELTA\textsubscript{SL}. Results of DELTA\textsubscript{SL} on the testing set of SLED are given in \cref{tab:results_sled}.
Compared to the state-of-the-art results of ALED\textsubscript{SL}~\cite{Brebion2023LearningTE} (also trained only on SLED, noted ALED\textsubscript{S} in~\cite{Brebion2023LearningTE}), a clear improvement can be seen across all metrics. Improvement is particularly important for nearby objects, as the mean depth error at the 10m cutoff is divided nearly by 2 and by 4 for the \verb|Town01| and \verb|Town03| maps respectively. Improvement is less significant at longer ranges, but we argue that close surroundings of the vehicle are of much more importance when driving than distant objects.

\begin{table*}
  \centering
  \resizebox{0.9\linewidth}{!}{
    \begin{tabular}{@{}cccccccccccccccc@{}}
      \toprule
      \multirow{2}[2]{*}{\textbf{Map}} & \multirow{2}[2]{*}{\textbf{Cutoff}} & \multicolumn{7}{c}{\textbf{ALED\textsubscript{SL}}~\cite{Brebion2023LearningTE}} & \multicolumn{7}{c}{\textbf{DELTA\textsubscript{SL}}} \\
      \cmidrule(lr){3-9}\cmidrule(lr){10-16}
      & & Mean\(\downarrow\) & AbsRel\(\downarrow\) & RMS\(\downarrow\) & RMSlog\(\downarrow\) & \(\delta^1\)\(\uparrow\) & \(\delta^2\)\(\uparrow\) & \(\delta^3\)\(\uparrow\) & Mean\(\downarrow\) & AbsRel\(\downarrow\) & RMS\(\downarrow\) & RMSlog\(\downarrow\) & \(\delta^1\)\(\uparrow\) & \(\delta^2\)\(\uparrow\) & \(\delta^3\)\(\uparrow\) \\
      \midrule
      \multirow{5}{*}{\Verb|Town01|} & 10m & 1.24 & 0.211 & 8.022 & 0.478 & 0.890 & 0.962 & 0.978 & \textbf{0.66} & \textbf{0.106} & \textbf{6.981} & \textbf{0.282} & \textbf{0.963} & \textbf{0.981} & \textbf{0.989} \\
      & 20m & 2.10 & 0.232 & 11.973 & 0.489 & 0.885 & 0.952 & 0.970 & \textbf{1.33} & \textbf{0.133} & \textbf{10.601} & \textbf{0.311} & \textbf{0.948} & \textbf{0.973} & \textbf{0.983} \\
      & 30m & 2.74 & 0.239 & 13.831 & 0.481 & 0.875 & 0.945 & 0.966 & \textbf{1.91} & \textbf{0.147} & \textbf{12.437} & \textbf{0.317} & \textbf{0.932} & \textbf{0.965} & \textbf{0.979} \\
      & 100m & 4.26 & 0.241 & 17.182 & 0.468 & 0.860 & 0.935 & 0.959 & \textbf{3.22} & \textbf{0.157} & \textbf{15.177} & \textbf{0.317} & \textbf{0.909} & \textbf{0.954} & \textbf{0.973} \\
      & 200m & \textbf{4.53} & 0.173 & \textbf{18.775} & 0.405 & 0.892 & 0.947 & 0.966 & 4.54 & \textbf{0.119} & 19.606 & \textbf{0.311} & \textbf{0.921} & \textbf{0.957} & \textbf{0.973} \\
      \midrule
      \multirow{5}{*}{\Verb|Town03|} & 10m & 2.01 & 0.290 & 14.900 & 0.456 & 0.904 & 0.952 & 0.966 & \textbf{0.54} & \textbf{0.082} & \textbf{5.656} & \textbf{0.184} & \textbf{0.958} & \textbf{0.973} & \textbf{0.986} \\
      & 20m & 2.87 & 0.301 & 17.134 & 0.516 & 0.883 & 0.939 & 0.961 & \textbf{1.31} & \textbf{0.117} & \textbf{9.958} & \textbf{0.234} & \textbf{0.934} & \textbf{0.962} & \textbf{0.980} \\
      & 30m & 3.35 & 0.292 & 17.706 & 0.507 & 0.874 & 0.935 & 0.959 & \textbf{1.93} & \textbf{0.133} & \textbf{11.831} & \textbf{0.251} & \textbf{0.919} & \textbf{0.952} & \textbf{0.974} \\
      & 100m & 4.62 & 0.275 & 18.840 & 0.484 & 0.860 & 0.928 & 0.955 & \textbf{3.40} & \textbf{0.146} & \textbf{15.119} & \textbf{0.265} & \textbf{0.897} & \textbf{0.941} & \textbf{0.968} \\
      & 200m & 4.87 & 0.216 & 20.059 & 0.438 & 0.882 & 0.937 & 0.960 & \textbf{4.63} & \textbf{0.122} & \textbf{19.363} & \textbf{0.271} & \textbf{0.906} & \textbf{0.944} & \textbf{0.967} \\
      \bottomrule
    \end{tabular}
  }
  \caption{Error metrics of ALED\textsubscript{SL}~\cite{Brebion2023LearningTE} and of DELTA\textsubscript{SL} on the SLED dataset for various cutoff depth distances. Best results are in bold.}\label{tab:results_sled}
\end{table*}

\begin{figure}
  \centering
  \setlength\tabcolsep{1pt}
  \renewcommand{\arraystretch}{0.5}
  \begin{tabular}{@{}cccc@{}}
    \multirow{2}{*}[-0.75cm]{\footnotesize \textbf{(\textcolor{cvprblue}{a})}} & \footnotesize Ground truth & \footnotesize ALED\textsubscript{SL}~\cite{Brebion2023LearningTE} result & \footnotesize DELTA\textsubscript{SL} result \\
    &
    \includegraphics[width=0.31\linewidth]{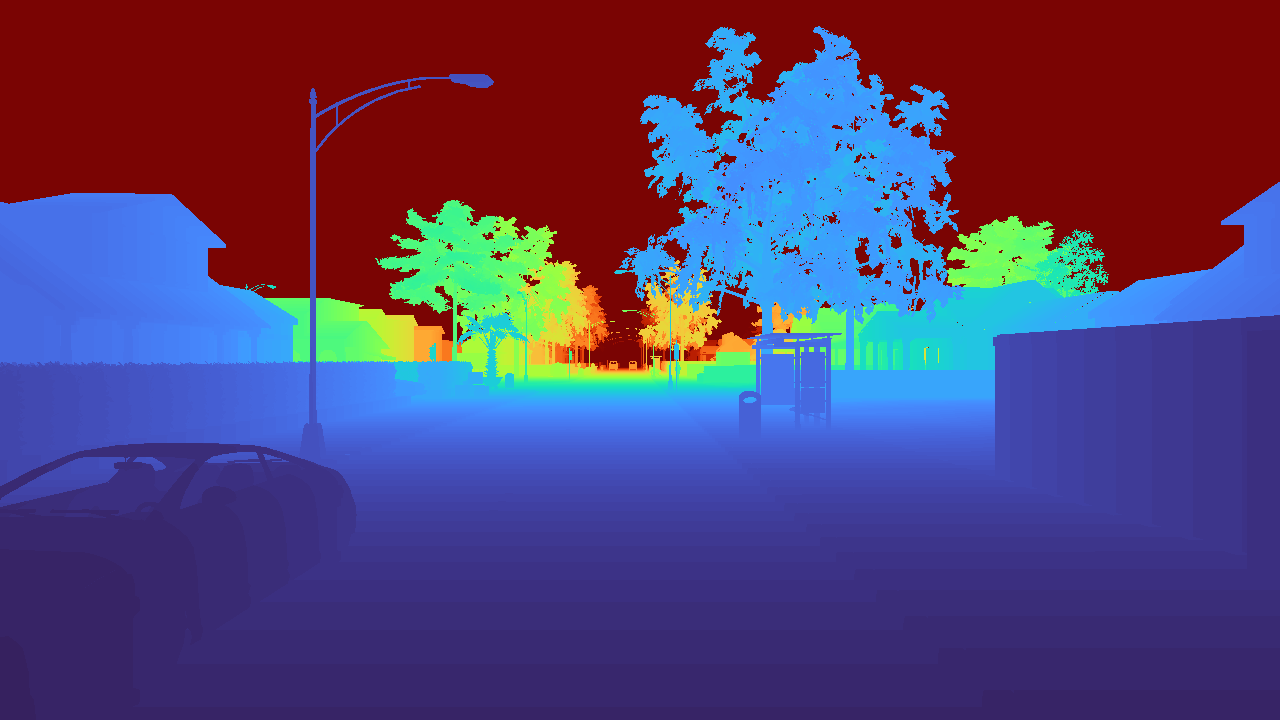} &
    \includegraphics[width=0.31\linewidth]{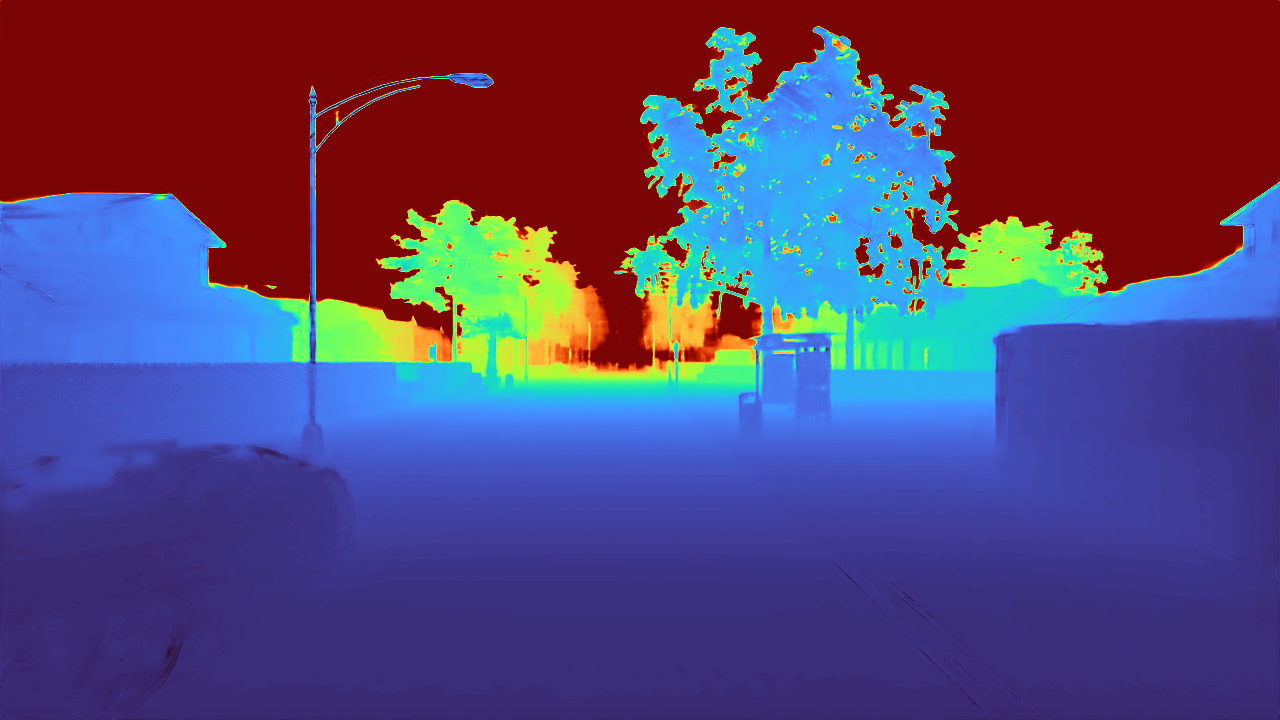} &
    \includegraphics[width=0.31\linewidth]{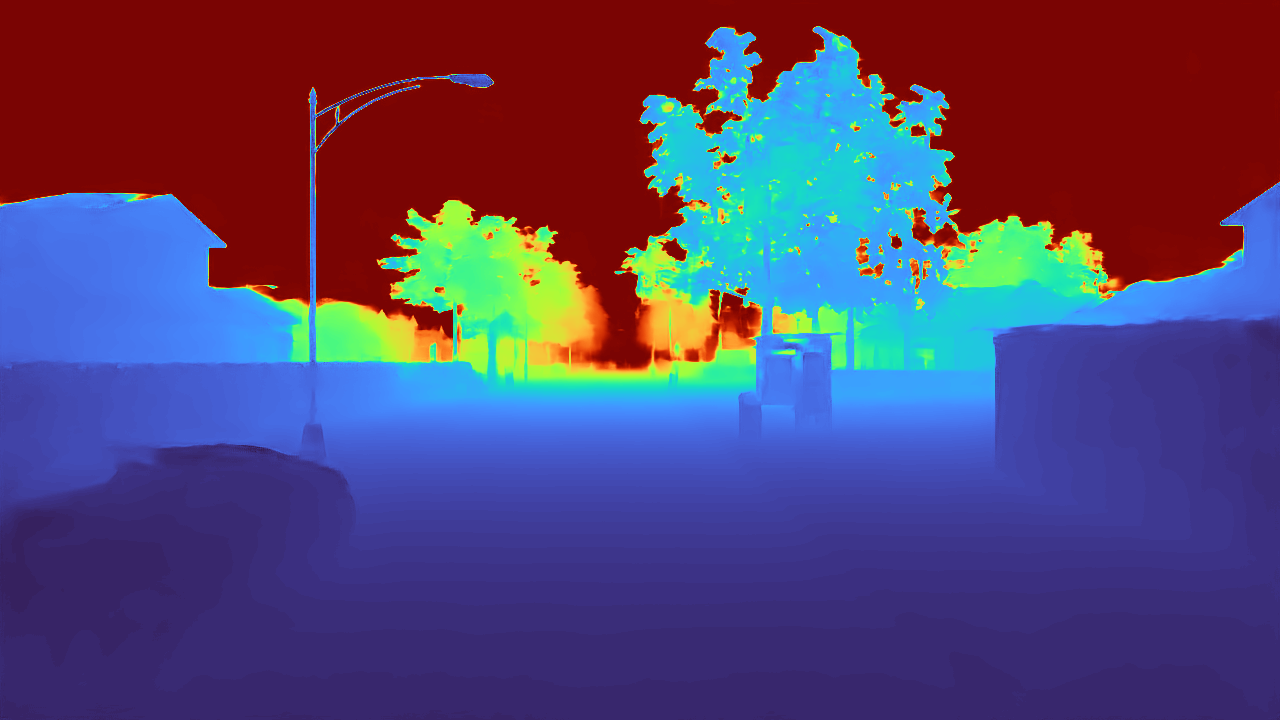} \\
    \multirow{2}{*}[-0.75cm]{\footnotesize \textbf{(\textcolor{cvprblue}{b})}} & & \footnotesize ALED\textsubscript{SL}~\cite{Brebion2023LearningTE} error map & \footnotesize DELTA\textsubscript{SL} error map \\
    &
    &
    \includegraphics[width=0.31\linewidth]{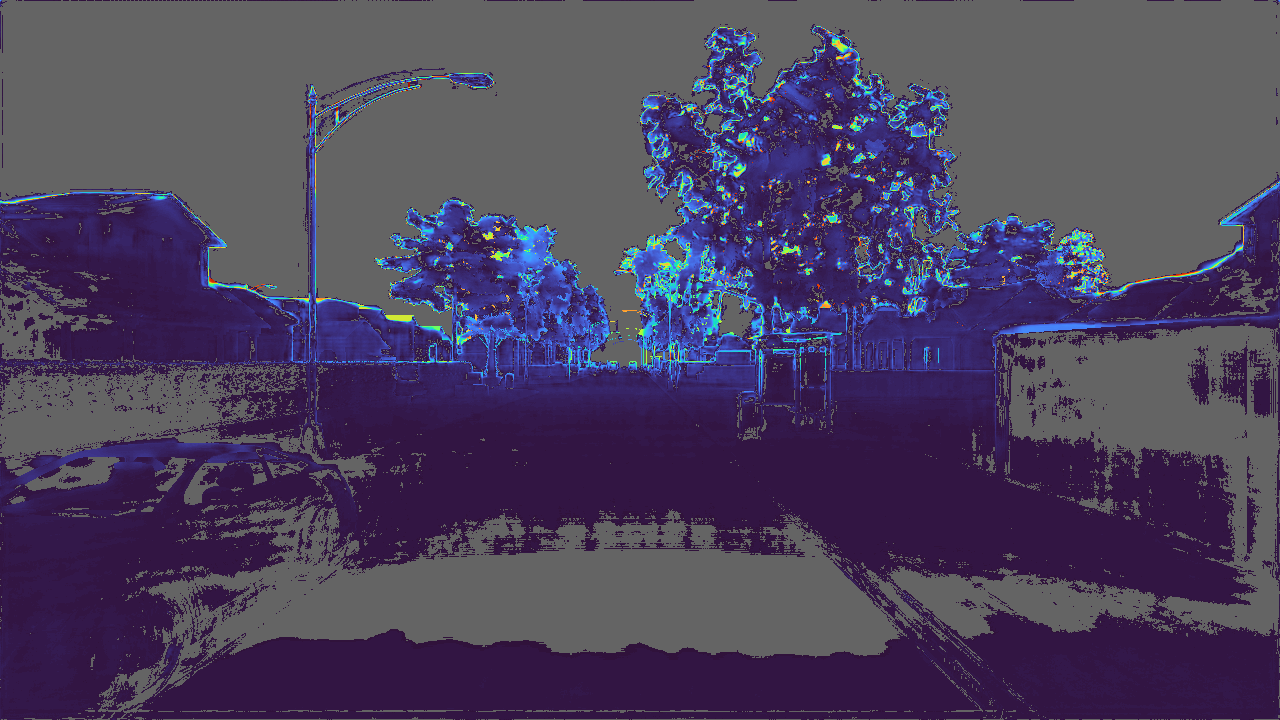} &
    \includegraphics[width=0.31\linewidth]{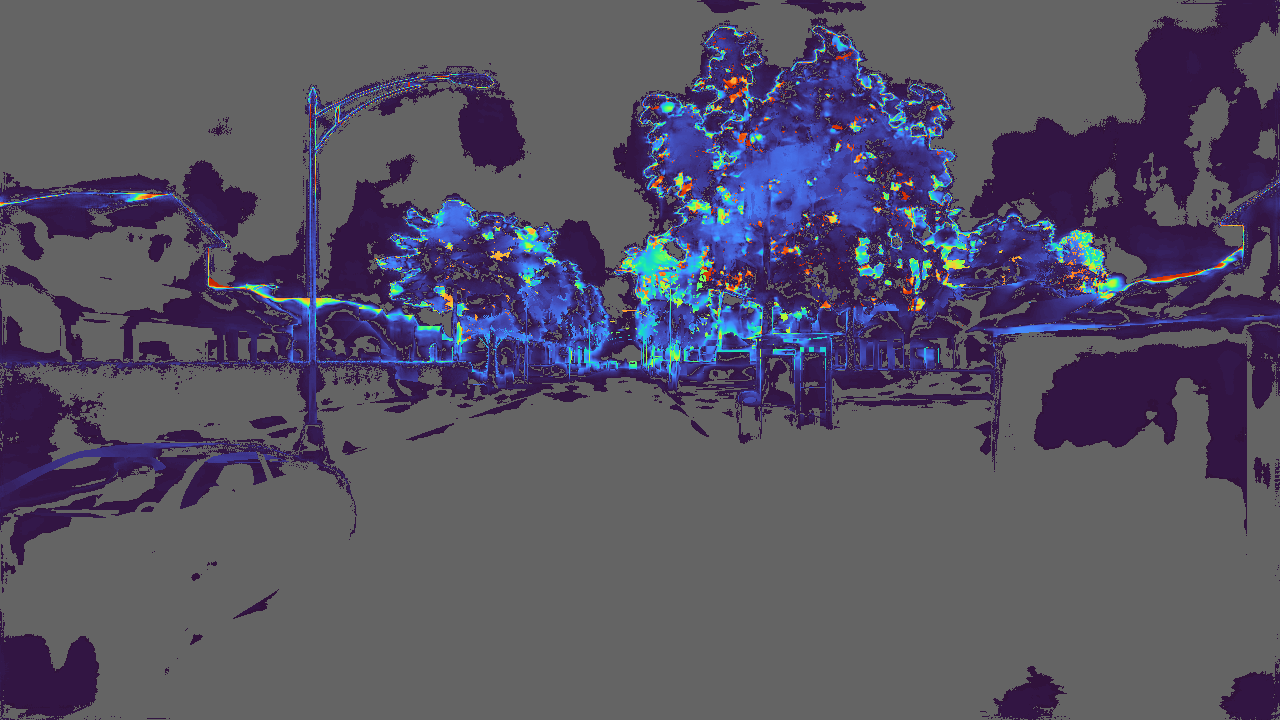} \\
    \\
    \multirow{2}{*}[-0.75cm]{\footnotesize \textbf{(\textcolor{cvprblue}{c})}} & \footnotesize Ground truth & \footnotesize ALED\textsubscript{SL}~\cite{Brebion2023LearningTE} result & \footnotesize DELTA\textsubscript{SL} result \\
    &
    \includegraphics[width=0.31\linewidth]{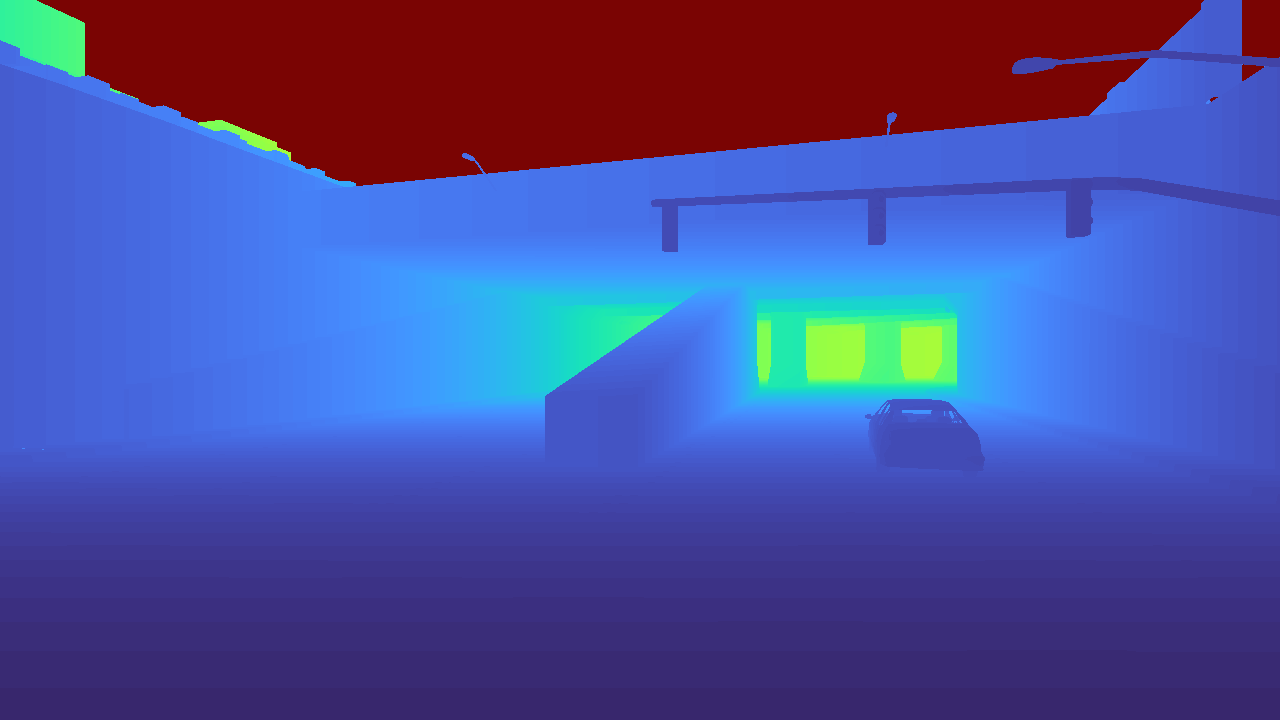} &
    \includegraphics[width=0.31\linewidth]{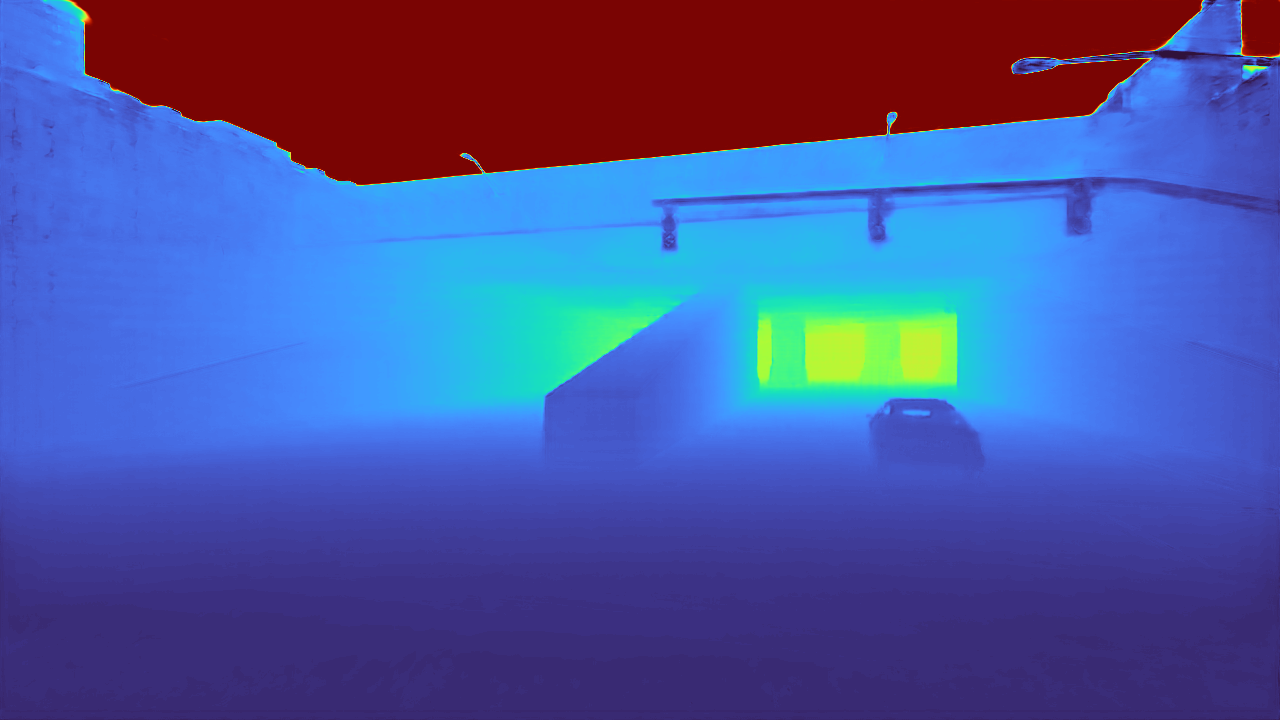} &
    \includegraphics[width=0.31\linewidth]{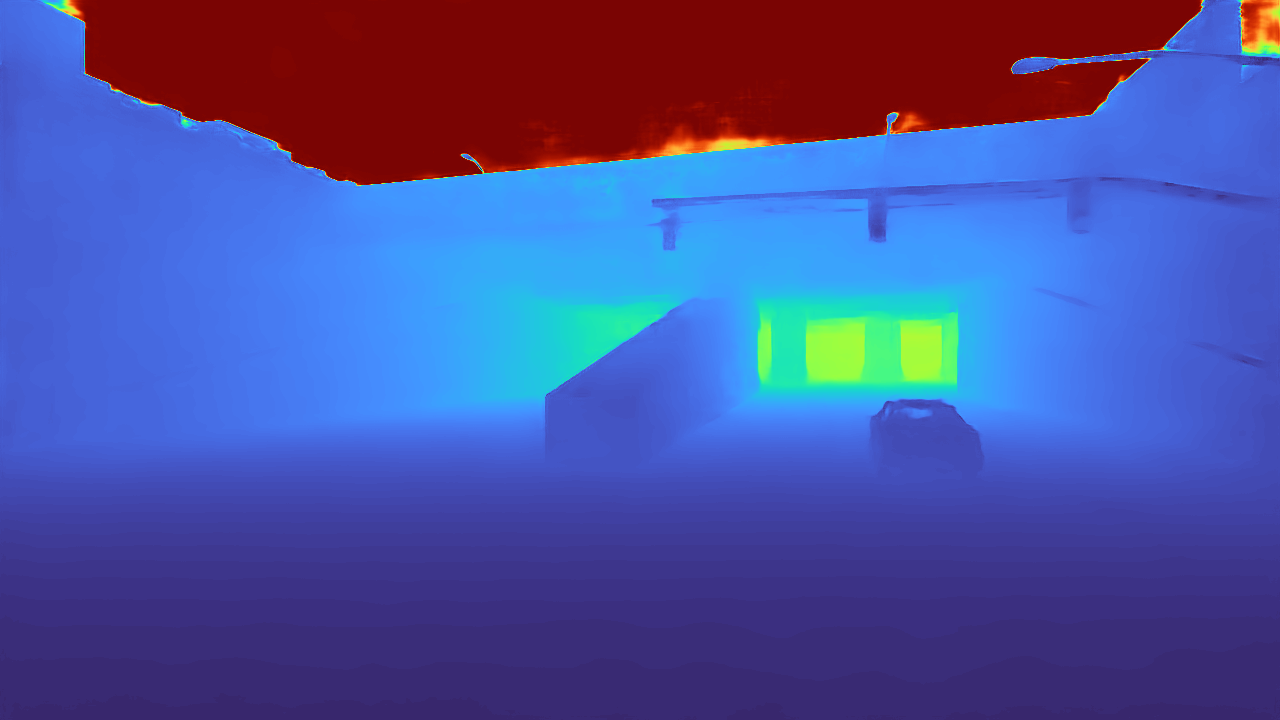} \\
    \multirow{2}{*}[-0.75cm]{\footnotesize \textbf{(\textcolor{cvprblue}{d})}} & & \footnotesize ALED\textsubscript{SL}~\cite{Brebion2023LearningTE} error map & \footnotesize DELTA\textsubscript{SL} error map \\
    &
    &
    \includegraphics[width=0.31\linewidth]{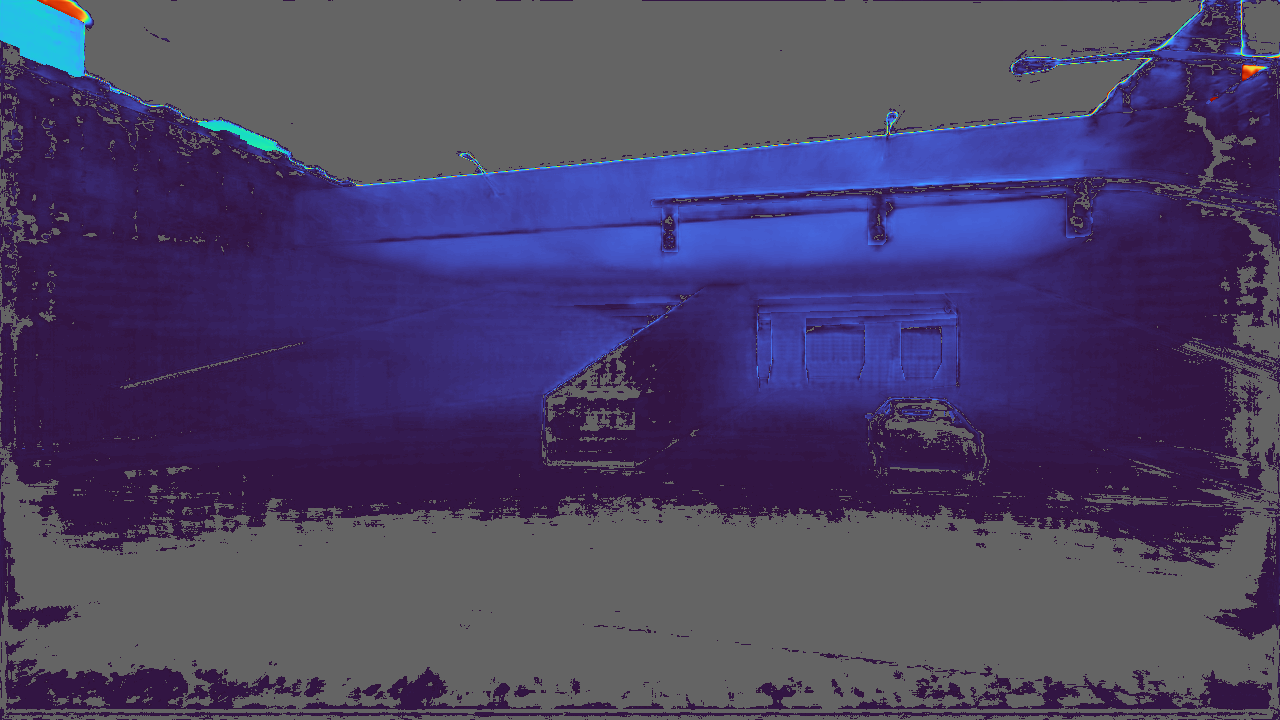} &
    \includegraphics[width=0.31\linewidth]{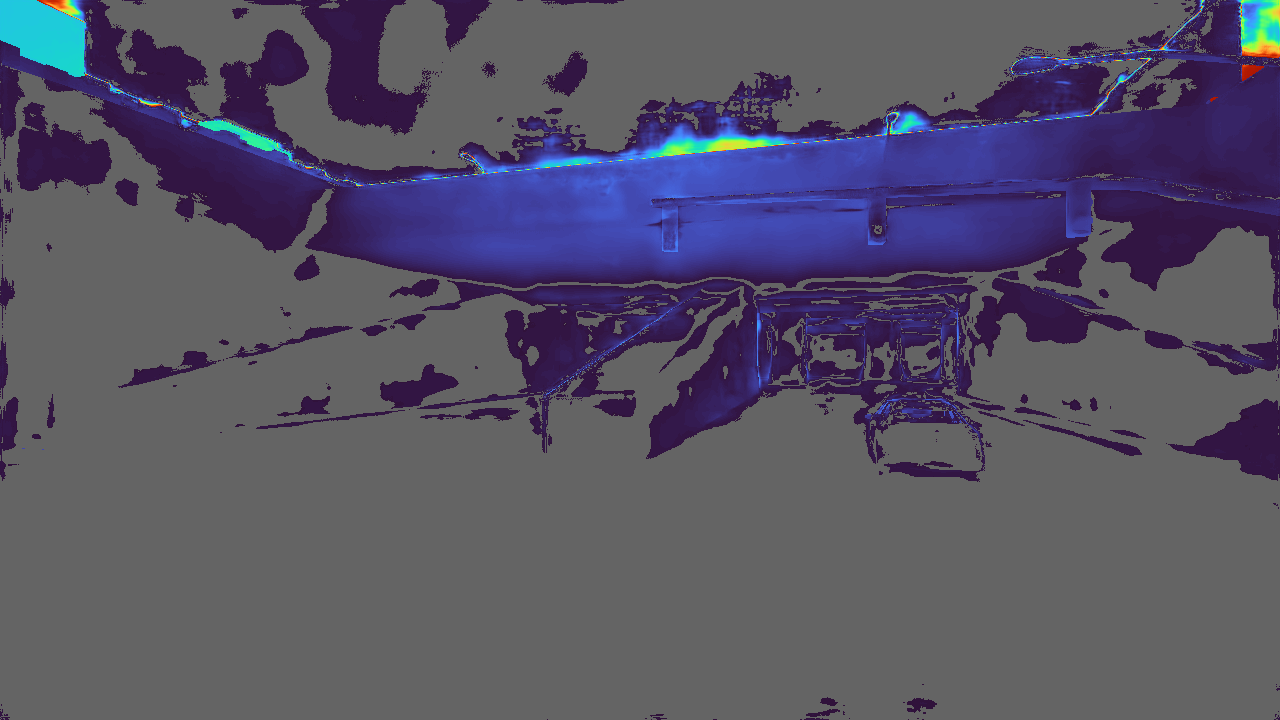} \\
  \end{tabular}
  \cprotect\caption{Results on the \verb|Town01_08| ({\footnotesize\textbf{(\textcolor{cvprblue}{a})}}, {\footnotesize\textbf{(\textcolor{cvprblue}{b})}}) and \verb|Town03_19| ({\footnotesize\textbf{(\textcolor{cvprblue}{c})}}, {\footnotesize\textbf{(\textcolor{cvprblue}{d})}}) sequences of SLED. Rows {\footnotesize\textbf{(\textcolor{cvprblue}{a})}} and {\footnotesize\textbf{(\textcolor{cvprblue}{c})}}, left to right: ground truth depth map; result from ALED\textsubscript{SL}; our result (DELTA\textsubscript{SL}). Differences between ALED\textsubscript{SL} and DELTA\textsubscript{SL} are better seen in rows {\footnotesize\textbf{(\textcolor{cvprblue}{b})}} and {\footnotesize\textbf{(\textcolor{cvprblue}{d})}}, showing the error maps of ALED\textsubscript{SL} and DELTA\textsubscript{SL} (where pixels with an error inferior to 0.5m are in gray). \textit{For a better visualization, an enlarged version of this figure is given in the Supplementary Material.}}\label{fig:cmp_sled}
\end{figure}

Visual results are presented in \cref{fig:cmp_sled}. Looking at the predicted depth maps (rows {\footnotesize\textbf{(\textcolor{cvprblue}{a})}} and {\footnotesize\textbf{(\textcolor{cvprblue}{c})}}), our network is able to infer very accurate results, visually close to the ground truth. Looking at the error maps  (rows {\footnotesize\textbf{(\textcolor{cvprblue}{b})}} and {\footnotesize\textbf{(\textcolor{cvprblue}{d})}}), they confirm the observations made on the quantitative results: we commit very small errors at close ranges (especially for the ground) and larger errors at longer ranges, while ALED commits medium to large errors over the whole depth map. These results also highlight that, despite a lack of LiDAR data for the road just in front of the vehicle (at the very bottom of the images), our network can accurately estimate depth for these areas thanks to its attention-based design, which can correlate all the patches of event and LiDAR data (compared to the convolutional-based design of ALED, which only operates on limited neighborhoods, and thus does not achieve good results for these areas).

\subsection{Results on the MVSEC Dataset}

\begin{figure*}
  \centering
  \setlength\tabcolsep{1pt}
  \renewcommand{\arraystretch}{0.5}
  \begin{tabular}{@{}ccccccc@{}}
    \small Events & \small LiDAR proj. & \small Ground truth & \small Cui \etal~\cite{Cui2022DenseDE} & \small ALED\textsubscript{SL\(\rightarrow\)MV}~\cite{Brebion2023LearningTE} & \small DELTA\textsubscript{MV} & \small DELTA\textsubscript{SL\(\rightarrow\)MV} \\
    \includegraphics[width=0.135\linewidth]{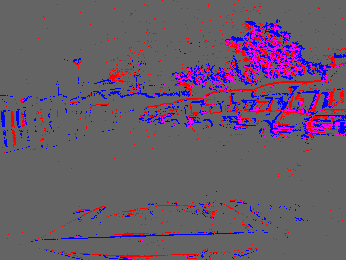} &
    \includegraphics[width=0.135\linewidth]{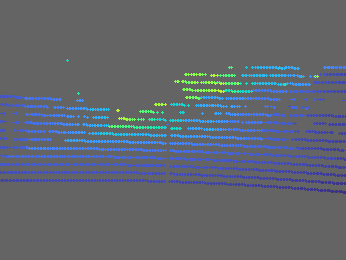} &
    \includegraphics[width=0.135\linewidth]{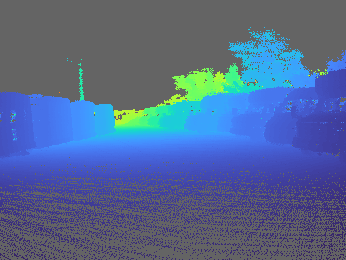} &
    \includegraphics[width=0.135\linewidth]{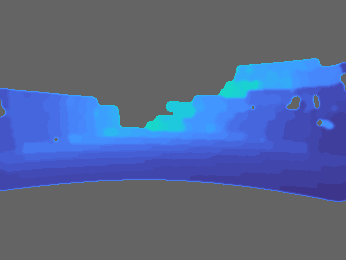} &
    \includegraphics[width=0.135\linewidth]{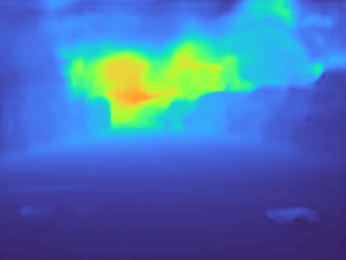} &
    \includegraphics[width=0.135\linewidth]{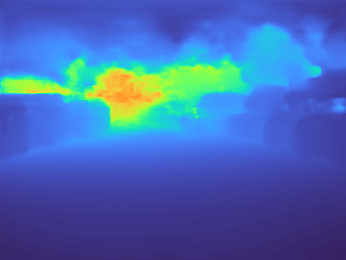} &
    \includegraphics[width=0.135\linewidth]{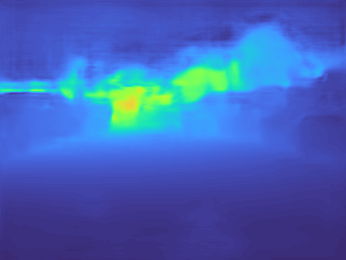} \\
    \includegraphics[width=0.135\linewidth]{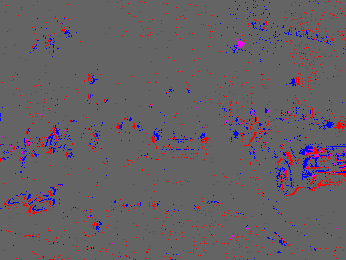} &
    \includegraphics[width=0.135\linewidth]{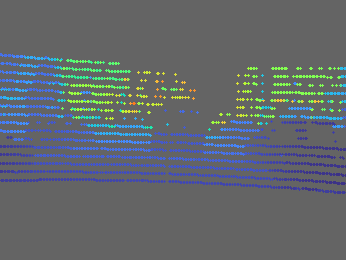} &
    \includegraphics[width=0.135\linewidth]{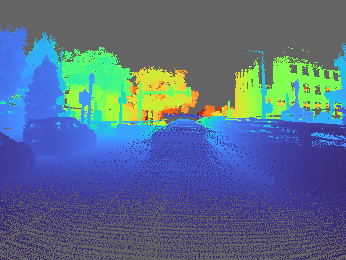} &
    \includegraphics[width=0.135\linewidth]{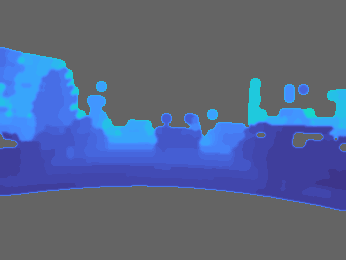} &
    \includegraphics[width=0.135\linewidth]{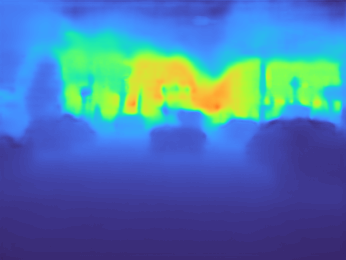} &
    \includegraphics[width=0.135\linewidth]{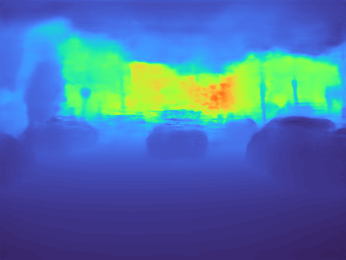} &
    \includegraphics[width=0.135\linewidth]{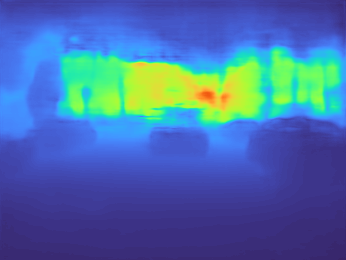} \\
    \includegraphics[width=0.135\linewidth]{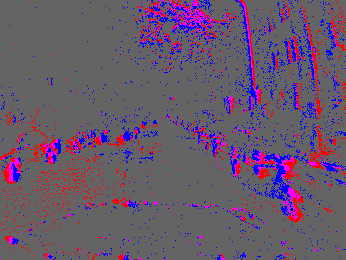} &
    \includegraphics[width=0.135\linewidth]{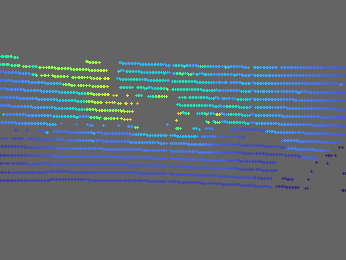} &
    \includegraphics[width=0.135\linewidth]{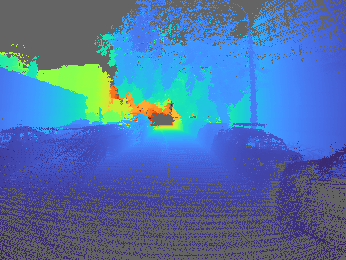} &
    \includegraphics[width=0.135\linewidth]{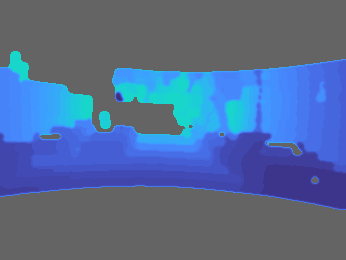} &
    \includegraphics[width=0.135\linewidth]{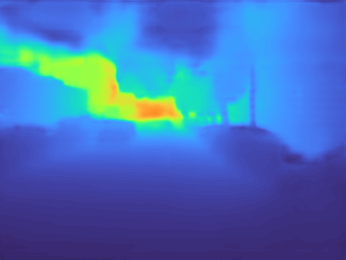} &
    \includegraphics[width=0.135\linewidth]{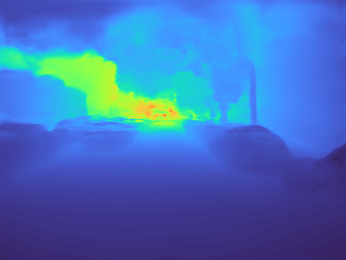} &
    \includegraphics[width=0.135\linewidth]{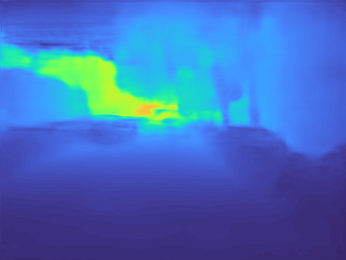}
  \end{tabular}
  \cprotect\caption{Qualitative results on the MVSEC dataset. Left to right: events; LiDAR projection (with larger points for a better readability); ground truth; results from Cui \etal~\cite{Cui2022DenseDE}; results from ALED~\cite{Brebion2023LearningTE}; our results. Top to bottom: {\footnotesize\verb|outdoor_day_1|}; {\footnotesize\verb|outdoor_night_1|}; {\footnotesize\verb|outdoor_night_2|}. \textit{For a better visualization, an enlarged version of this figure is given in the Supplementary Material.}}\label{fig:cmp_mvsec}
\end{figure*}

For the MVSEC dataset, we propose two variants:
\begin{itemize}
  \item DELTA\textsubscript{MV}, only trained on MVSEC;
  \item DELTA\textsubscript{SL\(\rightarrow\)MV}, after finetuning DELTA\textsubscript{SL} on MVSEC.
\end{itemize}
The results of both versions are given in \cref{tab:results_mvsec_mean,tab:results_mvsec_all_metrics}, in addition to the results of other methods of the state of the art.
Compared to the results of all other methods (except ALED~\cite{Brebion2023LearningTE}), our method yields consistently lower errors, especially after pre-training on the SLED dataset.
Compared to ALED\textsubscript{MV} (only trained on MVSEC, noted ALED\textsubscript{R} in~\cite{Brebion2023LearningTE}), DELTA\textsubscript{MV} offers a significant improvement, similar to the one observed on the SLED dataset in \cref{sec:eval:results_sled}. Compared to ALED\textsubscript{SL\(\rightarrow\)MV} (pre-trained on SLED and finetuned on MVSEC, noted ALED\textsubscript{S\(\rightarrow\)R} in~\cite{Brebion2023LearningTE}), DELTA\textsubscript{SL\(\rightarrow\)MV} also offers an improvement for close ranges, and remains close for more distant ranges.
Comparing DELTA\textsubscript{MV} and DELTA\textsubscript{SL\(\rightarrow\)MV} reveals however that the finetuning is not as efficient as between ALED\textsubscript{SL} and ALED\textsubscript{SL\(\rightarrow\)MV}, with the mean depth error only improving slightly in our case. We believe this is due to the large change of resolution between the SLED and MVSEC datasets, which is a well-documented issue with attention-based vision models~\cite{Fitzgerald2022MultiresolutionFO}, and which requires in our case a change in patch size and a redefinition of the positional encoding. This makes the finetuning naturally more complex for an attention-based network than for a purely convolutional one. Additionally, as showcased in \cref{fig:cmp_mvsec}, fewer ground truth points are available for the close ranges (where our method performs best), skewing the results in favor of ALED (which has more difficulties for close objects, as shown previously in \cref{sec:eval:results_sled}).

\begin{table}
  \centering
  \resizebox{\linewidth}{!}{
    \begin{tabular}{@{}cccccccccc@{}}
      \toprule
      \multirow{3}[2]{*}{\textbf{Recording}} & \multirow{3}[2]{*}{\textbf{Cutoff}} & \textbf{Events (stereo)} & \multicolumn{2}{c}{\textbf{Events \& Frames}} & \multicolumn{5}{c}{\textbf{Events \& LiDAR}} \\ \cmidrule(lr){3-3} \cmidrule(lr){4-5} \cmidrule(lr){6-10}
       & & StereoSpike & RAMNet & EvT\textsuperscript{+} & Cui \etal & ALED\textsubscript{MV} & \multirow{2}{*}{\textbf{DELTA\textsubscript{MV}}} & ALED\textsubscript{SL\(\rightarrow\)MV} & \multirow{2}{*}{\textbf{DELTA\textsubscript{SL\(\rightarrow\)MV}}} \\
      & & \cite{Ranon2021StereoSpikeDL} & \cite{Gehrig2021CombiningEA} & \cite{Sabater2022EventTA+} & \cite{Cui2022DenseDE} & \cite{Brebion2023LearningTE} & & \cite{Brebion2023LearningTE} \\
      \midrule
      \multirow{5}{*}{\Verb|outdoor_day_1|} & 10m & 0.79 & 1.39 & 1.24 & 1.24 & 0.91 & 0.51 & \textbf{0.50} & \textbf{0.50} \\
      & 20m & 1.47 & 2.17 & 1.91 & 1.28 & 1.22 & 0.86 & \textbf{0.80} & \underline{0.81} \\
      & 30m & 1.92 & 2.76 & 2.36 & 4.87 & 1.43 & 1.10 & \textbf{1.02} & \underline{1.06} \\
      & 50m & - & - & - & - & 1.67 & 1.42 & \textbf{1.31} & \underline{1.41} \\
      & 100m & 3.17 & - & - & - & 1.96 & \underline{1.73} & \textbf{1.60} & 1.80 \\
      \midrule
      \multirow{5}{*}{\Verb|outdoor_night_1|} & 10m & \textbf{1.38} & 2.50 & \underline{1.45} & 2.26 & 1.75 & 1.55 & 1.52 & 1.52 \\
      & 20m & 2.26 & 3.19 & 2.10 & 2.19 & 2.10 & 1.94 & \textbf{1.81} & \underline{1.92} \\
      & 30m & 2.97 & 3.82 & 2.88 & 4.50 & 2.25 & \underline{2.14} & \textbf{1.95} & 2.16 \\
      & 50m & - & - & - & - & 2.44 & \underline{2.42} & \textbf{2.20} & 2.46 \\
      & 100m & 4.82 & - & - & - & \underline{2.73} & 2.78 & \textbf{2.54} & 2.88 \\
      \midrule
      \multirow{5}{*}{\Verb|outdoor_night_2|} & 10m & - & 1.21 & 1.48 & 1.88 & 1.19 & 1.16 & 1.09 & \textbf{0.99} \\
      & 20m & - & 2.31 & 2.13 & 2.14 & 1.65 & 1.59 & \textbf{1.49} & \textbf{1.49} \\
      & 30m & - & 3.28 & 2.90 & 4.67 & 1.81 & 1.77 & \textbf{1.64} & \underline{1.75} \\
      & 50m & - & - & - & - & \underline{1.95} & \underline{1.95} & \textbf{1.80} & 1.98 \\
      & 100m & - & - & - & - & \underline{2.11} & 2.16 & \textbf{1.97} & 2.23 \\
      \midrule
      \multirow{5}{*}{\Verb|outdoor_night_3|} & 10m & - & 1.01 & 1.38 & 1.78 & 0.85 & 0.92 & \underline{0.81} & \textbf{0.74} \\
      & 20m & - & 2.34 & 2.03 & 1.93 & \underline{1.25} & 1.35 & \textbf{1.16} & 1.26 \\
      & 30m & - & 3.43 & 2.77 & 4.55 & \underline{1.42} & 1.57 & \textbf{1.33} & 1.56 \\
      & 50m & - & - & - & - & \underline{1.57} & 1.78 & \textbf{1.51} & 1.83 \\
      & 100m & - & - & - & - & \underline{1.73} & 1.96 & \textbf{1.66} & 2.09 \\
      \bottomrule
    \end{tabular}
  }
  \caption{Mean depth errors (in meters) on the MVSEC dataset for various cutoff depth distances.}\label{tab:results_mvsec_mean}
\end{table}

\begin{table}
  \centering
  \resizebox{\linewidth}{!}{
    \begin{tabular}{@{}ccccccccc@{}}
      \toprule
      \multirow{3}[2]{*}{\textbf{Recording}} & \multirow{3}[2]{*}{\textbf{Metric}} & \multicolumn{3}{c}{\textbf{Events \& Frames}} & \multicolumn{4}{c}{\textbf{Events \& LiDAR}} \\ \cmidrule(lr){3-5} \cmidrule(lr){6-9}
      & & RAMNet & HMNet & PCDepth & ALED\textsubscript{MV} & \multirow{2}{*}{\textbf{DELTA\textsubscript{MV}}} & ALED\textsubscript{SL\(\rightarrow\)MV} & \multirow{2}{*}{\textbf{DELTA\textsubscript{SL\(\rightarrow\)MV}}} \\
      & & \cite{Gehrig2021CombiningEA} & \cite{Hamaguchi2023HierarchicalNM} & \cite{Liu2024PCDepthPC} & \cite{Brebion2023LearningTE} & & \cite{Brebion2023LearningTE} \\
      \midrule
      \multirow{6}{*}{\Verb|outdoor_day_1|} & AbsRel\(\downarrow\) & 0.303 & 0.230 & 0.228 & 0.185 & \underline{0.118} & \textbf{0.114} & 0.121 \\
      & RMS\(\downarrow\) & 8.526 & 6.922 & 6.526 & 4.947 & \underline{4.793} & \textbf{4.574} & 4.910 \\
      & RMSlog\(\downarrow\) & 0.424 & 0.310 & 0.301 & 0.259 & \underline{0.215} & \textbf{0.200} & 0.236 \\
      & \(\delta^1\)\(\uparrow\) & 0.541 & 0.717 & 0.712 & 0.834 & \underline{0.890} & \textbf{0.895} & 0.866 \\
      & \(\delta^2\)\(\uparrow\) & 0.778 & 0.868 & 0.867 & 0.937 & 0.953 & \textbf{0.958} & \underline{0.954} \\
      & \(\delta^3\)\(\uparrow\) & 0.877 & 0.940 & 0.941 & 0.970 & \underline{0.977} & \textbf{0.980} & \textbf{0.980} \\
      \midrule
      \multirow{6}{*}{\Verb|outdoor_night_1|} & AbsRel\(\downarrow\) & 0.583 & 0.349 & \textbf{0.271} & 0.310 & 0.288 & \underline{0.274} & 0.289 \\
      & RMS\(\downarrow\) & 13.340 & 10.818 & 6.715 & 6.122 & \textbf{5.779} & \underline{5.791} & 6.042 \\
      & RMSlog\(\downarrow\) & 0.830 & 0.543 & \textbf{0.354} & 0.378 & \underline{0.360} & \underline{0.360} & 0.372 \\
      & \(\delta^1\)\(\uparrow\) & 0.296 & 0.497 & 0.632 & \underline{0.753} & 0.752 & \textbf{0.767} & 0.726 \\
      & \(\delta^2\)\(\uparrow\) & 0.502 & 0.661 & 0.822 & 0.861 & \underline{0.866} & \textbf{0.870} & 0.853 \\
      & \(\delta^3\)\(\uparrow\) & 0.635 & 0.784 & \underline{0.922} & 0.916 & \textbf{0.925} & \underline{0.922} & 0.917 \\
      \midrule
      \multirow{6}{*}{\Verb|outdoor_night_2|} & AbsRel\(\downarrow\) & - & - & - & 0.198 & 0.195 & \textbf{0.183} & \underline{0.184} \\
      & RMS\(\downarrow\) & - & - & - & 4.739 & \underline{4.636} & \textbf{4.543} & 4.870 \\
      & RMSlog\(\downarrow\) & - & - & - & 0.282 & \underline{0.281} & \textbf{0.274} & \underline{0.281} \\
      & \(\delta^1\)\(\uparrow\) & - & - & - & 0.802 & \underline{0.809} & \textbf{0.817} & 0.785 \\
      & \(\delta^2\)\(\uparrow\) & - & - & - & 0.911 & \underline{0.913} & \textbf{0.916} & 0.909 \\
      & \(\delta^3\)\(\uparrow\) & - & - & - & 0.958 & \underline{0.959} & \textbf{0.960} & \underline{0.959} \\
      \midrule
      \multirow{6}{*}{\Verb|outdoor_night_3|} & AbsRel\(\downarrow\) & - & - & - & \underline{0.142} & 0.155 & \textbf{0.133} & 0.145 \\
      & RMS\(\downarrow\) & - & - & - & \underline{3.861} & 4.053 & \textbf{3.726} & 4.514 \\
      & RMSlog\(\downarrow\) & - & - & - & \underline{0.225} & 0.239 & \textbf{0.221} & 0.243 \\
      & \(\delta^1\)\(\uparrow\) & - & - & - & \underline{0.839} & 0.831 & \textbf{0.849} & 0.803 \\
      & \(\delta^2\)\(\uparrow\) & - & - & - & \underline{0.934} & 0.929 & \textbf{0.937} & 0.921 \\
      & \(\delta^3\)\(\uparrow\) & - & - & - & \textbf{0.973} & \underline{0.969} & \textbf{0.973} & 0.968 \\
      \bottomrule
    \end{tabular}
  }
  \caption{Other error metrics on the MVSEC dataset (with the 100m cutoff).}\label{tab:results_mvsec_all_metrics}
\end{table}

Visual results are also presented in \cref{fig:cmp_mvsec}, comparing them with those of Cui \etal~\cite{Cui2022DenseDE} and of ALED\textsubscript{SL\(\rightarrow\)MV}~\cite{Brebion2023LearningTE}. While the method of Cui \etal{} allows for a good sharpness in the image, it fails at producing smooth depth gradients (especially on the ground, where the delimitations between LiDAR scans are clearly visible), and it is limited to the vertical range of the LiDAR sensor. Comparing our results to those of ALED\textsubscript{SL\(\rightarrow\)MV}, the visualizations reflect the observations made during the quantitative evaluation: our depth maps have a very good accuracy, but those of DELTA\textsubscript{SL\(\rightarrow\)MV} are slightly less sharp than those of ALED\textsubscript{SL\(\rightarrow\)MV}, while those of DELTA\textsubscript{MV} are much sharper but contain some small defects due to only learning on MVSEC (which contains erroneous ground truth values, as noted in \cref{sec:eval:datasets}). Like all other methods showcasing results on MVSEC~\cite{Brebion2023LearningTE,Gehrig2021CombiningEA,Hamaguchi2023HierarchicalNM}, the lack of ground truth data for the sky leads to blue blobs in the upper areas of all predictions.

\subsection{Results on the M3ED Dataset}
We propose two variants of DELTA on the M3ED dataset:
\begin{itemize}
  \item DELTA\textsubscript{M3}, only trained on M3ED;
  \item DELTA\textsubscript{SL\(\rightarrow\)M3}, after finetuning DELTA\textsubscript{SL} on M3ED.
\end{itemize}

\begin{figure*}
  \centering
  \setlength\tabcolsep{1pt}
  \renewcommand{\arraystretch}{0.5}
  \begin{tabular}{@{}ccccccc@{}}
    \small Events & \small LiDAR proj. & \small Ground truth & \small ALED\textsubscript{M3}~\cite{Brebion2023LearningTE} & \small ALED\textsubscript{SL\(\rightarrow\)M3}~\cite{Brebion2023LearningTE} & \small DELTA\textsubscript{M3} & \small DELTA\textsubscript{SL\(\rightarrow\)M3}\\
    \includegraphics[width=0.1375\linewidth]{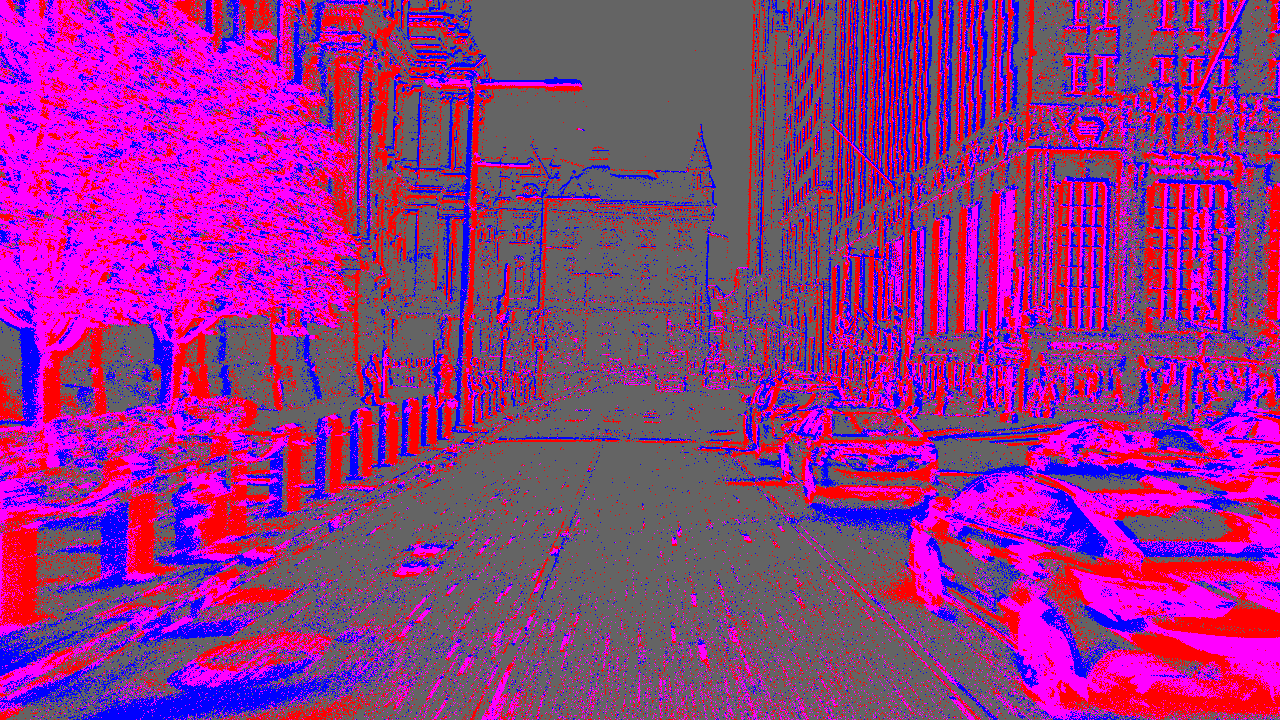} &
    \includegraphics[width=0.1375\linewidth]{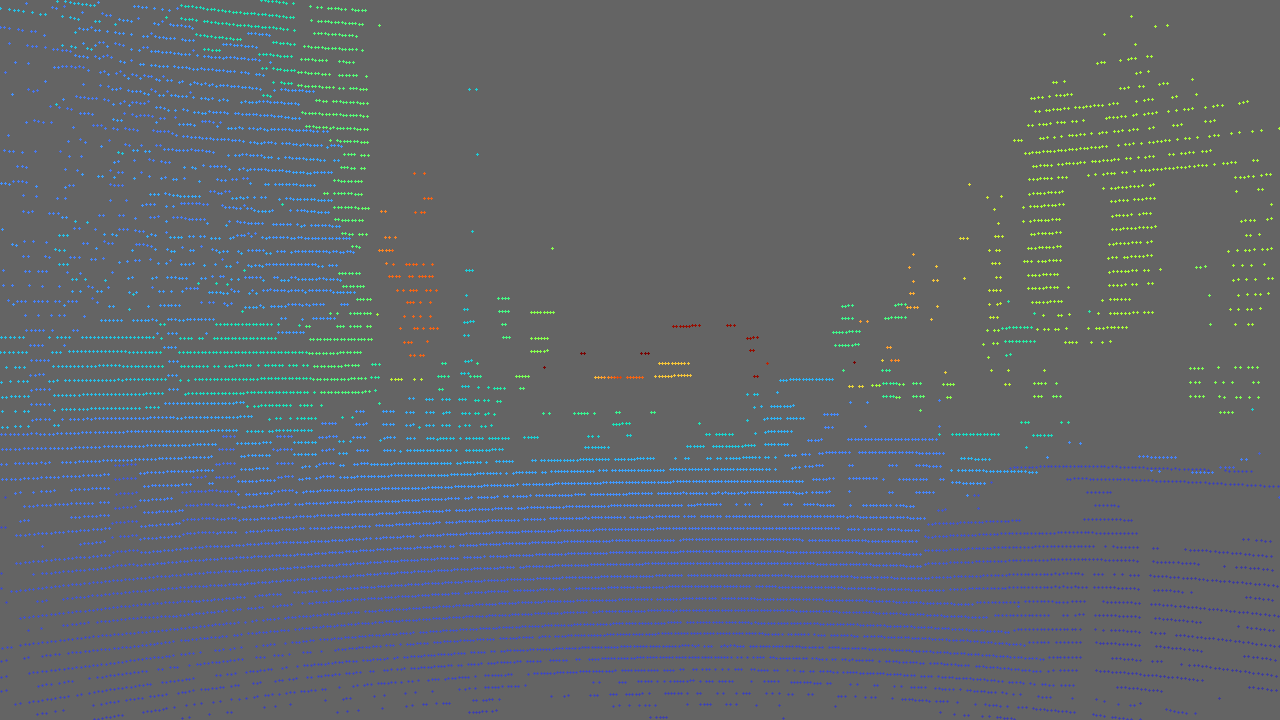} &
    \includegraphics[width=0.1375\linewidth]{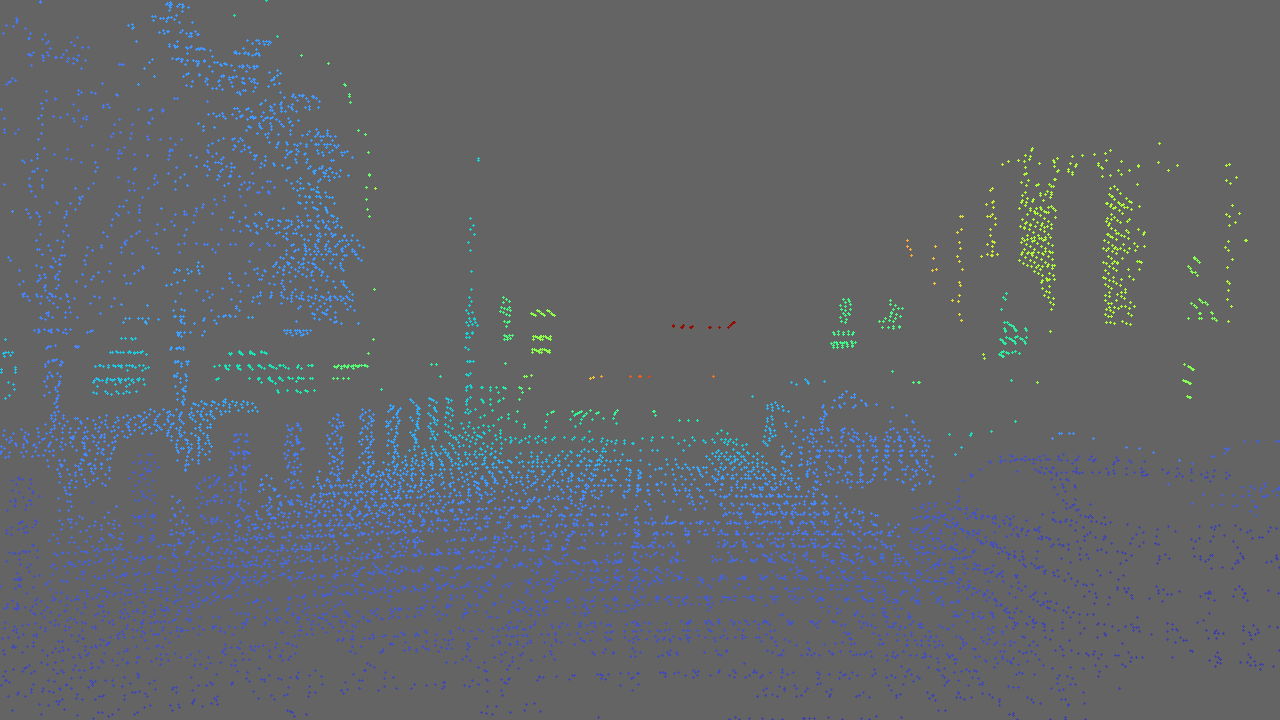} &
    \includegraphics[width=0.1375\linewidth]{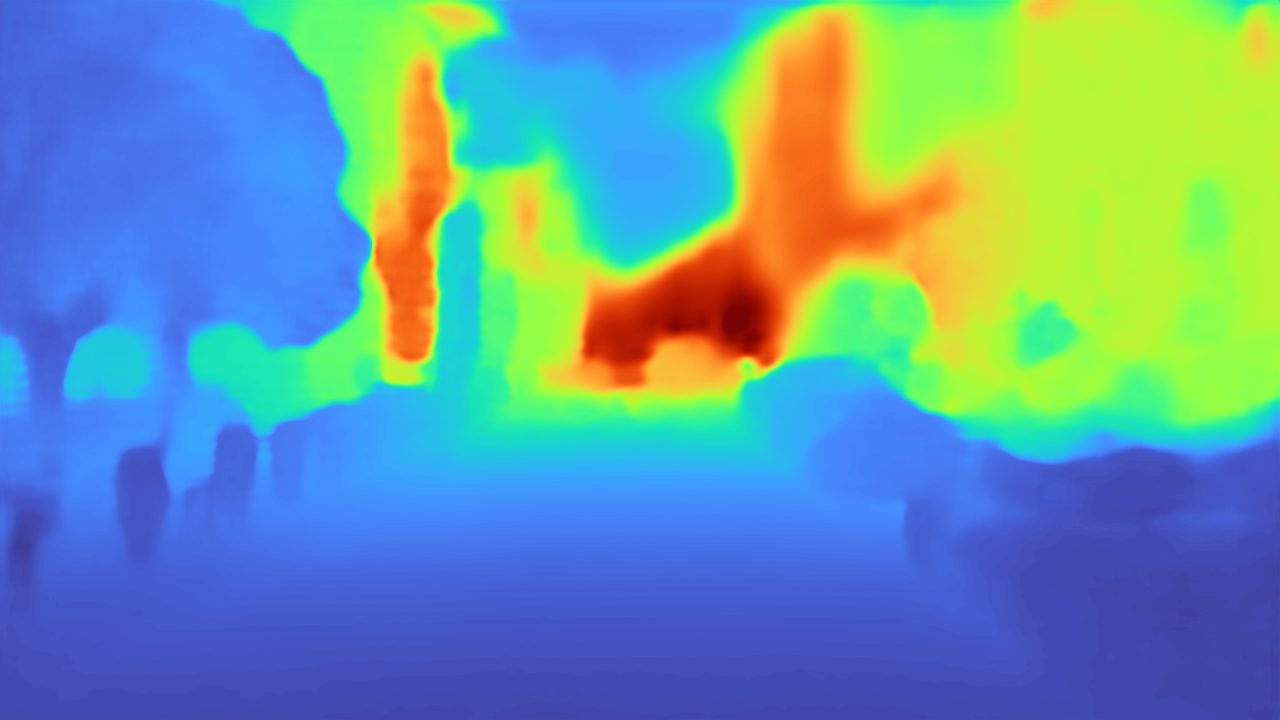} &
    \includegraphics[width=0.1375\linewidth]{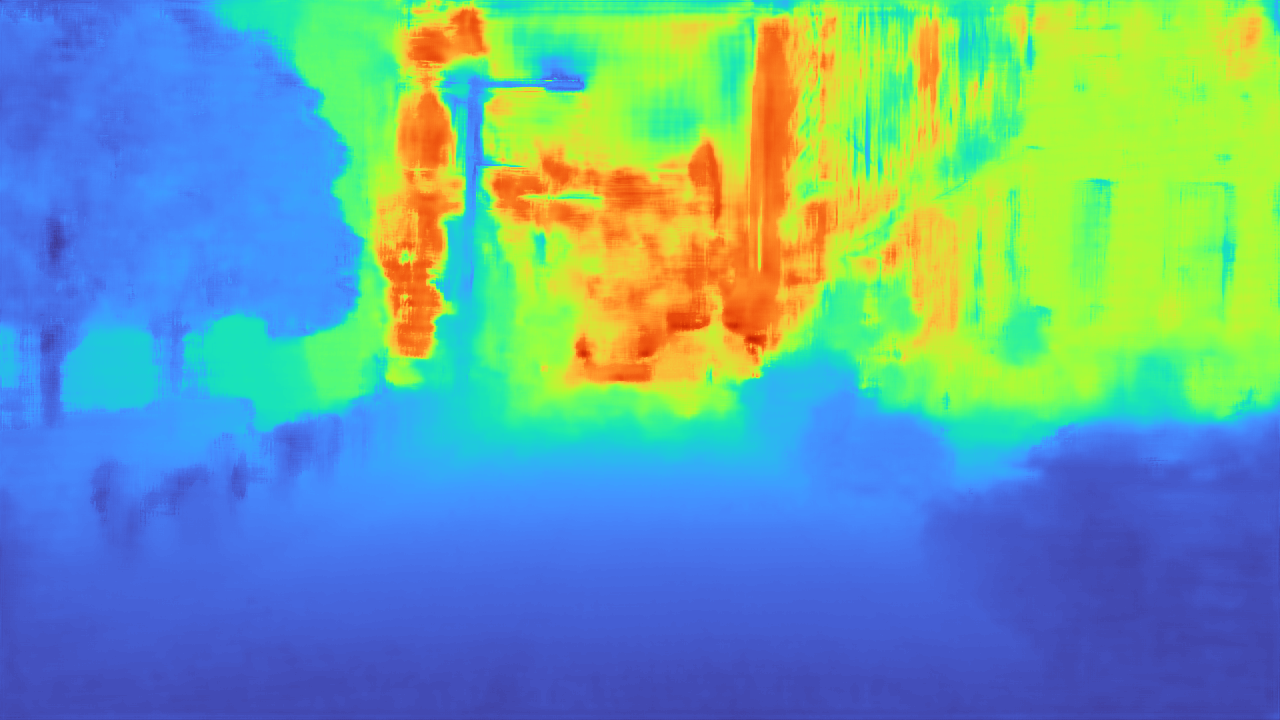} &
    \includegraphics[width=0.1375\linewidth]{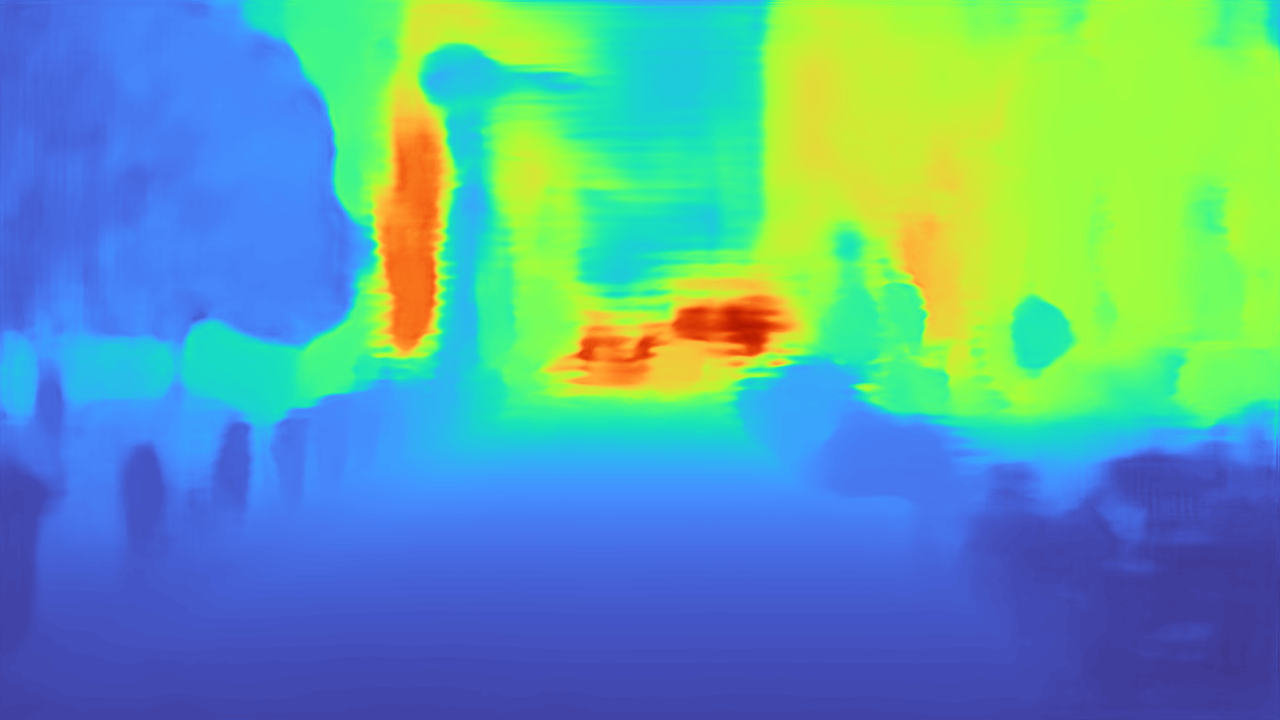} &
    \includegraphics[width=0.1375\linewidth]{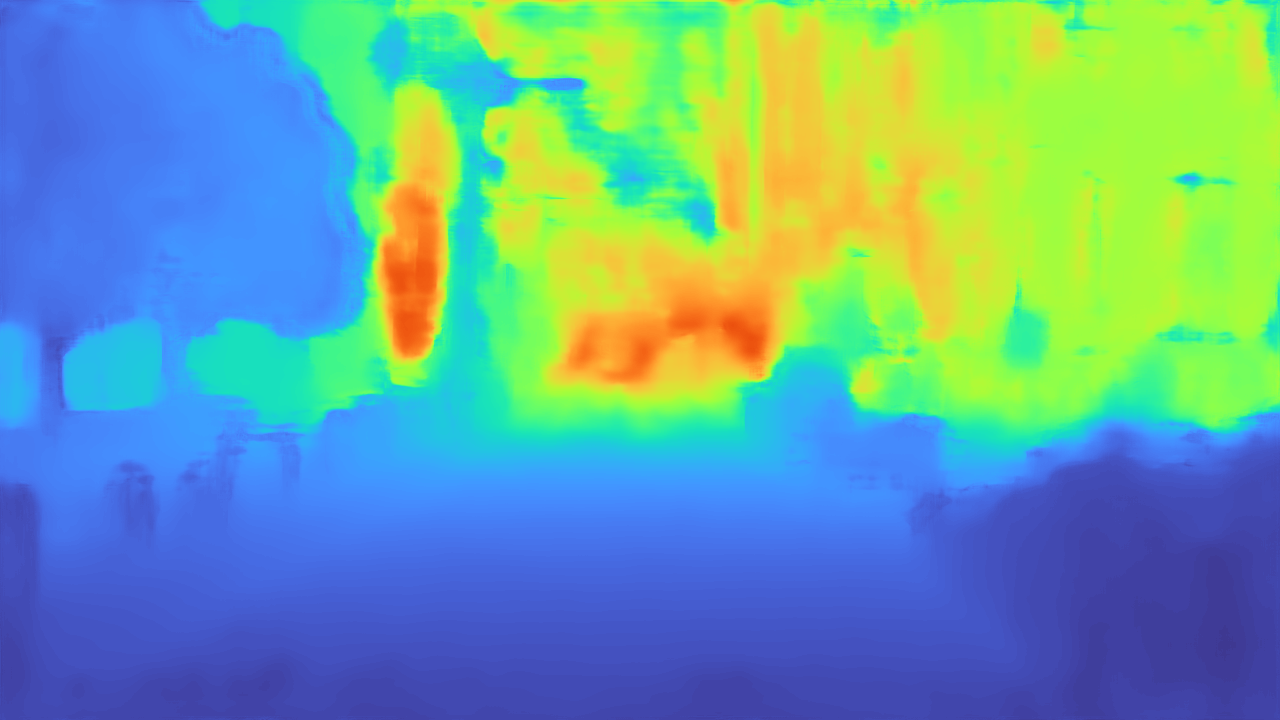}
  \end{tabular}
  \cprotect\caption{Qualitative results for the \verb|city_hall_day| sequence from M3ED. The size of points was increased for both the LiDAR projection and the ground truth. \textit{For a better visualization, an enlarged version of this figure is given in the Supplementary Material.}}\label{fig:cmp_m3ed}
\end{figure*}

Results are given in \cref{tab:results_m3ed}, compared with those of ALED~\cite{Brebion2023LearningTE}.
Again, DELTA performs very well on both sequences, with the mean error at the full cutoff distance being close to and lower than 1m for {\small\verb|city_hall_day|}, and even lower for {\small\verb|city_hall_night|}. This constitutes a significant improvement over ALED, as its mean error is always well over 1m for both recordings. Results on the other metrics also favor DELTA, although in a more even way. However, for both ALED and DELTA, pretraining on SLED brings little to no improvement. This apparent discrepancy can be explained through \cref{fig:cmp_m3ed}: the ground truth depth maps of M3ED are very sparse, with a similar density than the one of the LiDAR input, and with very few ground truth values at the edges of the objects. Because of these limitations, both ALED and DELTA showcase depth maps with accurate depth estimations, but give a blob-like appearance to all the objects as they could not learn the notion of edges from the training set. Pretraining on SLED allows for a better identification of these edges, producing sharper depth images, but it also leads to the appearance of artifacts (\eg, on the streetlamp on the left side in \cref{fig:cmp_m3ed}).

\begin{table}
  \centering
  \resizebox{\linewidth}{!}{
    \begin{tabular}{@{}ccccccccc@{}}
      \toprule
      \textbf{Recording} & \textbf{Method} & Mean\(\downarrow\) & AbsRel\(\downarrow\) & RMS\(\downarrow\) & RMSlog\(\downarrow\) & \(\delta^1\)\(\uparrow\) & \(\delta^2\)\(\uparrow\) & \(\delta^3\)\(\uparrow\) \\
      \midrule
      \multirow{4}{*}{\small\Verb|city_hall_day|} & ALED\textsubscript{M3}~\cite{Brebion2023LearningTE} & 1.41 & 0.074 & 2.867 & \underline{0.120} & 0.958 & 0.986 & \textbf{0.995} \\
      & \textbf{DELTA\textsubscript{M3}} & \underline{1.03} & \textbf{0.054} & 2.769 & \underline{0.120} & 0.953 & 0.985 & \underline{0.994} \\
      & ALED\textsubscript{SL\(\rightarrow\)M3}~\cite{Brebion2023LearningTE} & 1.31 & 0.090 & \underline{2.616} & 0.128 & \underline{0.962} & \underline{0.987} & \textbf{0.995} \\
      & \textbf{DELTA\textsubscript{SL\(\rightarrow\)M3}} & \textbf{0.89} & \underline{0.055} & \textbf{2.258} & \textbf{0.113} & \textbf{0.968} & \textbf{0.989} & \textbf{0.995} \\
      \midrule
      \multirow{4}{*}{\small\Verb|city_hall_night|} & ALED\textsubscript{M3}~\cite{Brebion2023LearningTE} & 1.41 & 0.068 & 2.499 & \underline{0.102} & \textbf{0.976} & \textbf{0.992} & \textbf{0.997} \\
      & \textbf{DELTA\textsubscript{M3}} & \textbf{0.82} & \textbf{0.042} & \underline{2.318} & \textbf{0.099} & 0.969 & \underline{0.990} & \underline{0.996} \\
      & ALED\textsubscript{SL\(\rightarrow\)M3}~\cite{Brebion2023LearningTE} & 1.28 & 0.085 & 2.398 & 0.117 & 0.971 & \underline{0.990} & \underline{0.996} \\
      & \textbf{DELTA\textsubscript{SL\(\rightarrow\)M3}} & \underline{0.85} & \underline{0.048} & \textbf{2.243} & 0.103 & \underline{0.974} & \underline{0.990} & 0.995 \\
      \bottomrule
    \end{tabular}
  }
  \caption{Errors of ALED~\cite{Brebion2023LearningTE} and DELTA on M3ED, without and with pretraining on SLED (with the maximum 120m cutoff). Results of ALED on M3ED were computed for this article.}\label{tab:results_m3ed}
\end{table}

Therefore, while M3ED remains a dataset of interest in the event-based community, we believe that it may not be the best suited for the training and/or evaluation of an event-and-LiDAR dense depth estimation method like ours.

\subsection{Ablation Studies}\label{sec:eval:ablation}
In this final subsection, we investigate the importance of the two memories in the network, the role of the central cross-attention module CA\textsubscript{F}, and the information provided by each of the input modalities. For that purpose, we propose five variants of DELTA:
\begin{itemize}
  \item DELTA\textsuperscript{NPM}, with no propagation memory;
  \item DELTA\textsuperscript{NCM}, with no central memory;
  \item DELTA\textsuperscript{NCA}, with no central cross-attention module (replaced by a GRU module for each encoding branch to update the central memory);
  \item DELTA\textsuperscript{NL}, with no LiDAR input;
  \item DELTA\textsuperscript{NE}, with no event input.
\end{itemize}
We give an overview of the structure of these modified networks in the Suppl.\ Material. For the evaluation, as was done for \cref{sec:eval:results_sled}, we trained these variants of DELTA on the SLED dataset, allowing for comparison with the proposed version of the network. The choice of doing this analysis on SLED was motivated by the fact that both the MVSEC and the M3ED datasets suffer from shortcomings in their respective ground truth, which could alter this ablation study.

\begin{table}
  \centering
  \setlength\tabcolsep{5pt}
  \resizebox{0.8\linewidth}{!}{
    \begin{tabular}{@{}ccc@{\hspace{0.5\tabcolsep}}lc@{\hspace{0.5\tabcolsep}}lc@{\hspace{0.5\tabcolsep}}l@{}}
      \toprule
      Cutoff & DELTA\textsubscript{SL} & \multicolumn{2}{c}{DELTA\rlap{\textsubscript{SL}}\textsuperscript{NPM}} & \multicolumn{2}{c}{DELTA\rlap{\textsubscript{SL}}\textsuperscript{NCM}} & \multicolumn{2}{c}{DELTA\rlap{\textsubscript{SL}}\textsuperscript{NCA}} \\
      \midrule
      10m & \textbf{0.60} & \underline{0.85} & \((+0.25)\) & 0.91 & \((+0.31)\) & 0.95 & \((+0.35)\) \\
      20m & \textbf{1.32} & \underline{1.64} & \((+0.32)\) & 1.71 & \((+0.39)\) & 1.80 & \((+0.48)\) \\
      30m & \textbf{1.92} & 2.24 & \((+0.32)\) & \underline{2.21} & \((+0.29)\) & 2.40 & \((+0.48)\) \\
      100m & \textbf{3.32} & 3.73 & \((+0.41)\) & \underline{3.53} & \((+0.21)\) & 3.85 & \((+0.53)\) \\
      200m & \underline{4.58} & 5.01 & \((+0.43)\) & \textbf{4.49} & \((-0.09)\) & 5.03 & \((+0.45)\) \\
      \bottomrule
    \end{tabular}
  }
  \caption{Absolute and relative mean depth errors (in meters) on the full testing set of SLED, for alternative versions of DELTA (No Propagation Memory, No Central Memory, No Cross-Attention).}\label{tab:results_ablation}
\end{table}

Results for the first three variants are presented in \cref{tab:results_ablation}. As can be seen, 
\begin{enumerate*}[label=\textbf{(\textcolor{cvprblue}{\arabic*})}]
  \item the base variant DELTA\textsubscript{SL} produces the best results, except for the maximum 200m cutoff range, where it is only slightly beaten by the DELTA\rlap{\textsubscript{SL}}\textsuperscript{NCM} variant.
  \item The version without a propagation memory DELTA\rlap{\textsubscript{SL}}\textsuperscript{NPM} has constantly an additional error of around 0.2m to 0.4m, highlighting the importance of temporally propagating the LiDAR data with the events.
  \item The version without the central memory DELTA\rlap{\textsubscript{SL}}\textsuperscript{NCM} also performs worse, albeit with an additional error that diminishes the greater the cutoff distance is, beating DELTA\textsubscript{SL} by 0.09m at the maximum cutoff range. It should be noted however that the car which the sensors are mounted on in the SLED dataset rarely stops, and if it does, it is for very short periods of time. As explained in \cref{sec:method:memory}, since the main role of the central memory is to help provide accurate results in these cases, it only serves here its secondary role of being a stabilization medium, which is still valuable given the numerical results.
  \item The version without the central cross-attention module DELTA\rlap{\textsubscript{SL}}\textsuperscript{NCA} is the worst performing variant, as it has the largest error across all cutoff ranges. As such, the central cross-attention module CA\textsubscript{F} is crucial for encoding the cross-relations between the encoded LiDAR and event data before updating the central memory.
\end{enumerate*}

\begin{figure}
  \centering
  \setlength\tabcolsep{1pt}
  \renewcommand{\arraystretch}{0.5}
  \begin{tabular}{@{}ccc@{}}
    \small Ground truth & \hphantom{b} & \small DELTA\textsubscript{SL} \\
    \includegraphics[width=0.4\linewidth]{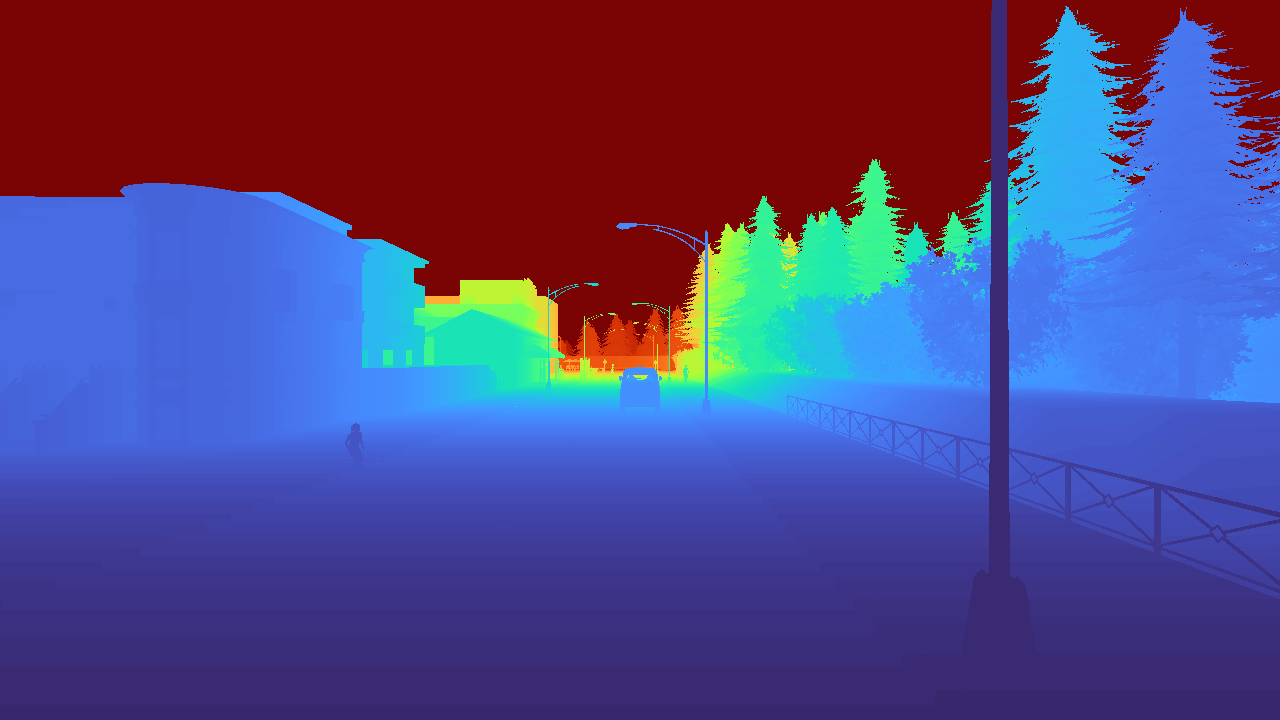} & &
    \includegraphics[width=0.4\linewidth]{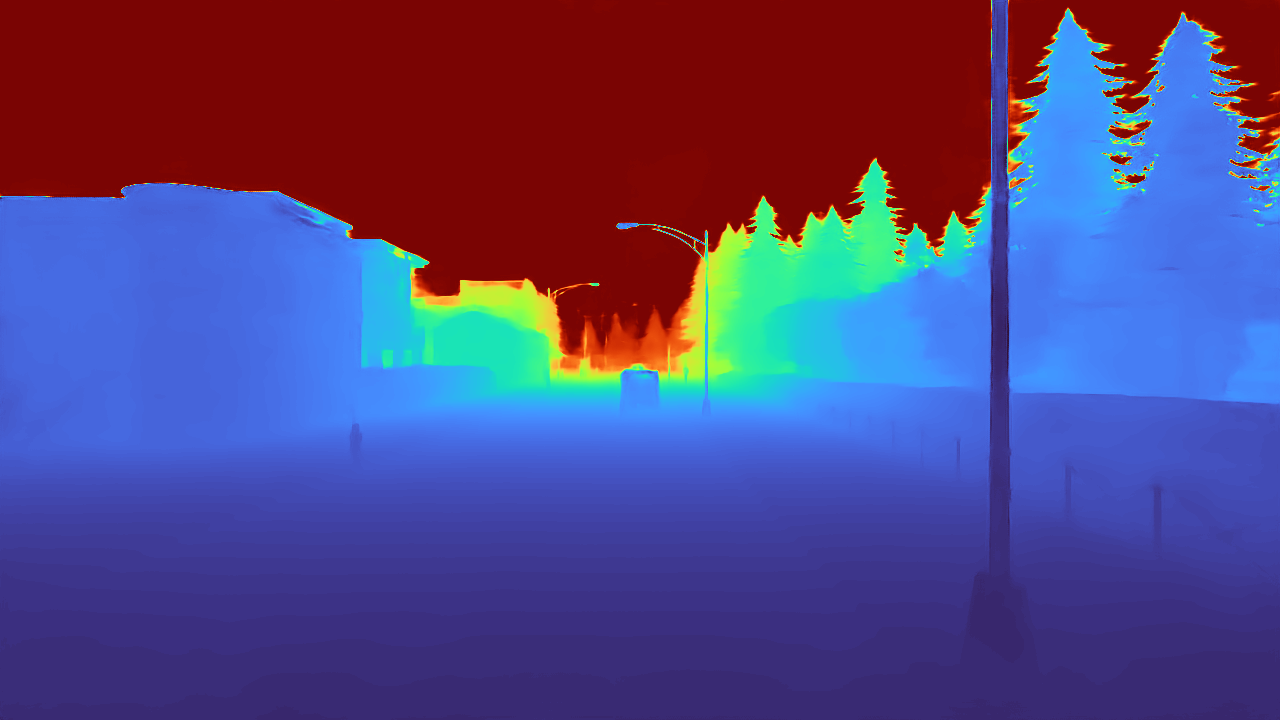} \\
    \small DELTA\rlap{\textsubscript{SL}}\textsuperscript{NL} & & \small DELTA\rlap{\textsubscript{SL}}\textsuperscript{NE} \\
    \includegraphics[width=0.4\linewidth]{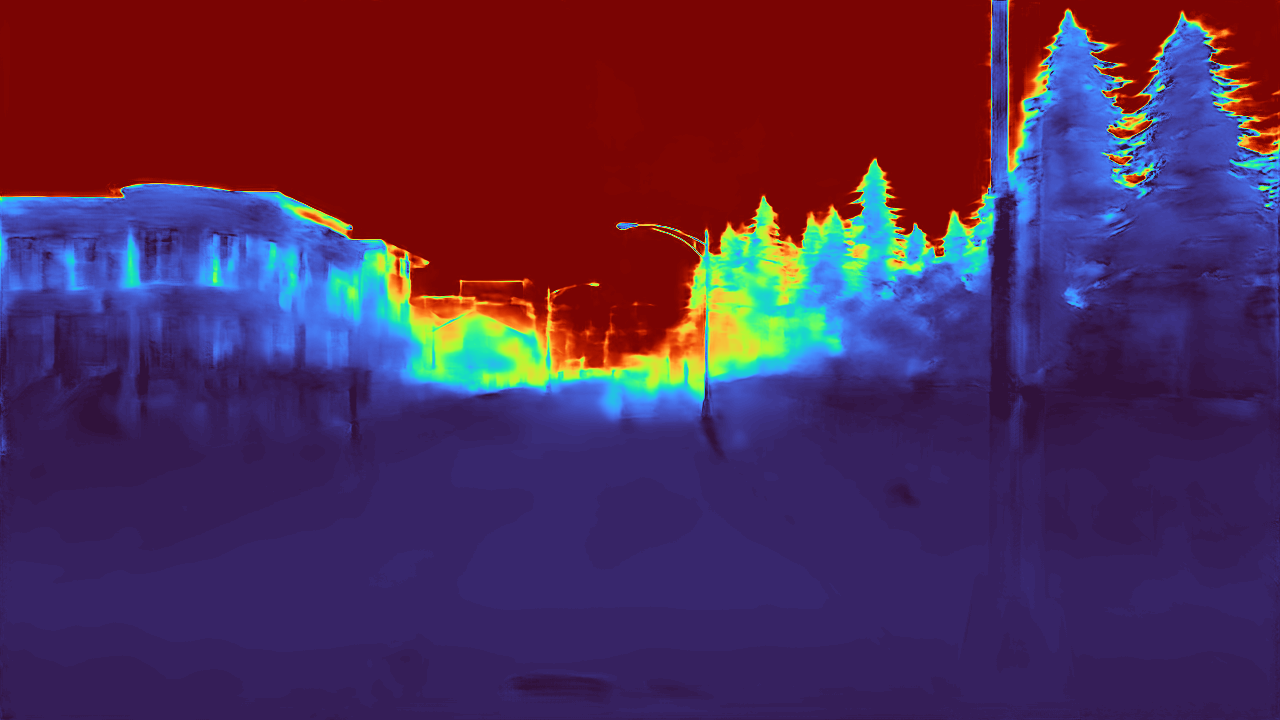} & &
    \includegraphics[width=0.4\linewidth]{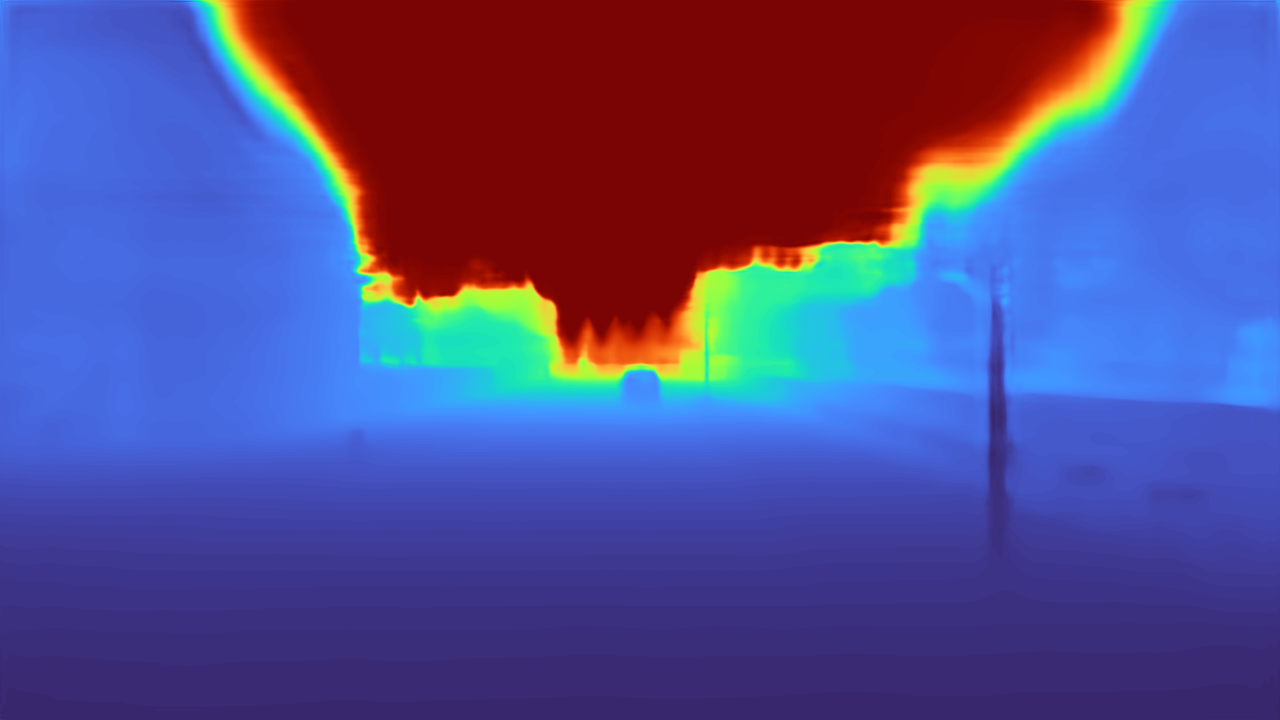} \\
  \end{tabular}
  \cprotect\caption{Qualitative results for the DELTA\textsuperscript{NL} and DELTA\textsuperscript{NE} variants, on the \verb|Town01_19| sequence of SLED.}\label{fig:results_ablation_nl_ne}
\end{figure}

Finally, results for the DELTA\textsuperscript{NL} and DELTA\textsuperscript{NE} variants are given in \cref{fig:results_ablation_nl_ne}. They highlight the complementarity of the two inputs: the events help identify the edges of the objects but not their depth (DELTA\textsuperscript{NL}), while the LiDAR helps identify the depth but not the edges of the objects, especially those above or under its narrow vertical FOV (DELTA\textsuperscript{NE}).

\section{Conclusion and Discussions}
In this article, a new attention-based network for fusing projected LiDAR data and temporal windows of events to construct dense depth maps was presented. Thanks to the introduction of a propagation memory between cross-attentions, DELTA is able to extrapolate LiDAR with events at higher rate for an optimal fusion.
A GRU is also added between the encoding and decoding stages, allowing for a short-term central memory and more robust outputs. A thorough evaluation including ablation studies was conducted on three datasets of the state of the art to demonstrate the relevance of these propositions. On the synthetic SLED dataset, large improvements are achieved, with for instance the mean error being reduced up to four times for short ranges when compared to the state of the art. On MVSEC and M3ED, despite limitations in the ground truth data of these datasets, DELTA remains very competitive, being close to or achieving the state of the art, with low errors across all cutoff ranges. We believe in particular that our new state-of-the-art architecture can serve as a basis for other applications or for fusion with other sensors.

In hindsight, some modifications could still be brought to this work in order to further improve its overall performance.
\begin{enumerate*}[label=\textbf{(\textcolor{cvprblue}{\arabic*})}]
  \item As shown throughout this article, the reduction of errors over short ranges is sometimes done at the cost of lower precision at longer ranges. Adapting our training procedure by introducing variable weights for every cutoff distance could be a solution to make sure all of them are optimized equally.
  \item We chose to use the event volume~\cite{Zhu2019UnsupervisedEL}, as it is a standard, compact, and high-performing representation of event data. While it allows DELTA to achieve very accurate results, this representation can lead to some information being lost in case of very fast motion. Novel representations~\cite{Baldwin2021TimeOrderedRE,Gehrig2019EndtoEndLO,Zubic2023FromCC} could be examined for these cases.
  \item Finally, projecting the LiDAR data allows for a pixel-wise fusion with the event data, but makes the three-dimensional structure of the point cloud disappear. Working directly in 3D like~\cite{Cui2022DenseDE} could therefore constitute an interesting research opportunity.
\end{enumerate*}

\section*{Acknowledgment}
This work was supported in part by the Hauts-de-France Region and in part by the SIVALab Joint Laboratory (Renault Group -- Universit\'e de technologie de Compi\`egne (UTC) -- Centre National de la Recherche Scientifique (CNRS)).
This work was also supported by The Royal Physiographic Society in Lund.

{
    \small
    \bibliographystyle{ieeenat_fullname}
    \bibliography{main}
}

\clearpage
\setcounter{page}{1}
\maketitlesupplementary

\section{Enlarged Views of the Results on SLED, MVSEC, and M3ED}
As described in the main article, an enlarged version of \cref{fig:cmp_sled} is given in \cref{fig:cmp_sled_enlarged} (this version also includes the input LiDAR and event data), an enlarged version of \cref{fig:cmp_mvsec} is given in \cref{fig:cmp_mvsec_enlarged}, and an enlarged version of \cref{fig:cmp_m3ed} is given in \cref{fig:cmp_m3ed_enlarged}.

\section{Alternative Versions of DELTA}
\begin{figure*}
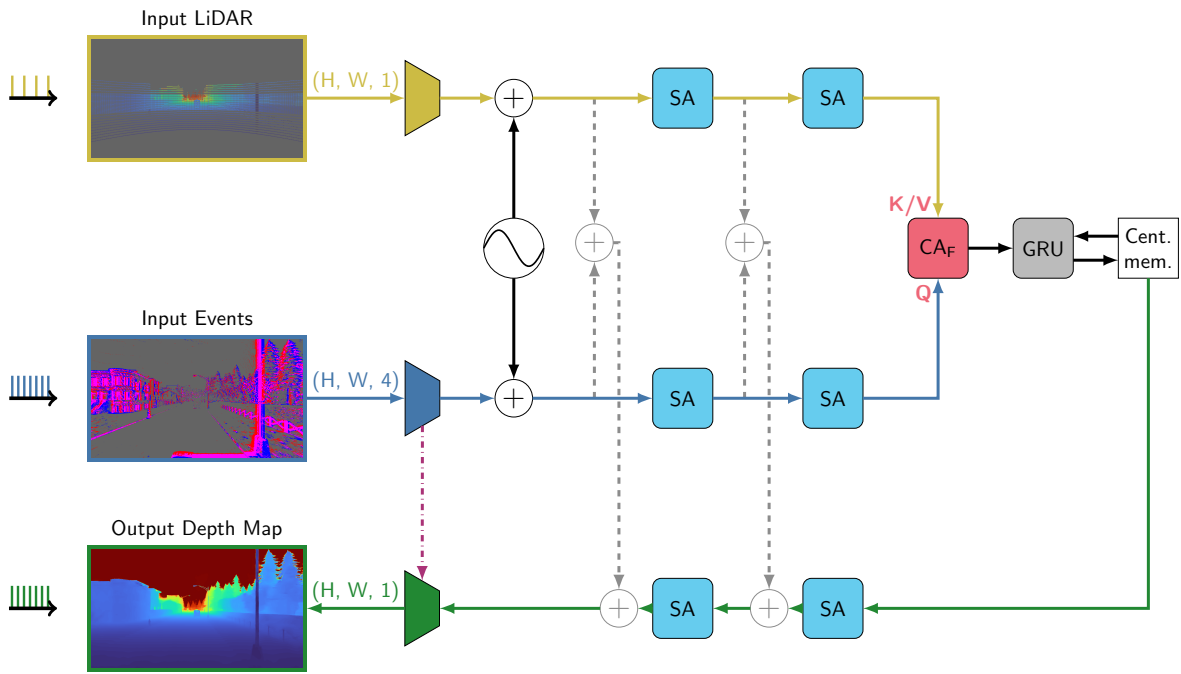

  \centering
  \resizebox{0.9\linewidth}{!}{
    %
  }

  \caption{The alternative architecture without propagation memory, DELTA\textsuperscript{NPM}.}\label{fig:network_alt_npm}
  \vspace{2cm}
\end{figure*}

\begin{figure*}
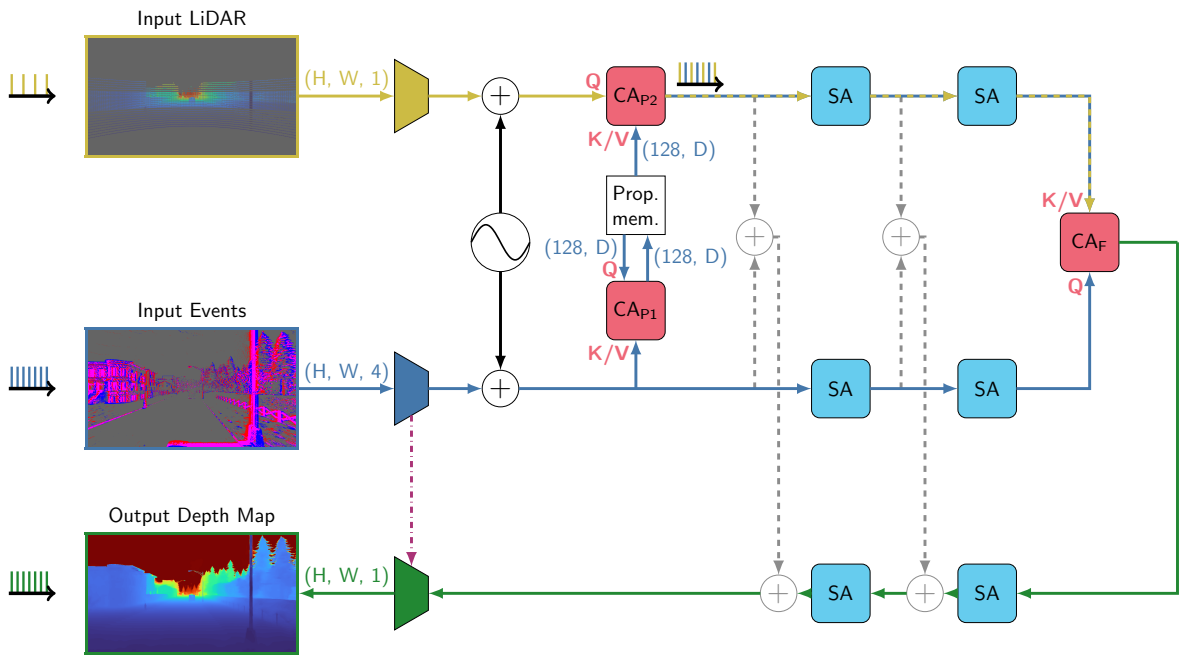

  \centering
  \resizebox{0.9\linewidth}{!}{
    %
  }

  \caption{The alternative architecture without central memory, DELTA\textsuperscript{NCM}.}\label{fig:network_alt_ncm}
\end{figure*}

\begin{figure*}
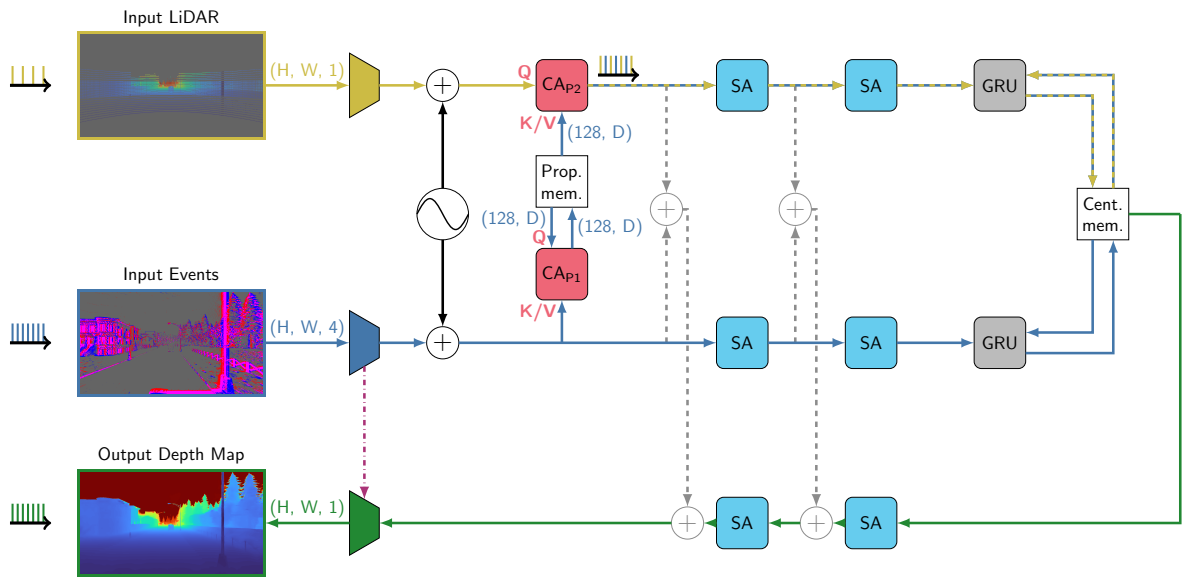

  \centering
  \resizebox{0.9\linewidth}{!}{
    %
  }

  \caption{The alternative architecture without the central cross-attention, DELTA\textsuperscript{NCA}.}\label{fig:network_alt_nca}
\end{figure*}

\begin{figure*}
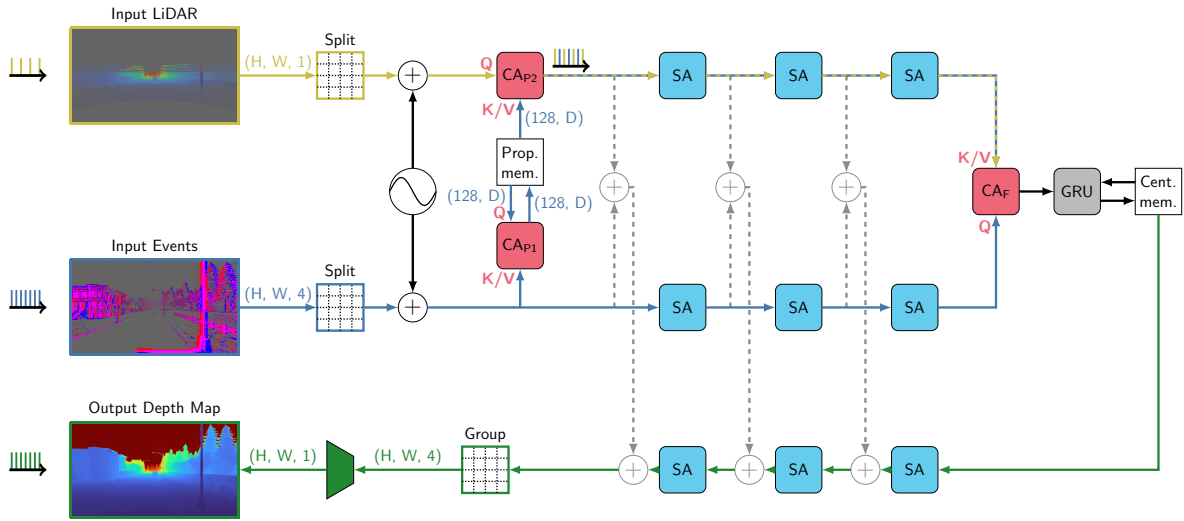

  \centering
  \resizebox{0.9\linewidth}{!}{
    %
  }

  \caption{The alternative architecture where the convolutional encoding heads are replaced by a simple splitting into patches, DELTA\textsuperscript{NEH}.}\label{fig:network_alt_neh}
  \vspace{1cm}
\end{figure*}

Alternative versions of our DELTA network are given in \cref{fig:network_alt_npm,fig:network_alt_ncm,fig:network_alt_nca,fig:network_alt_nl,fig:network_alt_ne,fig:network_alt_neh}. Versions of the network illustrated in \cref{fig:network_alt_npm,fig:network_alt_ncm,fig:network_alt_nca,fig:network_alt_nl,fig:network_alt_ne} are used as part of the ablation study in \cref{sec:eval:ablation} of the main article, while the version illustrated in \cref{fig:network_alt_neh} is used as part of \cref{sec:ablation_study} of this Supplementary Material.

\section{Ablation Study on Encoding Heads}\label{sec:ablation_study}
In addition to the ablation studies conducted in the main article, we propose here an additional variant of the network, DELTA\textsuperscript{NEH}, showcased in \cref{fig:network_alt_neh}. Here, the convolutional encoding heads are replaced by a more direct splitting into patches, as originally done in the Vision Transformer~\cite{Dosovitskiy2020AnII}. To compensate for the reduced number of parameters in the network, we add a third layer of self-attention modules. At the end of the decoding, the patches are simply grouped back to an image-based format, and a final small convolutional head reduces the number of channels and smooths the resulting depth maps.

Results of DELTA\textsuperscript{NEH} on the SLED dataset are shown in \cref{tab:results_ablation_suppl}. As can be seen, DELTA\rlap{\textsubscript{SL}}\textsuperscript{NEH} does not perform well at all, even worse than all the variants showcased in the main article. As visually illustrated in \cref{fig:cmp_neh}, DELTA\rlap{\textsubscript{SL}}\textsuperscript{NEH} produces depth maps with large errors (especially for the tunnel in the bottom row), and where the junction between the patches remains visible, creating numerous artifacts. While a simple splitting into patches can be conducted for the Vision Transformer, as classification is the end task, here we require a dense reconstruction at the end, \ie, we need to keep information about the structure of the scene. Therefore, in our case, computing the patches using a convolutional head allows for a better re-grouping of the patches at the end of the network, by allowing the decoding head to be guided by the corresponding data from the encoding head through our use of the convex upsampling module of~\cite{Teed2020RAFTRA}.

\begin{table}
  \centering
  \setlength\tabcolsep{5pt}
  \resizebox{0.65\linewidth}{!}{
    \begin{tabular}{@{}cccc@{\hspace{0.5\tabcolsep}}l@{}}
      \toprule
      Map & Cutoff & DELTA\textsubscript{SL} & \multicolumn{2}{c}{DELTA\rlap{\textsubscript{SL}}\textsuperscript{NEH}} \\
      \midrule
      \multirow{5}{*}{\Verb|Town01|} & 10m & \textbf{0.66} & 1.58 & \((+0.92)\) \\
      & 20m & \textbf{1.33} & 2.89 & \((+1.56)\) \\
      & 30m & \textbf{1.91} & 3.90 & \((+1.99)\) \\
      & 100m & \textbf{3.22} & 6.73 & \((+3.51)\) \\
      & 200m & \textbf{4.54} & 9.85 & \((+5.41)\) \\
      \midrule
      \multirow{5}{*}{\Verb|Town03|} & 10m & \textbf{0.54} & 2.13 & \((+1.59)\) \\
      & 20m & \textbf{1.31} & 3.17 & \((+1.86)\) \\
      & 30m & \textbf{1.93} & 3.89 & \((+1.96)\) \\
      & 100m & \textbf{3.40} & 6.14 & \((+2.74)\) \\
      & 200m & \textbf{4.63} & 9.23 & \((+4.60)\) \\
      \bottomrule
    \end{tabular}
  }
  \caption{Absolute and relative mean depth errors (in meters) on the SLED dataset, for the base version of DELTA and the ``No Encoding Head'' variant shown in \cref{fig:network_alt_neh}.}\label{tab:results_ablation_suppl}
\end{table}

\section{Computational Complexity}
We report in \cref{tab:time_mem} several metrics of the computational complexity of DELTA, computed on a single NVIDIA L40 GPU. For high- (1280\texttimes720), mid- (640\texttimes480), and low-resolution (346\texttimes260) data, DELTA has a mean inference rate of 6.3Hz, 20.5Hz, and 47.8Hz respectively. Compared to the method of Cui \etal~\cite{Cui2022DenseDE} with its reported output rate of 56Hz on the MVSEC dataset, our method is only 1.17 times slower, but for a much better accuracy as shown in \cref{tab:results_mvsec_mean} of the main article. Compared to ALED, despite the significant increase in the number of parameters due to the use of attention modules, DELTA requires a similar amount of FLOPS and of GPU memory, allowing its deployment on standard consumer-grade GPUs. Inference times of ALED are of course smaller, and while we can not exactly call our method real-time, we want to remind the reader here that the focus of our work was set on accuracy rather on real-time compatibility. As such, inference time and/or memory usage could be further reduced, as we are not using advanced optimizations like \verb|torch.compile()|, and as we believe the DELTA architecture could be slightly revised to reduce its number of parameters while keeping a similar accuracy. Implementation on specialized hardware could also be considered for real-time robotic applications, but is beyond the scope of this work.
\begin{table*}
  \centering
  \setlength\tabcolsep{5pt}
  \resizebox{0.85\linewidth}{!}{
    \begin{tabular}{@{}lllcrrrr@{}}
      \toprule
      Model & Resolution (with padding) & Dataset(s) & Patch size & Nb.\ param.\ & FLOPS & Inference time & Max.\ GPU mem.\ \\
      \midrule
      \multirow{3}{*}{DELTA} & 1280 \texttimes{} 720 & SLED, M3ED & 16 & 180.9M & 1786.1B & 157.9ms \rpm{} 2.0ms & 10.64 GB \\
      & 640 \texttimes{} 480 & DSEC & 16 & 180.9M & 596.4B & 48.7ms \rpm{} 0.4ms & 4.68 GB \\
      & 346 \texttimes{} 260 (348 \texttimes{} 264) & MVSEC & 12 & 181.2M & 300.4B & 20.9ms \rpm{} 0.2ms & 3.09 GB \\
      \midrule
      \multirow{3}{*}{ALED} & 1280 \texttimes{} 720 & SLED, M3ED & / & 26.2M & 1546.0B & 91.2ms \rpm{} 16.8ms & 7.02 GB \\
      & 640 \texttimes{} 480 & DSEC & / & 26.2M & 515.3B & 26.9ms \rpm{} 5.0ms & 2.74 GB \\
      & 346 \texttimes{} 260 (352 \texttimes{} 264) & MVSEC & / & 26.2M & 155.9B & 7.5ms \rpm{} 1.4ms & 1.36 GB \\
      \bottomrule
    \end{tabular}
  }
  \caption{Computational complexity metrics (number of parameters, FLOPS, mean inference time, maximum GPU memory usage) for DELTA and ALED, for both high-, mid-, and low-resolution data.}\label{tab:time_mem}
\end{table*}

\section{Additional Visual Results on the SLED Dataset}
\begin{figure*}
  \centering
  \includegraphics[width=0.475\linewidth]{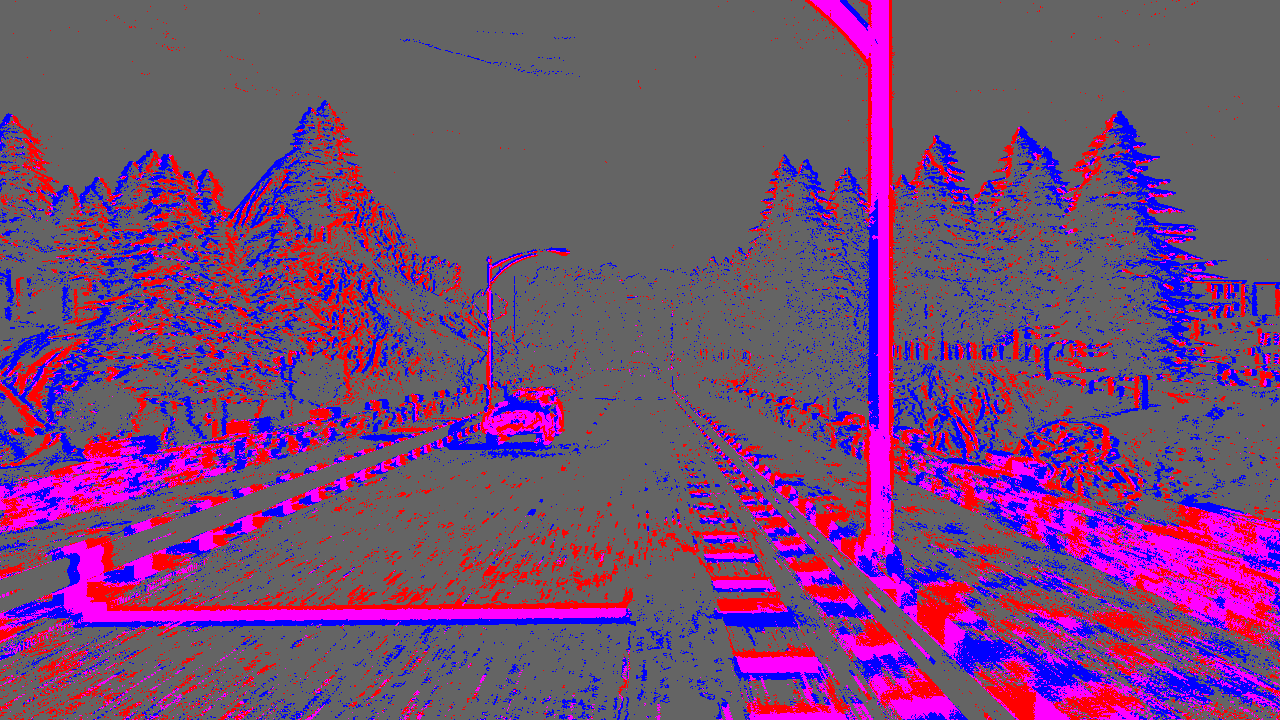}
  \includegraphics[width=0.475\linewidth]{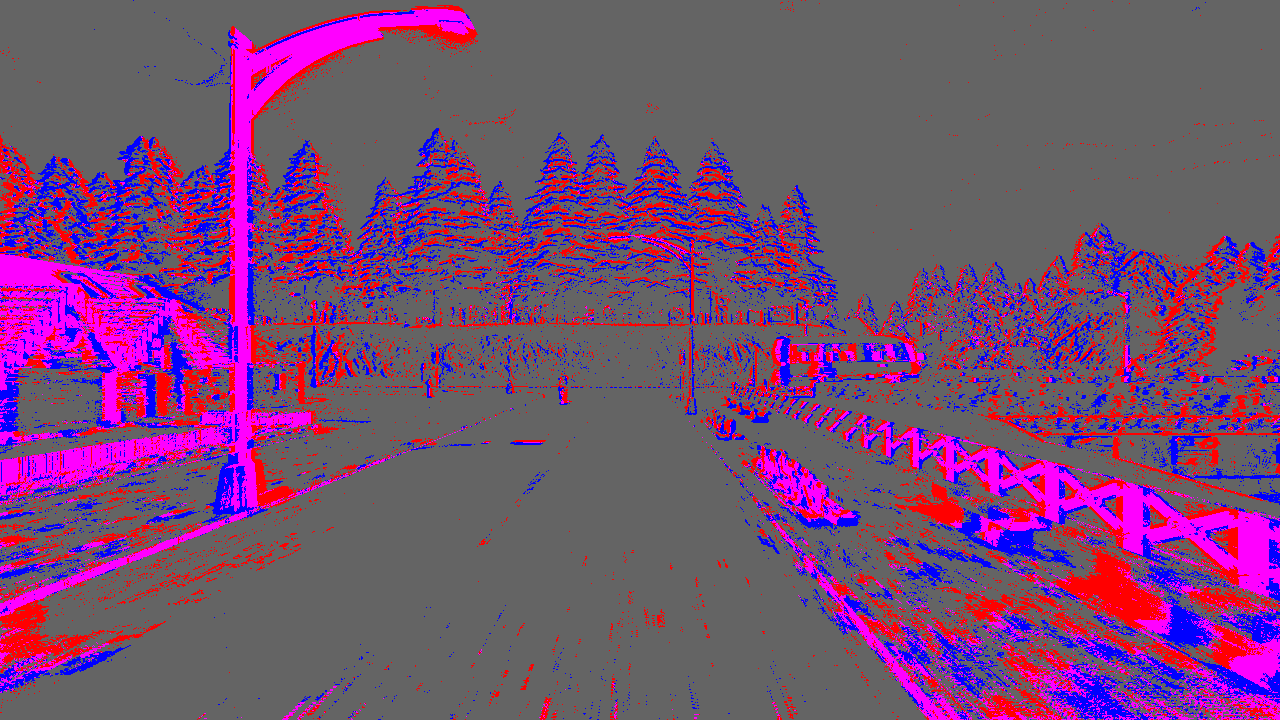}\\
  \includegraphics[width=0.475\linewidth]{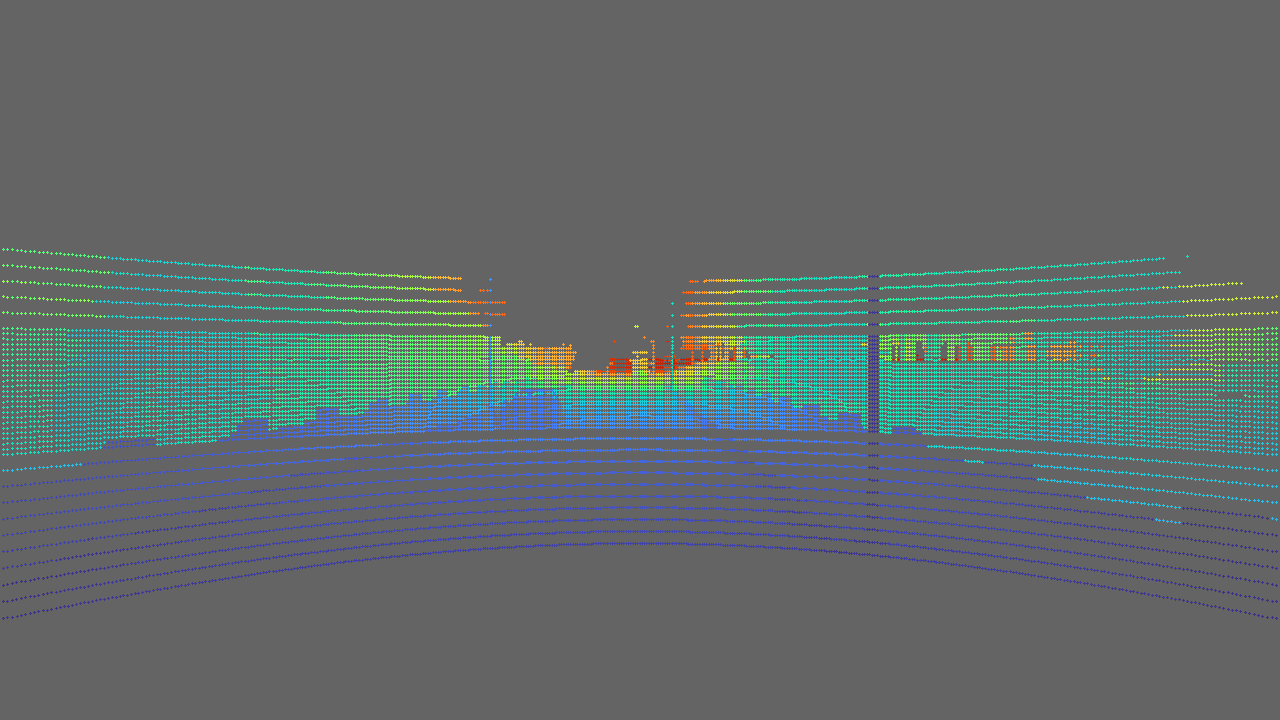}
  \includegraphics[width=0.475\linewidth]{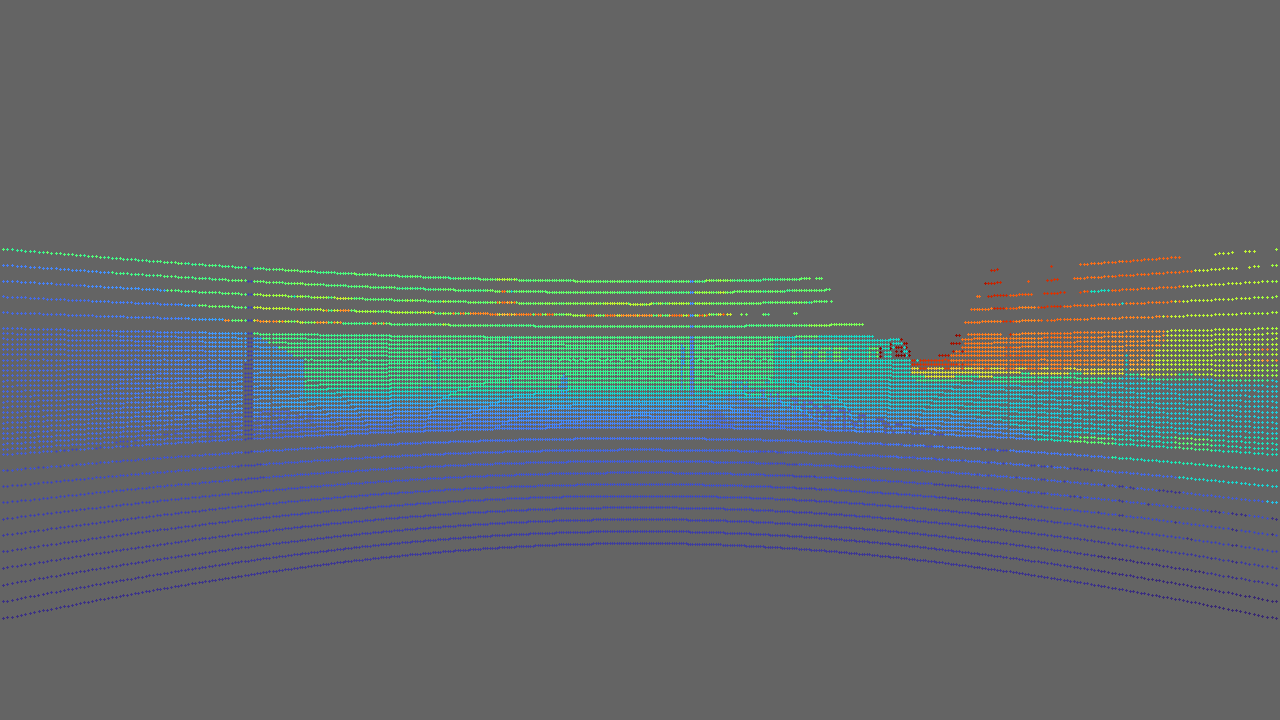}\\
  \includegraphics[width=0.475\linewidth]{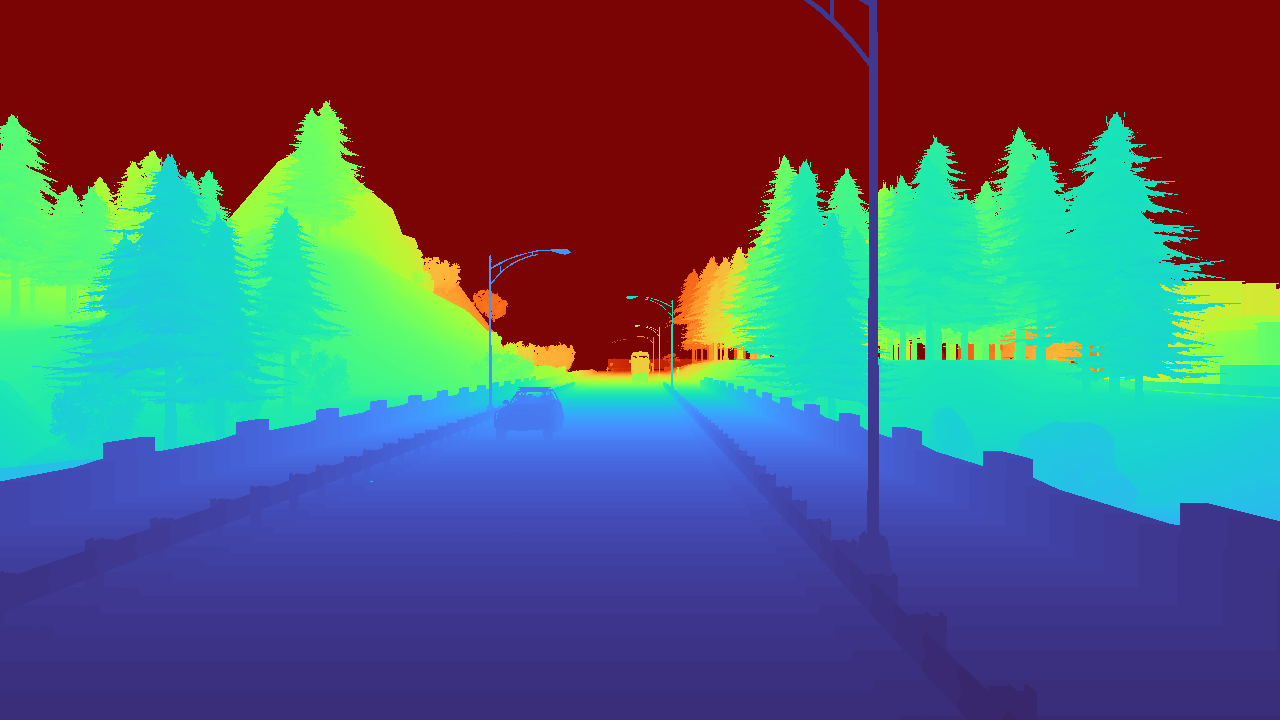}
  \includegraphics[width=0.475\linewidth]{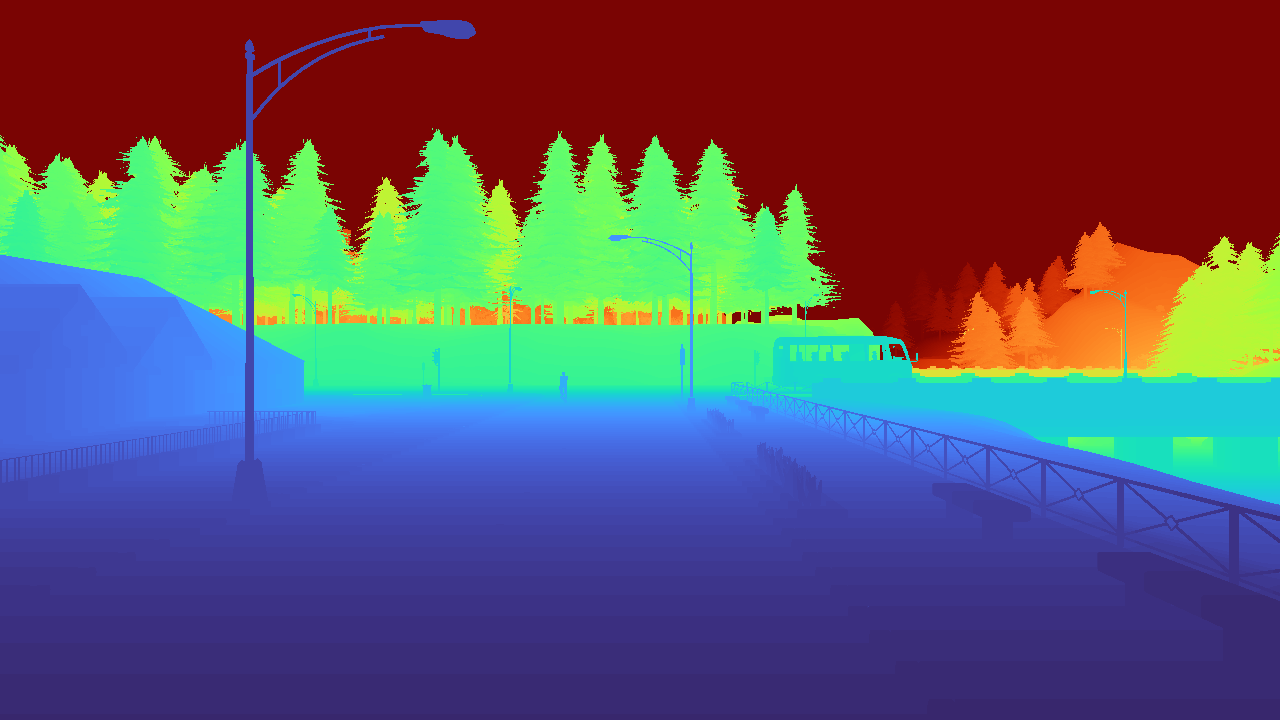}\\
  \includegraphics[width=0.475\linewidth]{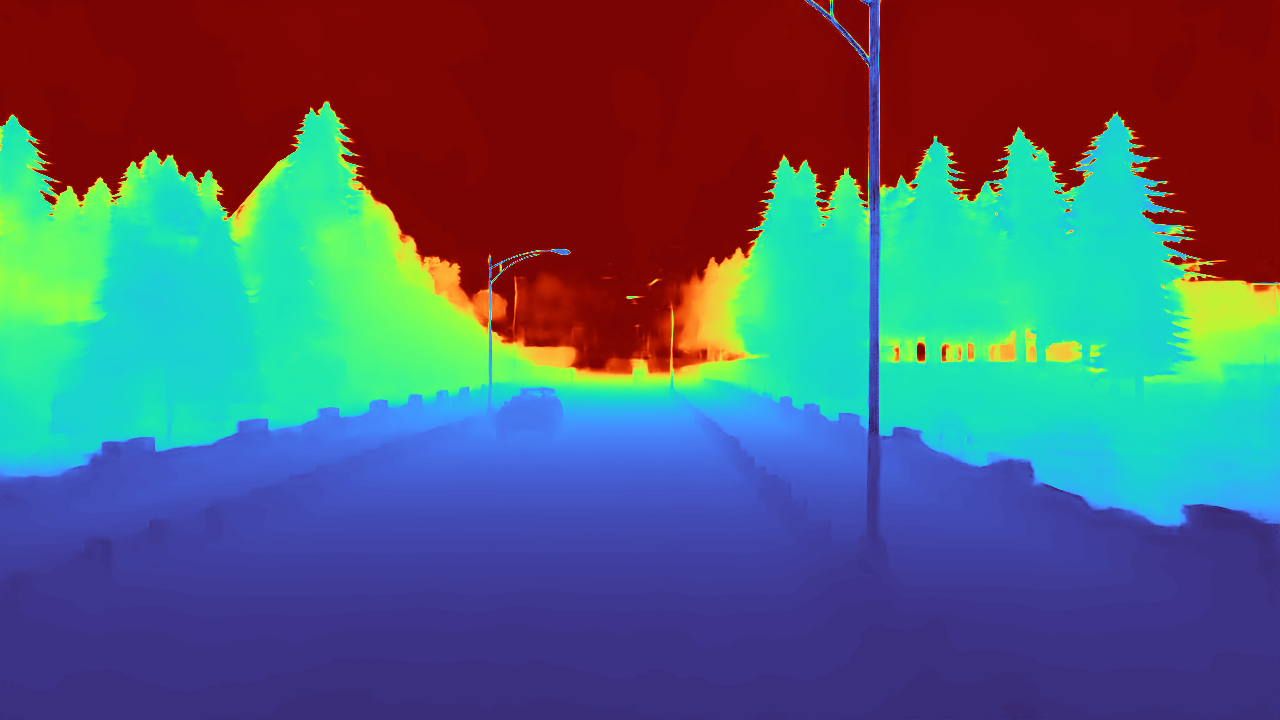}
  \includegraphics[width=0.475\linewidth]{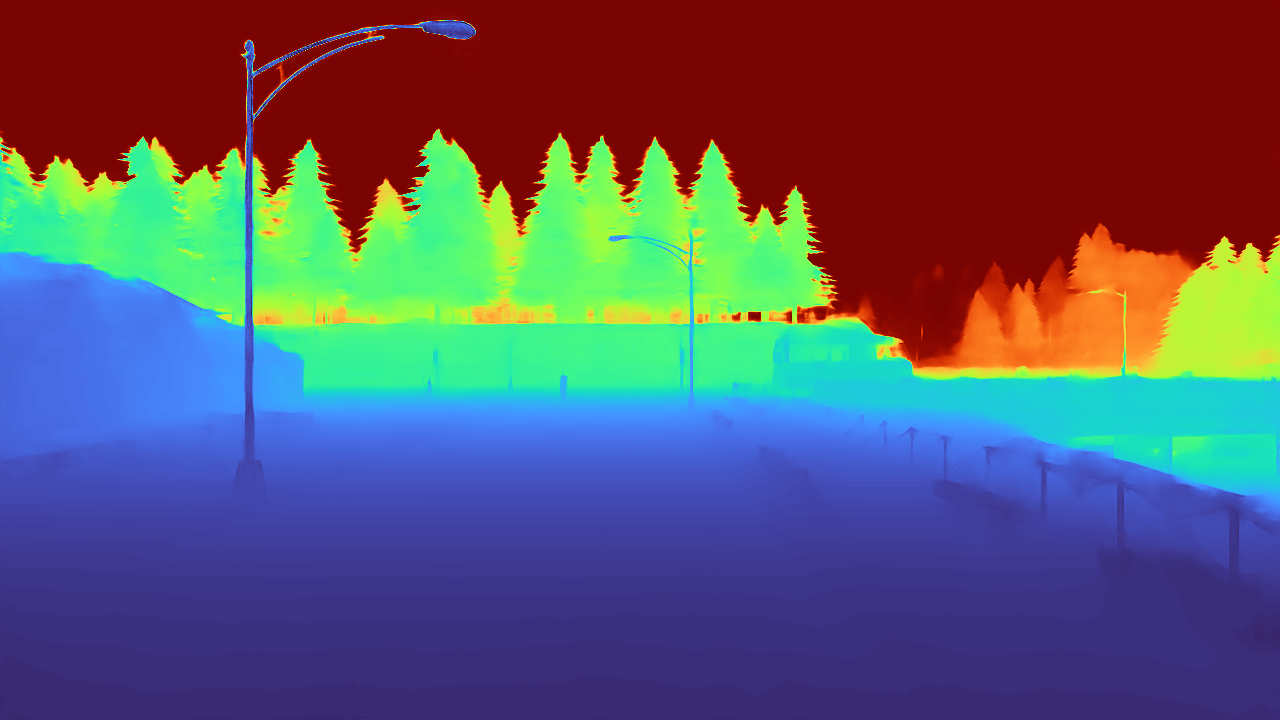}
  \cprotect\caption{Additional results on the SLED dataset, on sequences \verb|Town01_03| and \verb|Town01_05|. From top to bottom: events, LiDAR projection, ground truth, our results.}\label{fig:cmp_sled_additional_good_0}
\end{figure*}

\begin{figure*}
  \centering
  \includegraphics[width=0.475\linewidth]{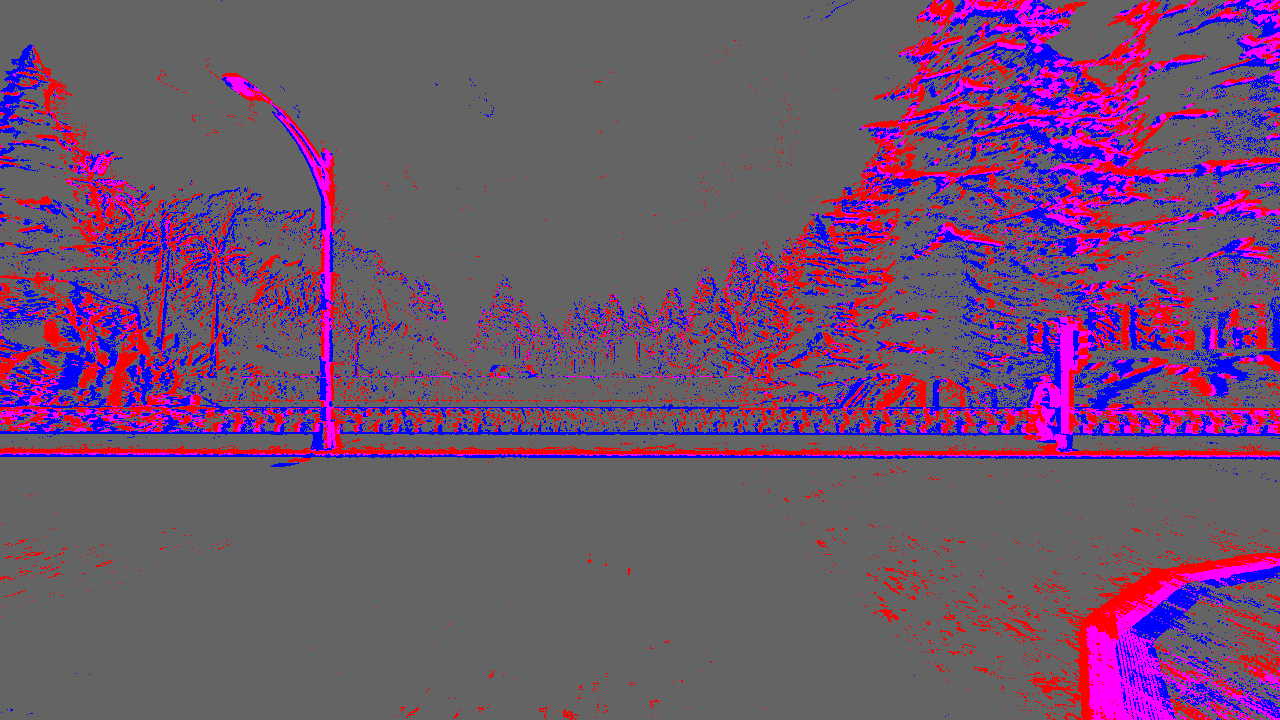}
  \includegraphics[width=0.475\linewidth]{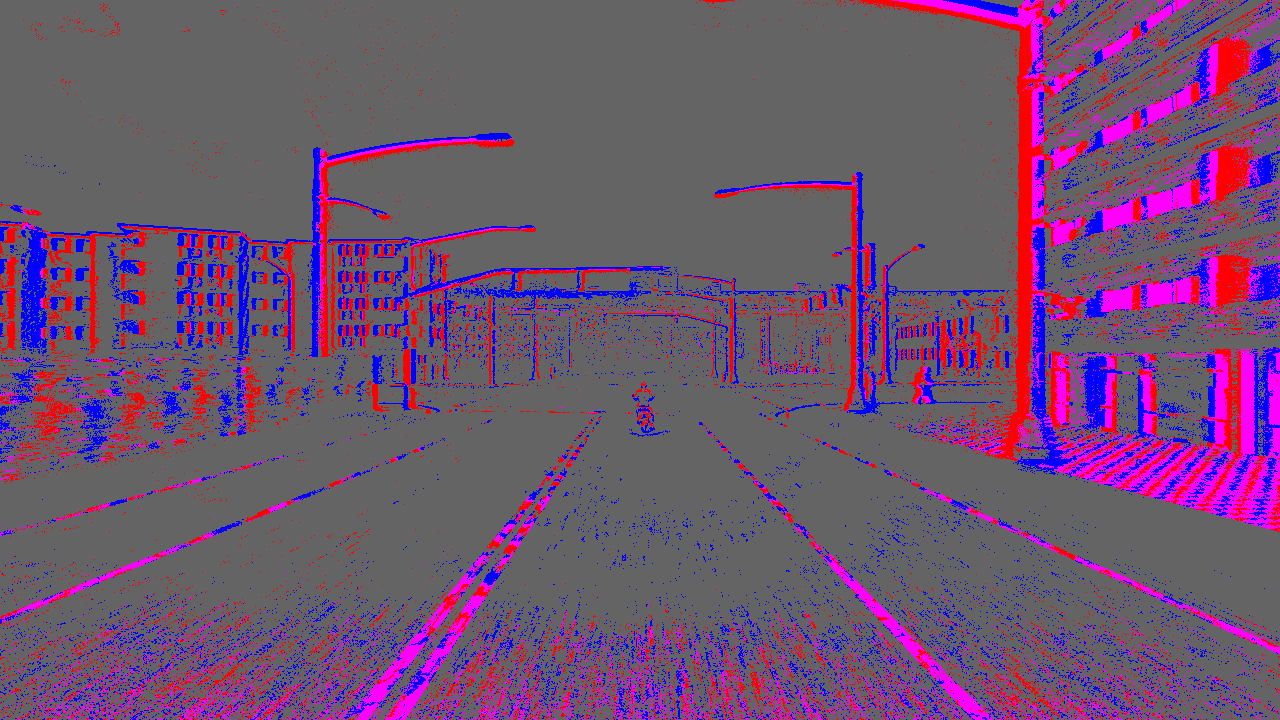}\\
  \includegraphics[width=0.475\linewidth]{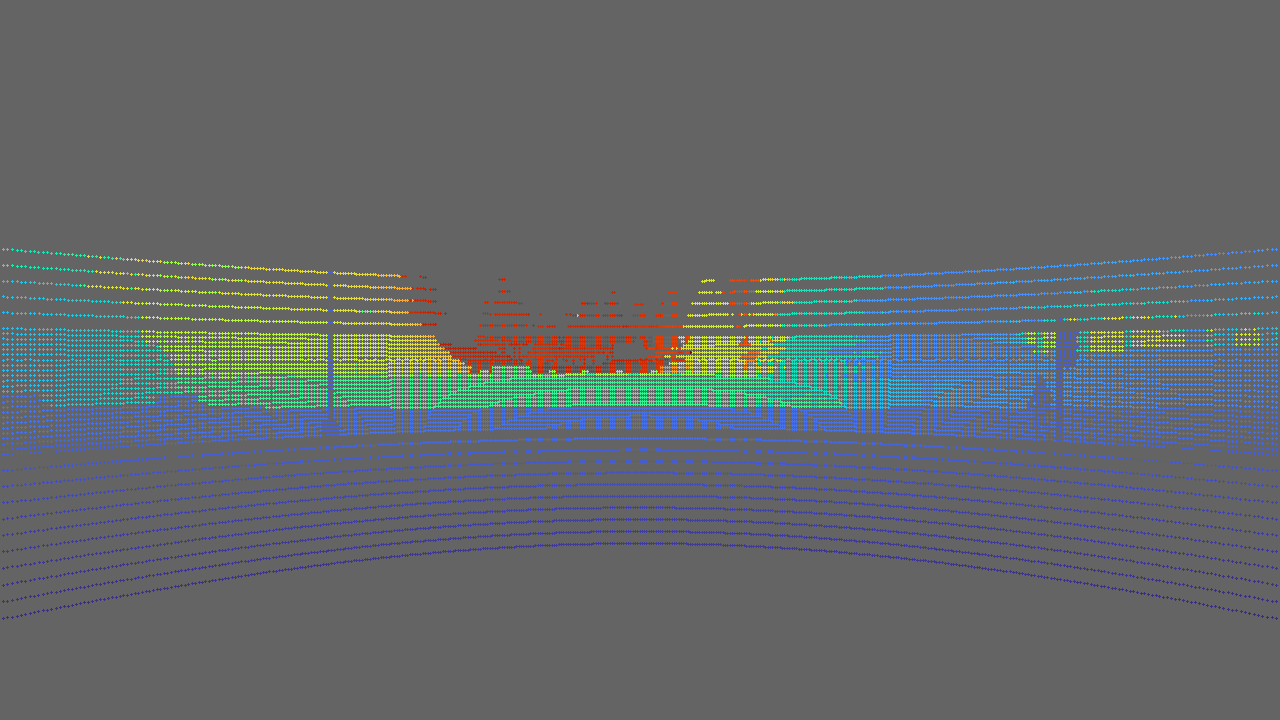}
  \includegraphics[width=0.475\linewidth]{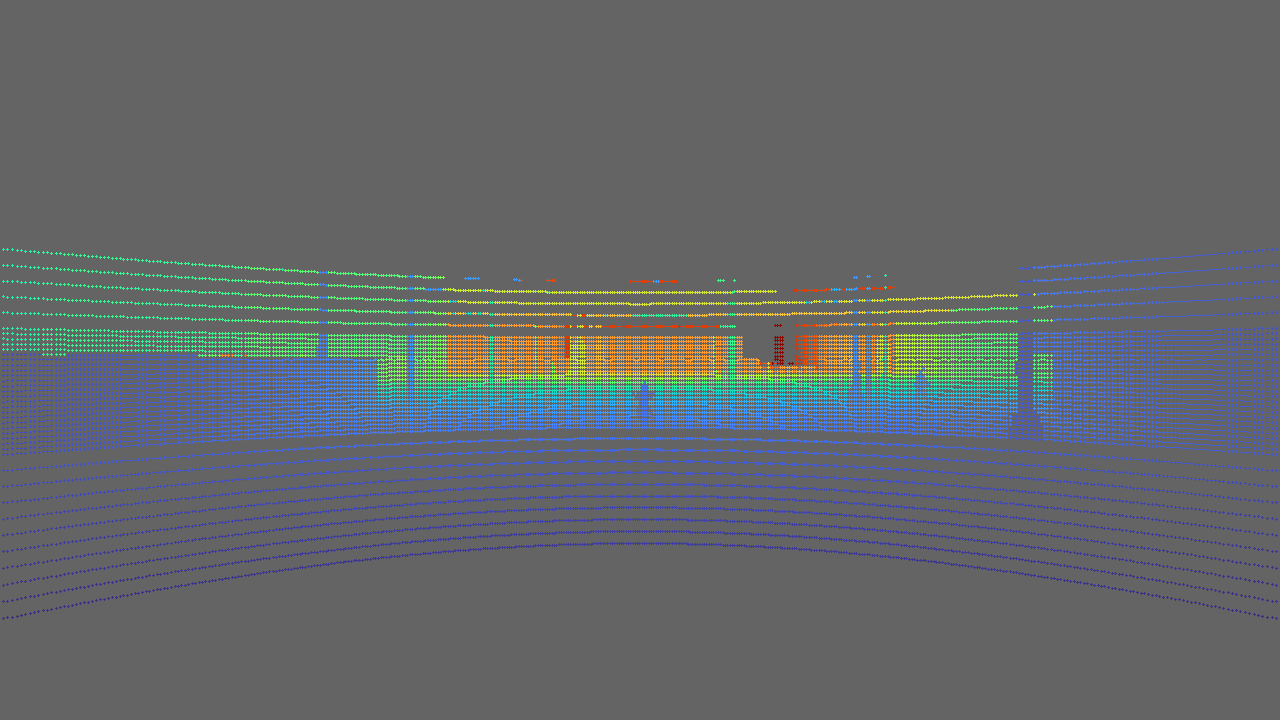}\\
  \includegraphics[width=0.475\linewidth]{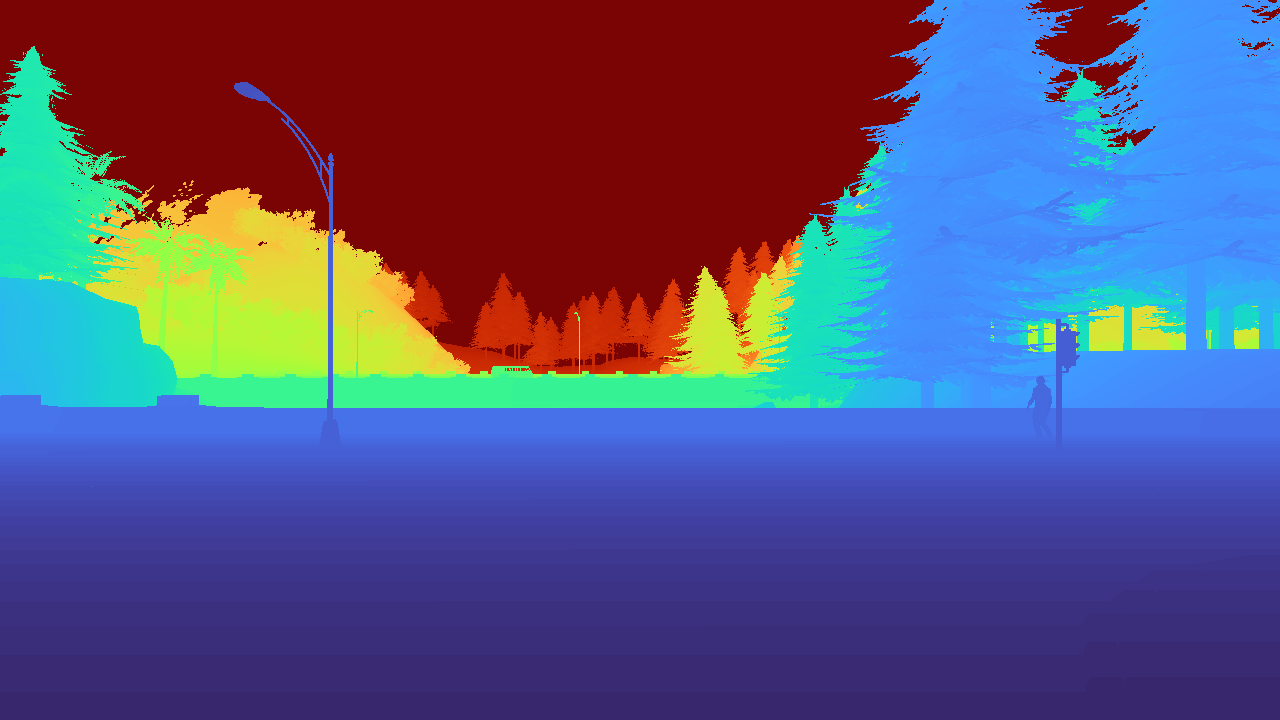}
  \includegraphics[width=0.475\linewidth]{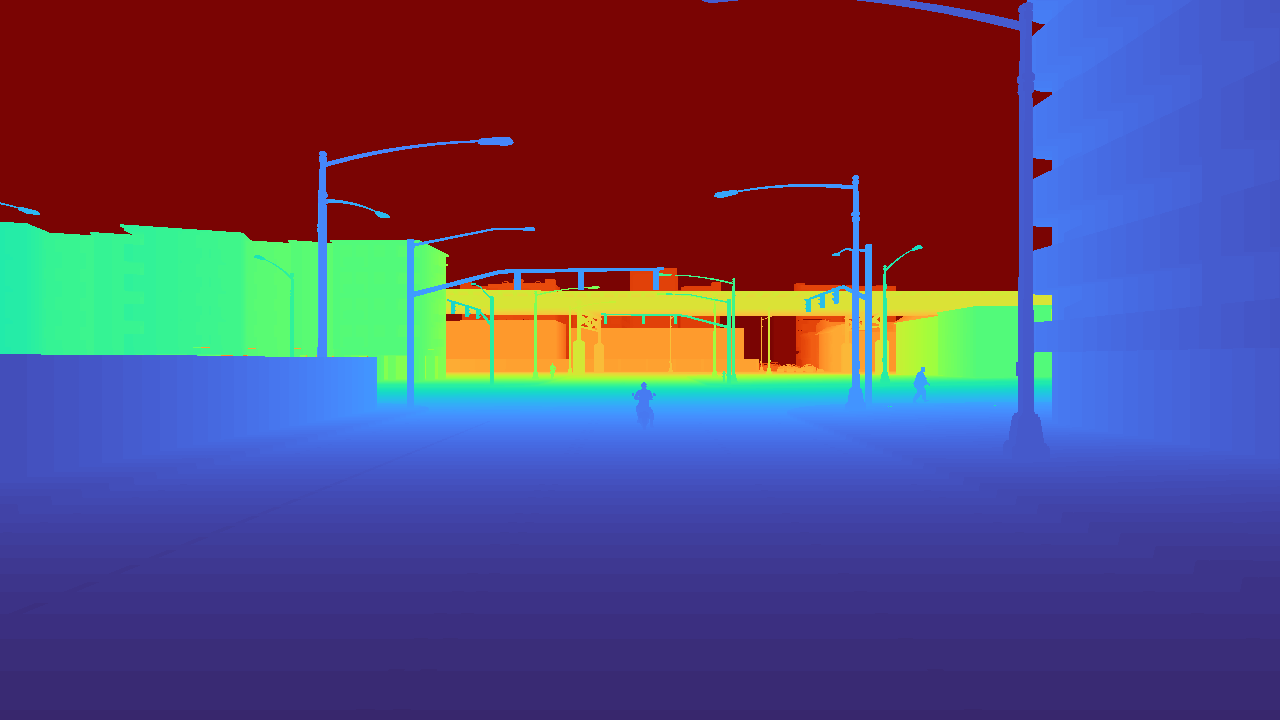}\\
  \includegraphics[width=0.475\linewidth]{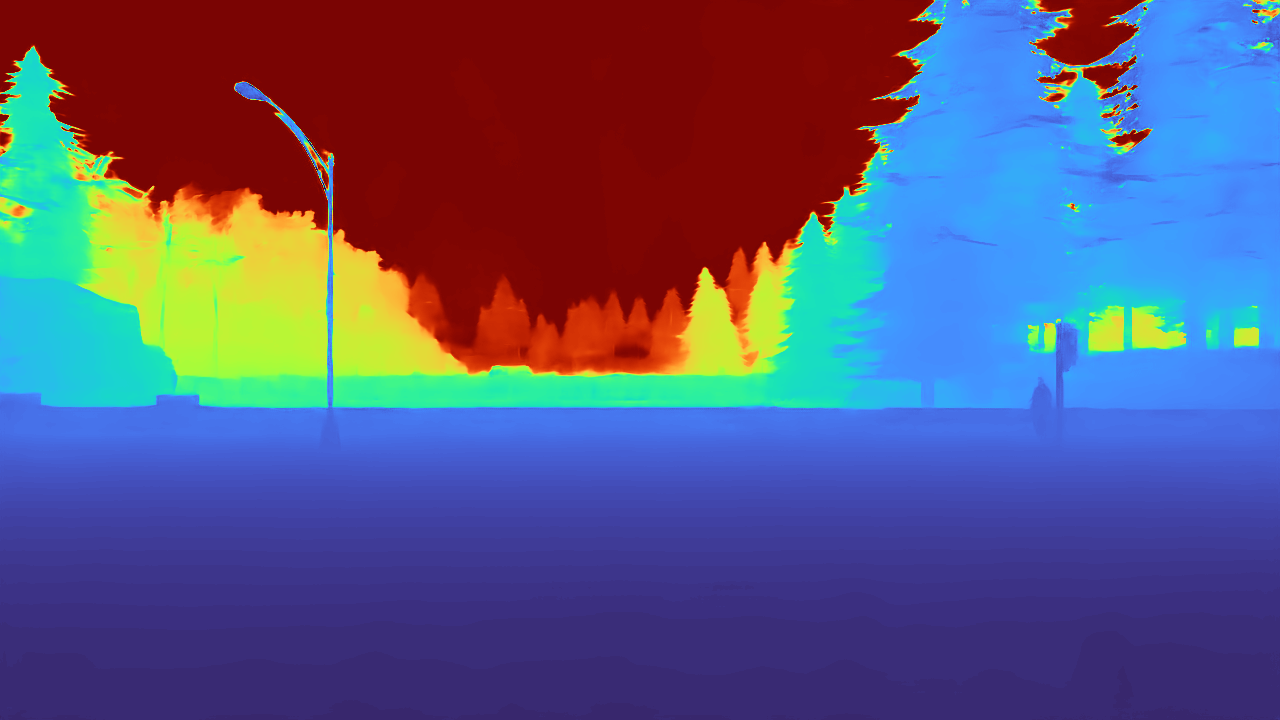}
  \includegraphics[width=0.475\linewidth]{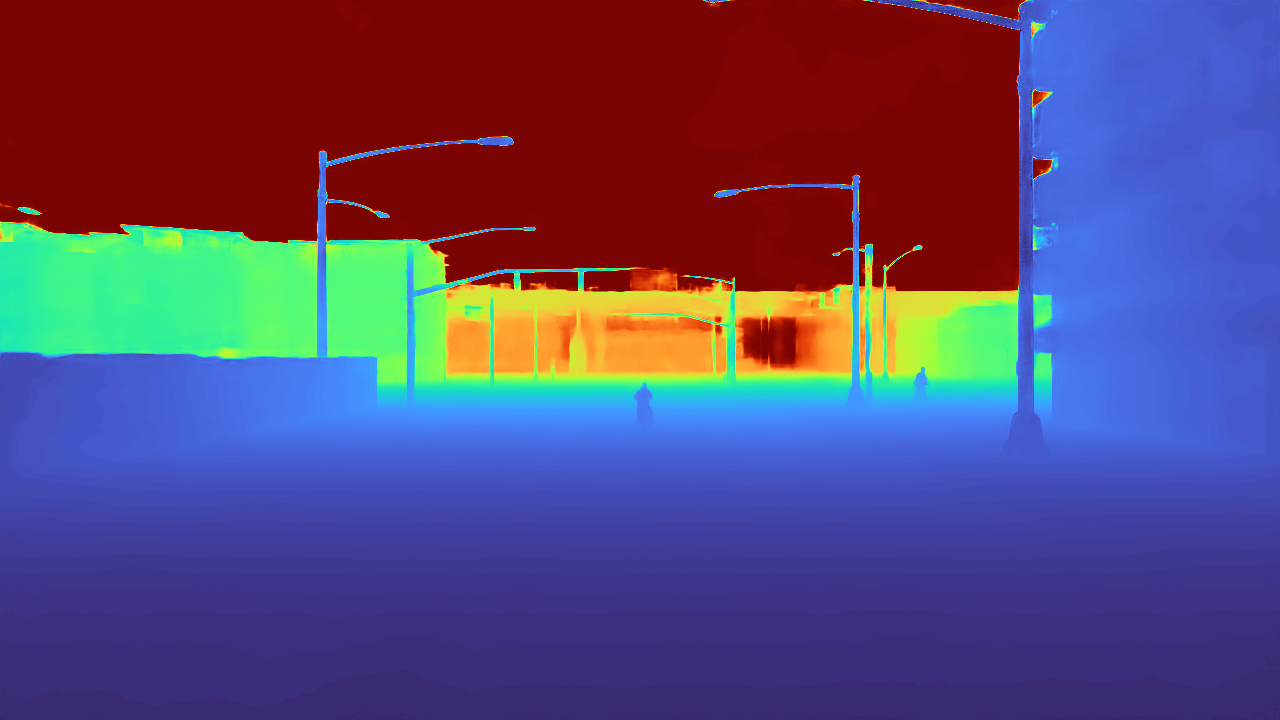}
  \cprotect\caption{Additional results on the SLED dataset, on sequences \verb|Town01_18| and \verb|Town03_02|. From top to bottom: events, LiDAR projection, ground truth, our results.}\label{fig:cmp_sled_additional_good_1}
\end{figure*}

\begin{figure*}
  \centering
  \includegraphics[width=0.475\linewidth]{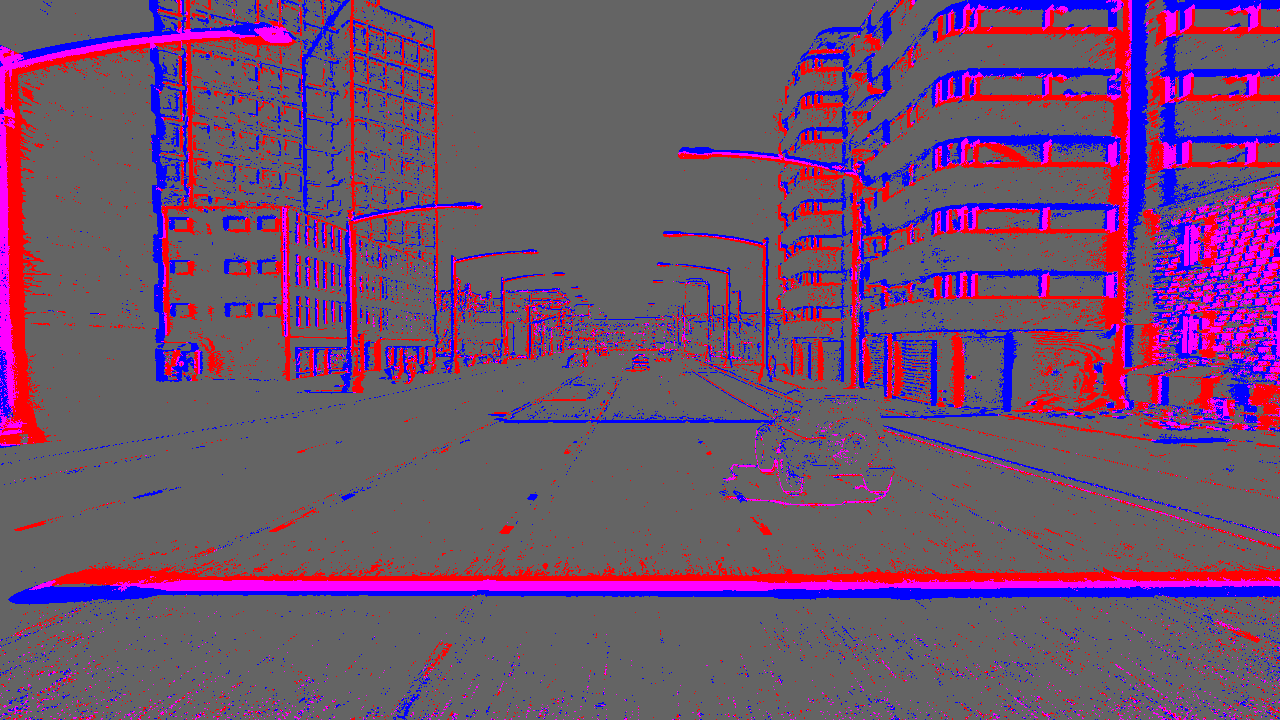}
  \includegraphics[width=0.475\linewidth]{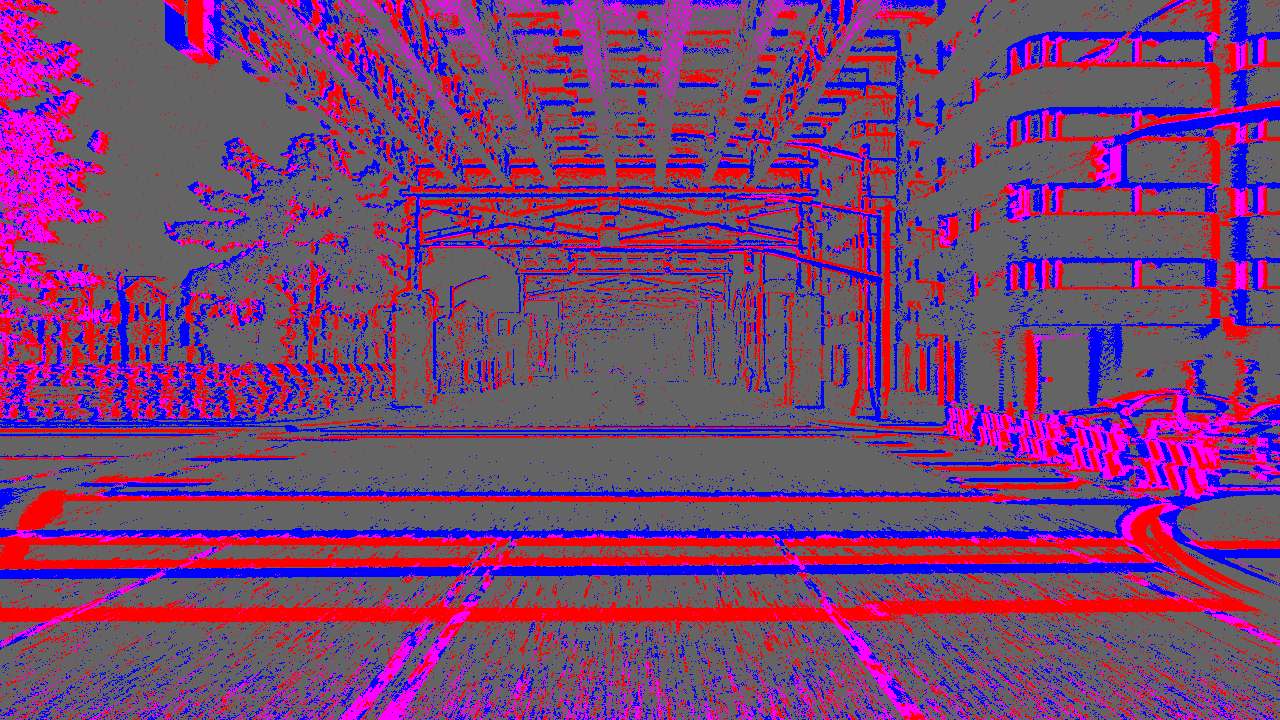}\\
  \includegraphics[width=0.475\linewidth]{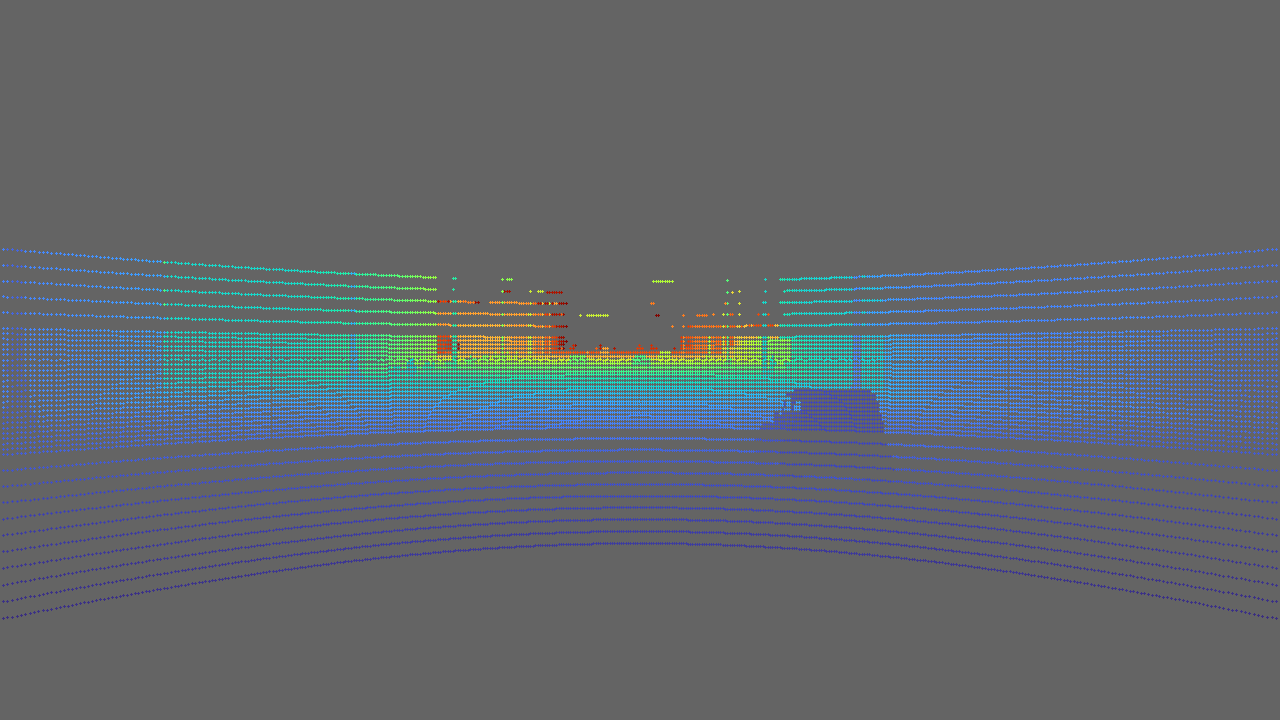}
  \includegraphics[width=0.475\linewidth]{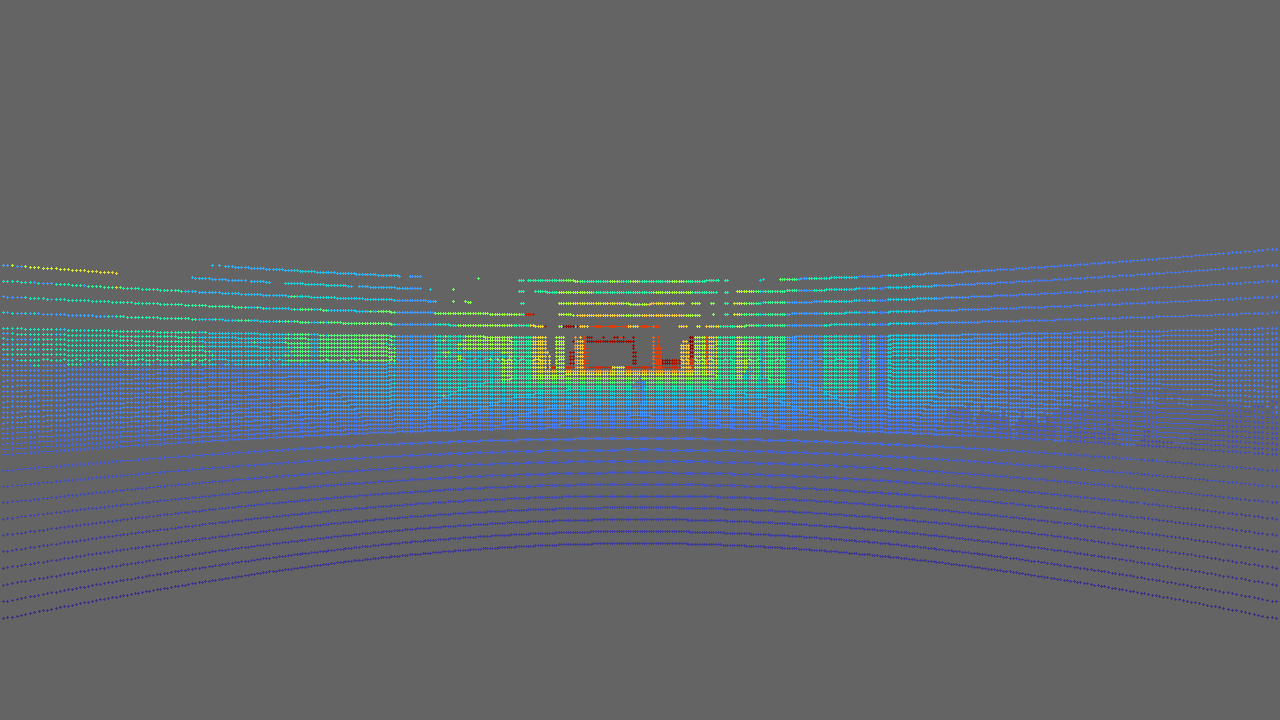}\\
  \includegraphics[width=0.475\linewidth]{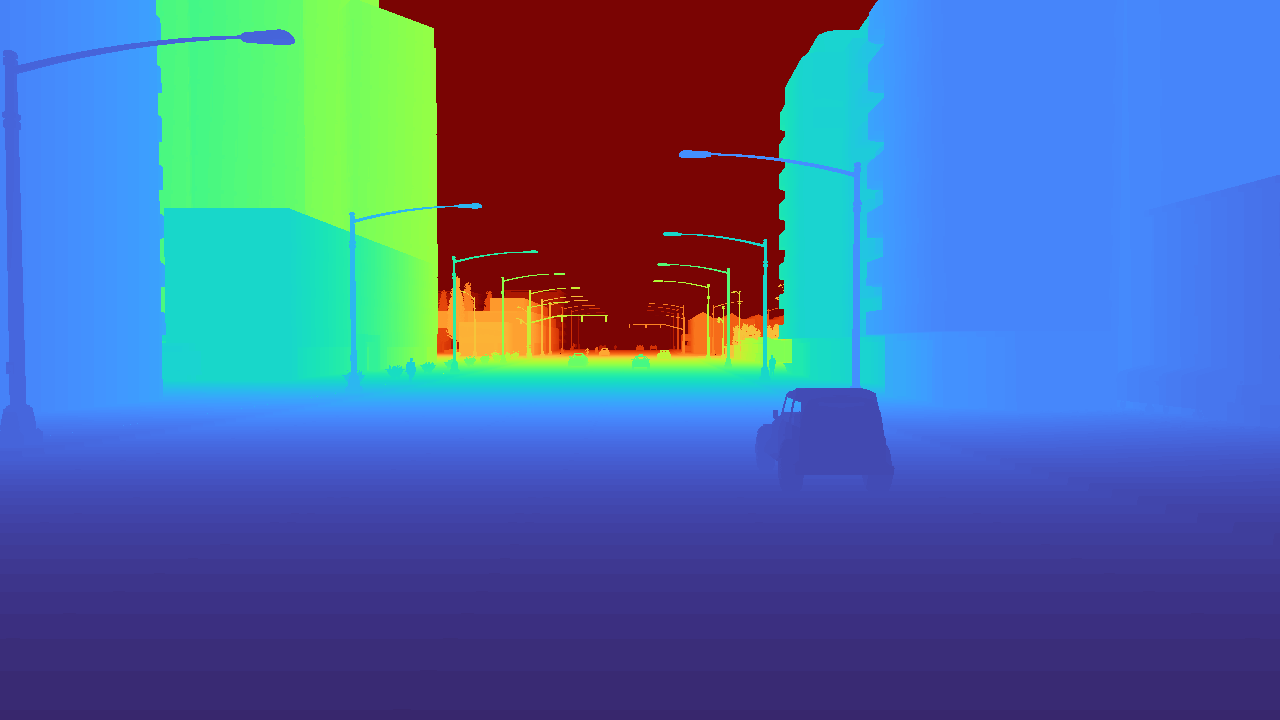}
  \includegraphics[width=0.475\linewidth]{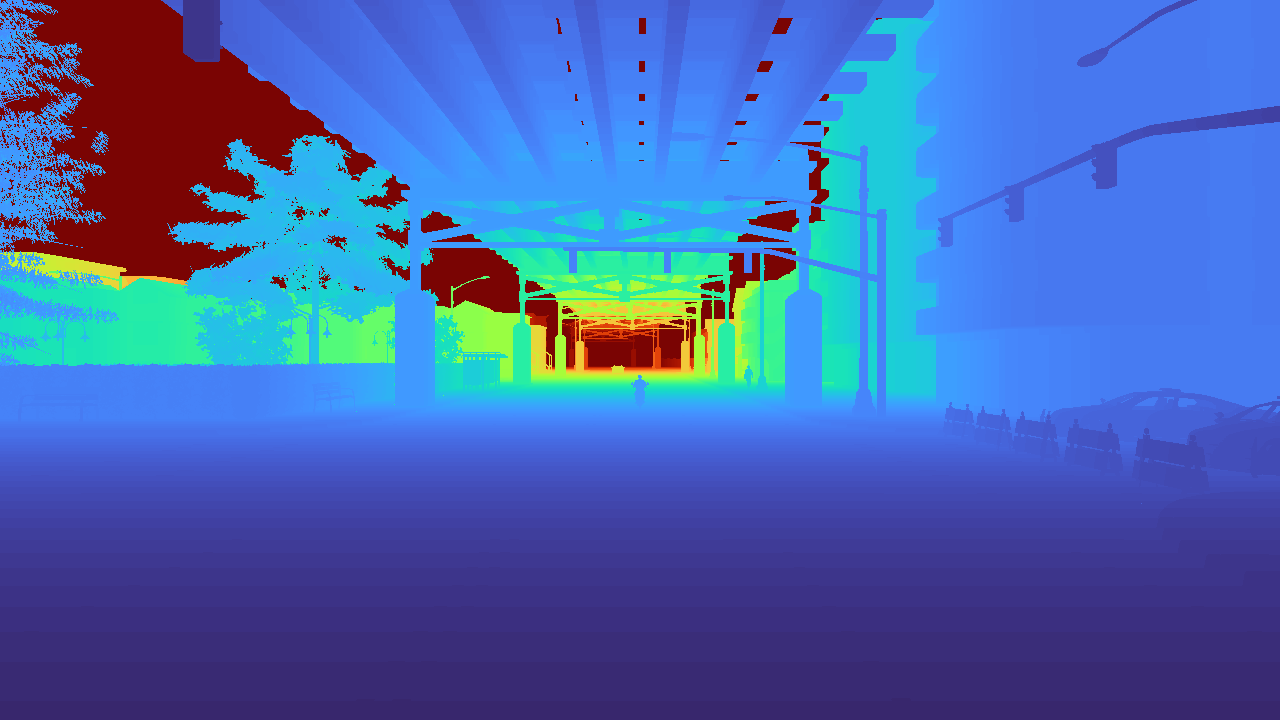}\\
  \includegraphics[width=0.475\linewidth]{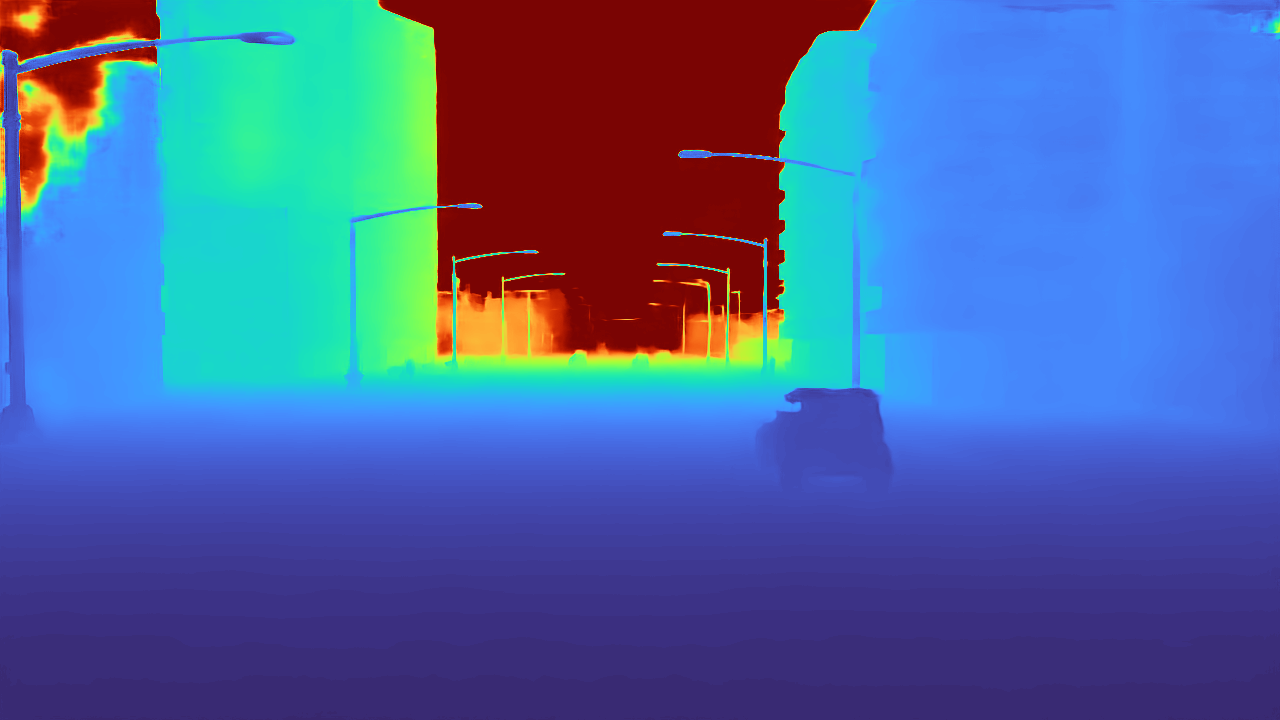}
  \includegraphics[width=0.475\linewidth]{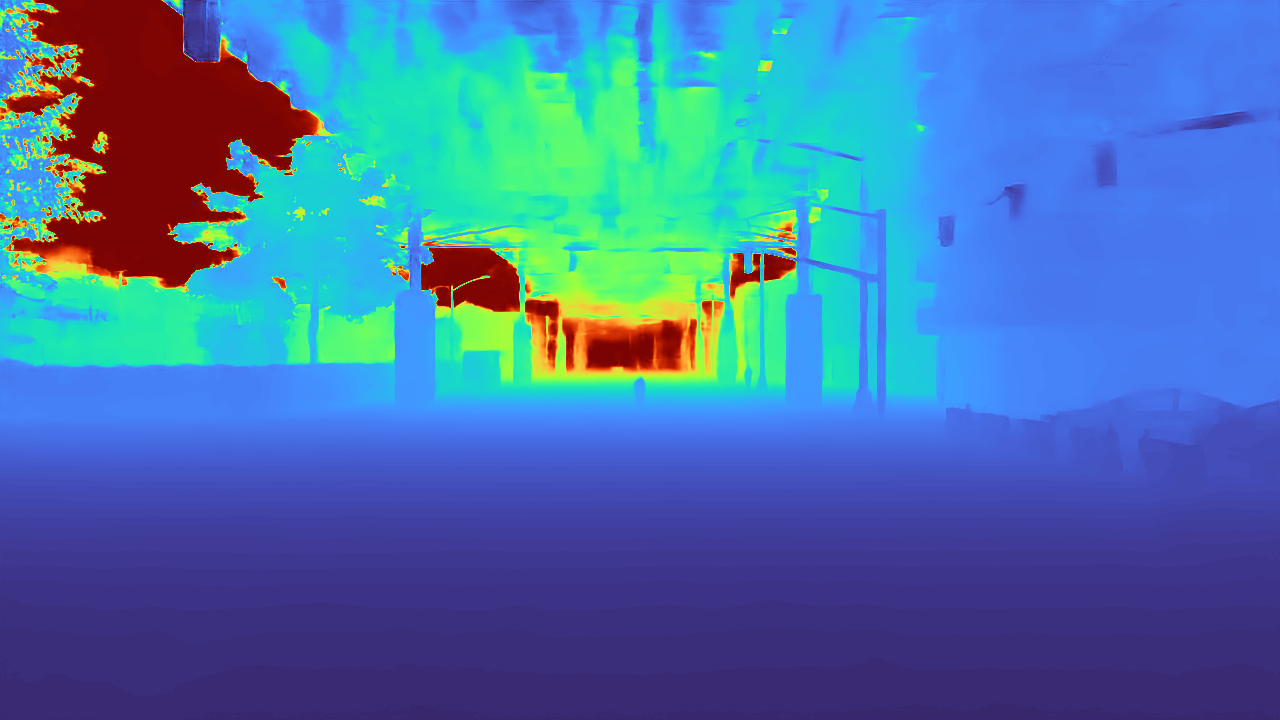}
  \cprotect\caption{Additional results on the SLED dataset, on sequences \verb|Town03_06| and \verb|Town03_13|. From top to bottom: events, LiDAR projection, ground truth, our results. Shown here are two cases where DELTA\textsubscript{SL} displays moderate to large errors for objects in the upper part of the depth maps (where no LiDAR data is available), like the building on the top left for the left column, and the suspended railway on the top for the right column.}\label{fig:cmp_sled_additional_bad_0}
\end{figure*}

\begin{figure*}
  \centering
  \includegraphics[width=0.475\linewidth]{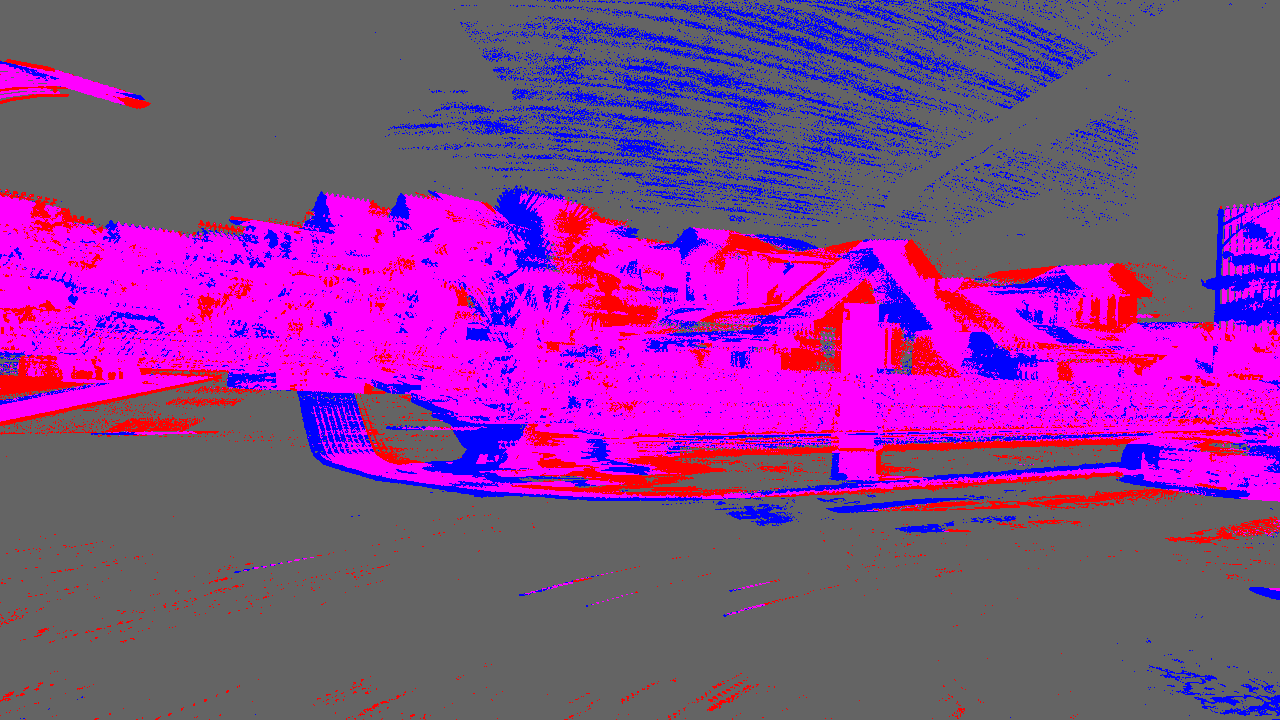}
  \includegraphics[width=0.475\linewidth]{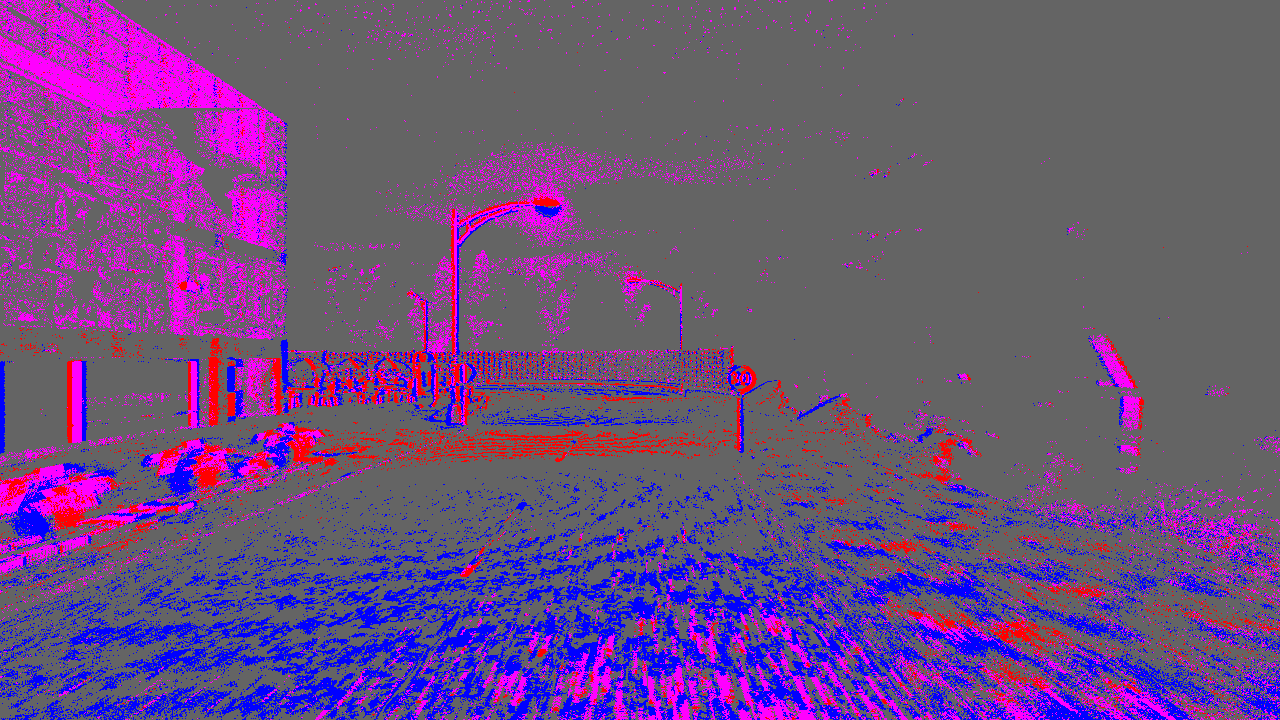}\\
  \includegraphics[width=0.475\linewidth]{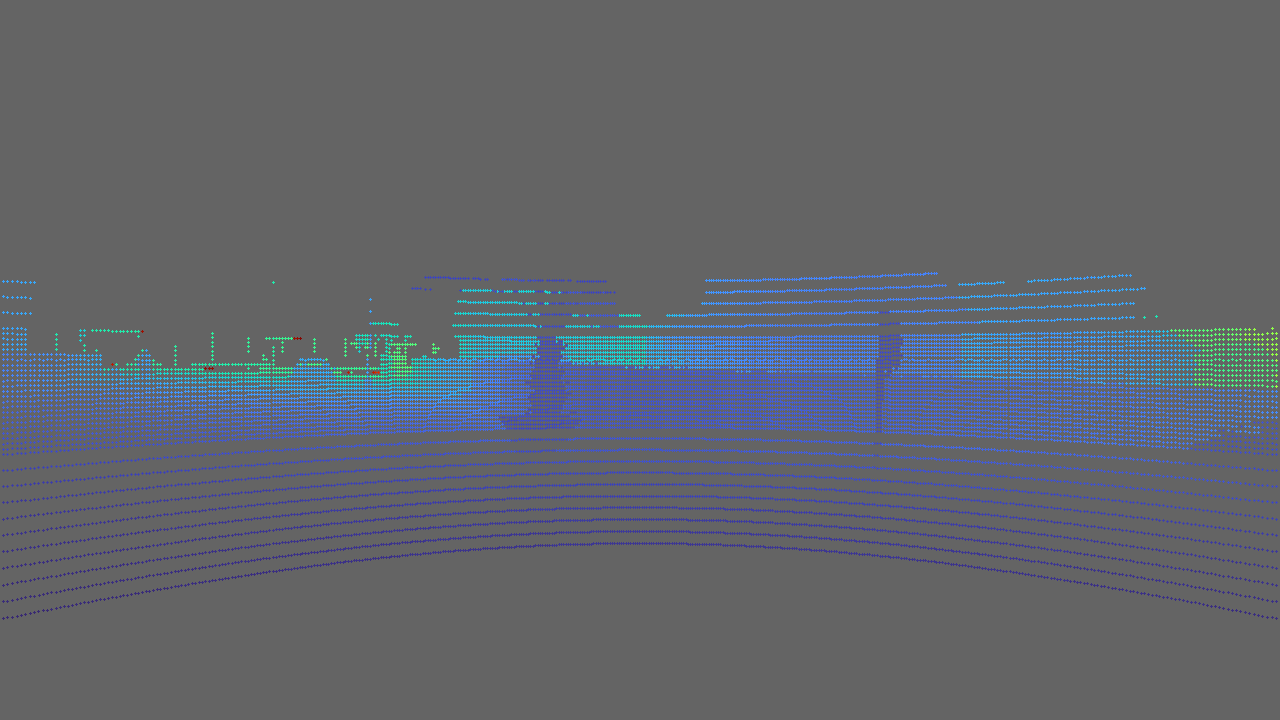}
  \includegraphics[width=0.475\linewidth]{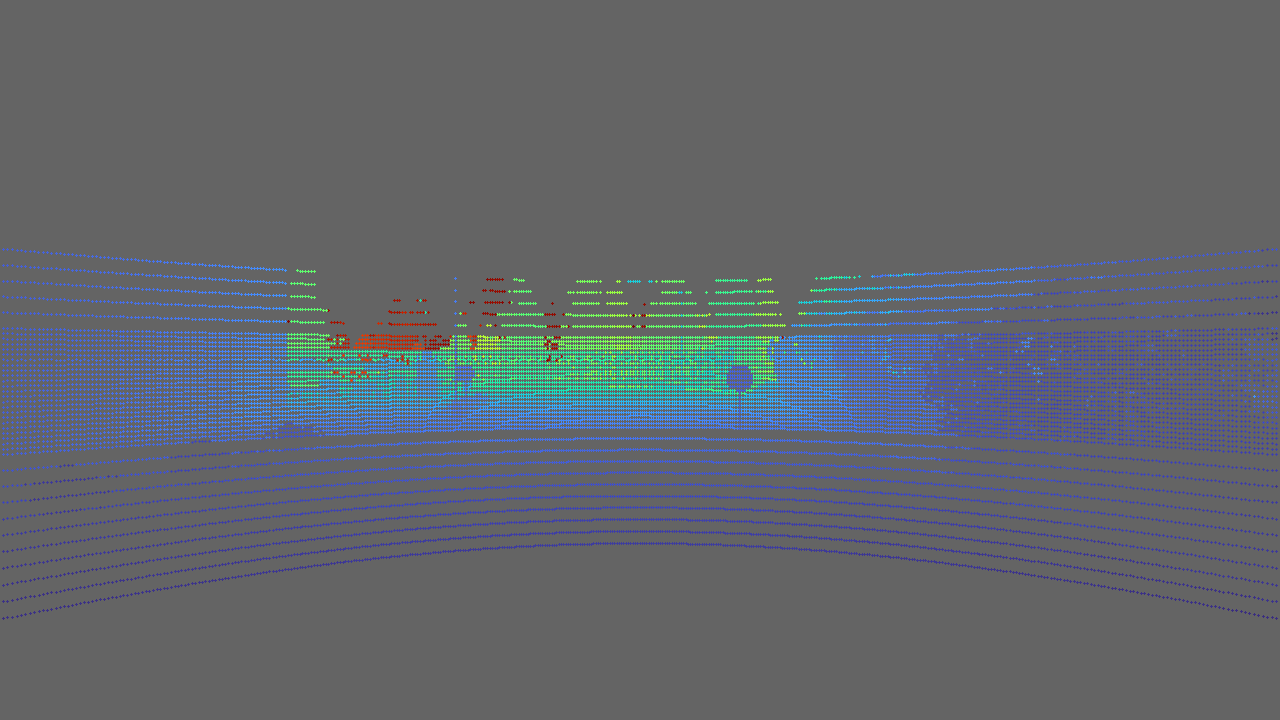}\\
  \includegraphics[width=0.475\linewidth]{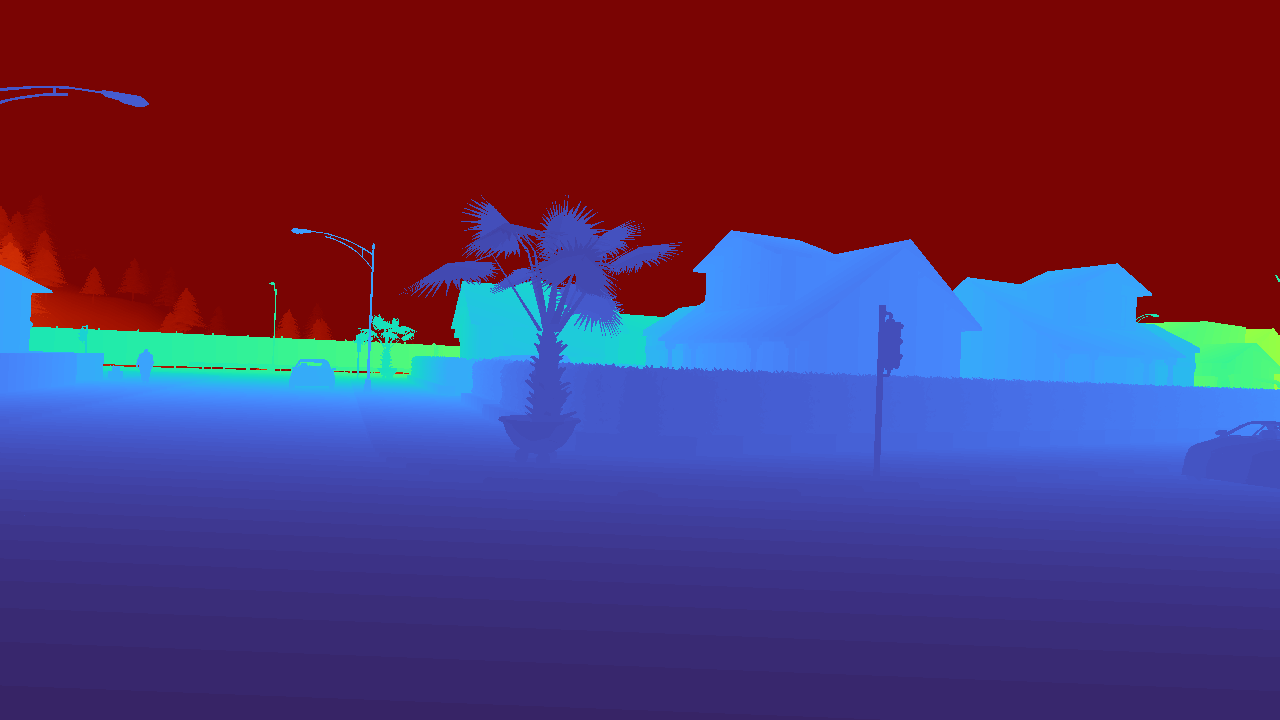}
  \includegraphics[width=0.475\linewidth]{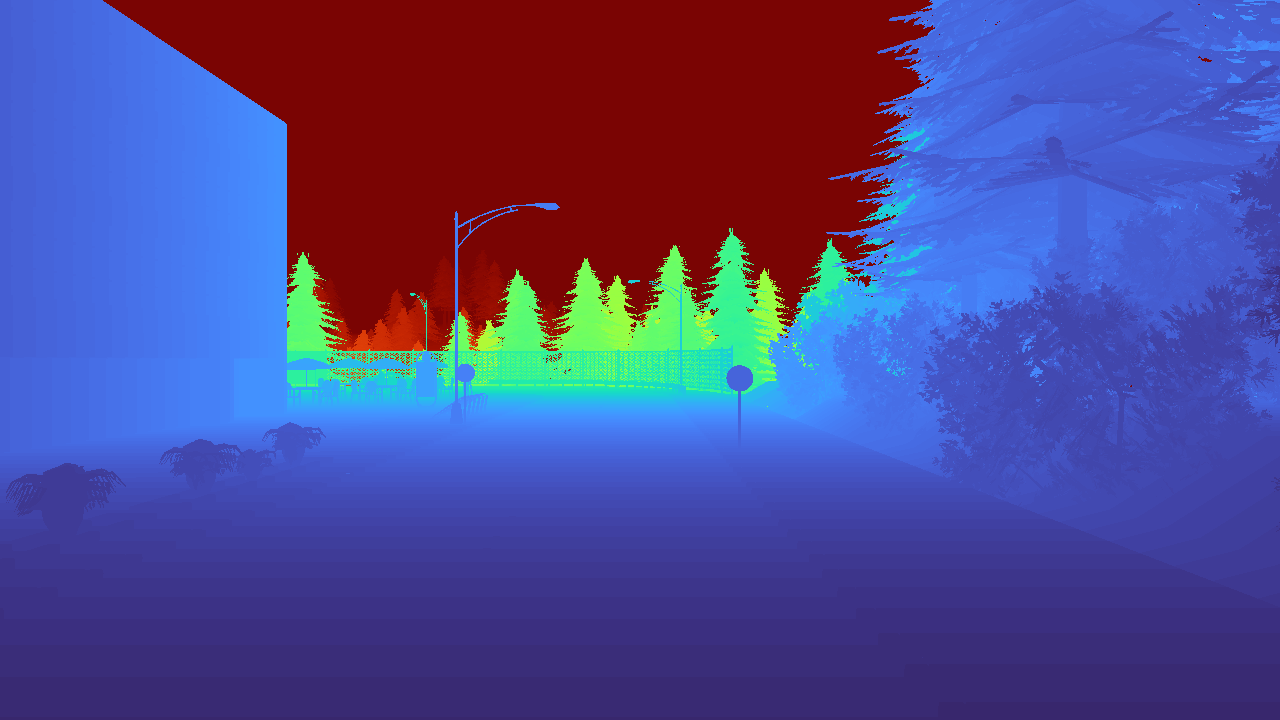}\\
  \includegraphics[width=0.475\linewidth]{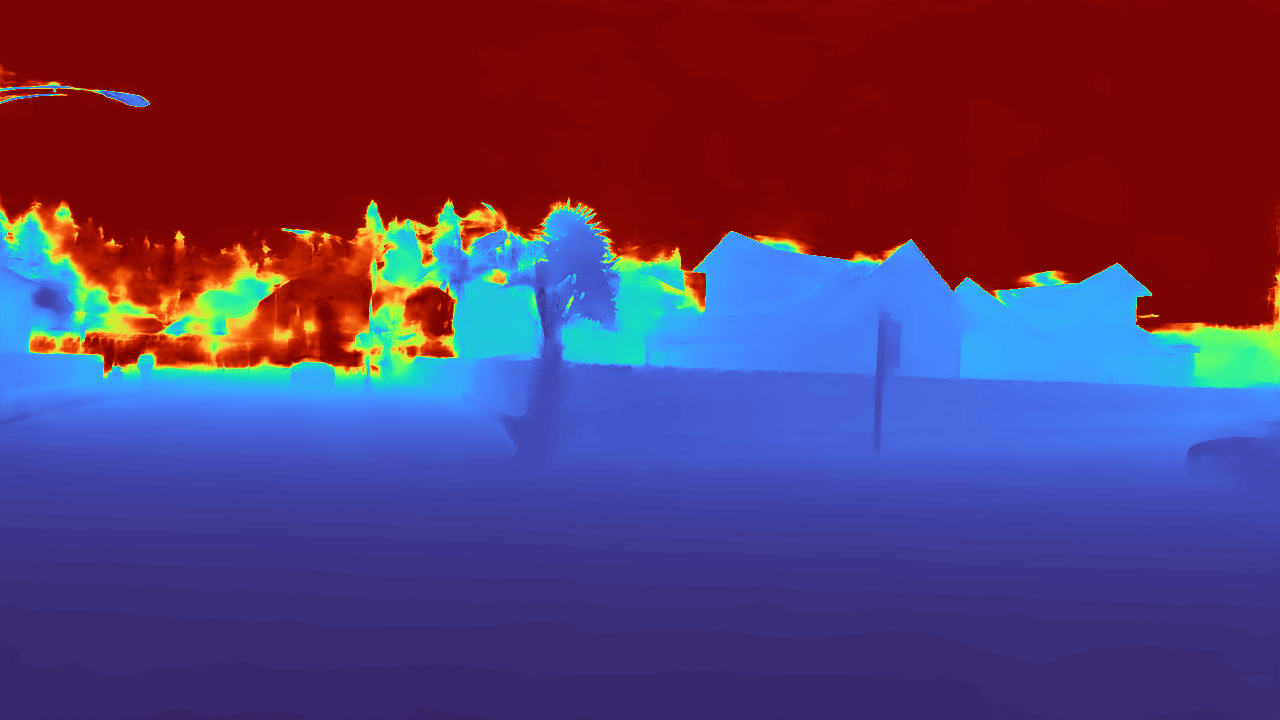}
  \includegraphics[width=0.475\linewidth]{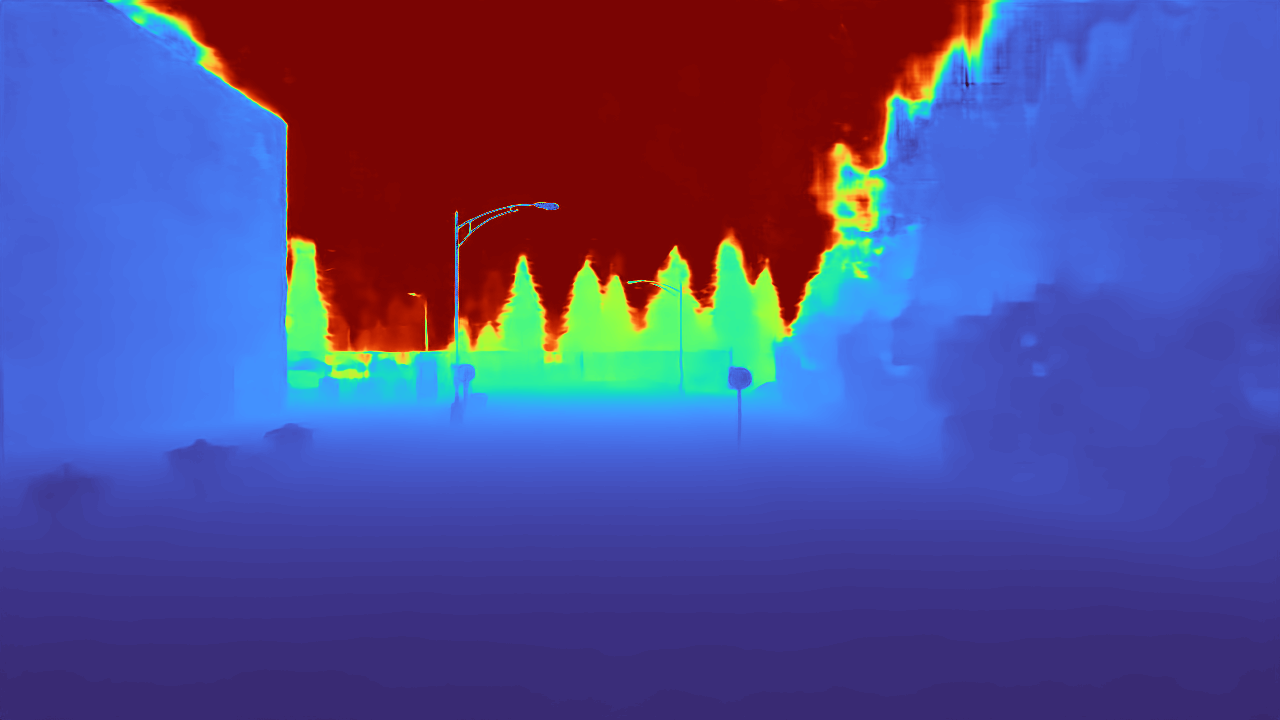}
  \cprotect\caption{Additional results on the SLED dataset, on sequences \verb|Town01_08| and \verb|Town01_11|. From top to bottom: events, LiDAR projection, ground truth, our results. Shown here are two failure cases where DELTA\textsubscript{SL} displays large errors. Left: due to a high-speed sharp turn, a very high quantity of events is produced in the time window of accumulation, leading to information being lost in the event volume, and thus leading to an inaccurate depth estimation for background objects. Right: due to the limitations of the event camera in the CARLA simulator, dark objects in a night scene like the trees in the middle and on the right of the scene are not captured in the event stream of the SLED dataset, resulting in blurry depth estimations for these objects.}\label{fig:cmp_sled_additional_bad_1}
\end{figure*}

Additional qualitative results on the SLED dataset are given in \cref{fig:cmp_sled_additional_good_0,fig:cmp_sled_additional_good_1,fig:cmp_sled_additional_bad_0,fig:cmp_sled_additional_bad_1}. We showcase in \cref{fig:cmp_sled_additional_good_0,fig:cmp_sled_additional_good_1} scenes with accurate estimations, but also some small and larger failure cases in \cref{fig:cmp_sled_additional_bad_0,fig:cmp_sled_additional_bad_1}.

\section{Additional Visual Results on the MVSEC Dataset}
\begin{figure*}
  \centering
  \includegraphics[width=0.24\linewidth]{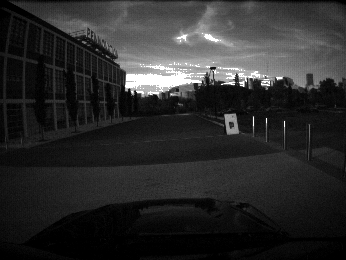}
  \includegraphics[width=0.24\linewidth]{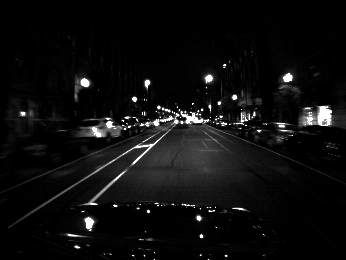}
  \includegraphics[width=0.24\linewidth]{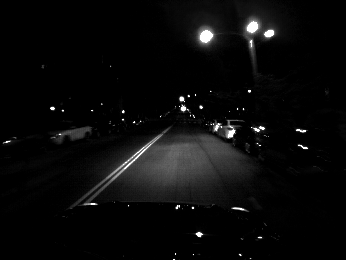}
  \includegraphics[width=0.24\linewidth]{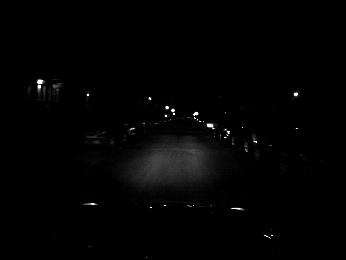}\\
  \includegraphics[width=0.24\linewidth]{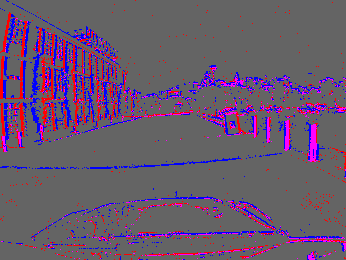}
  \includegraphics[width=0.24\linewidth]{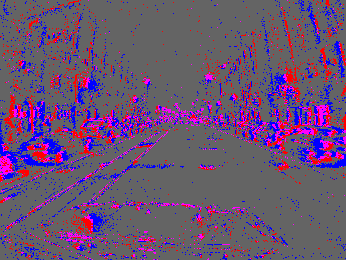}
  \includegraphics[width=0.24\linewidth]{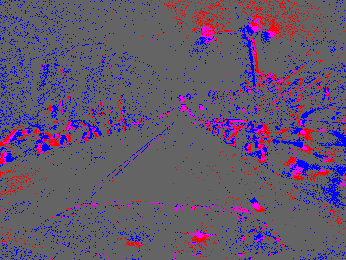}
  \includegraphics[width=0.24\linewidth]{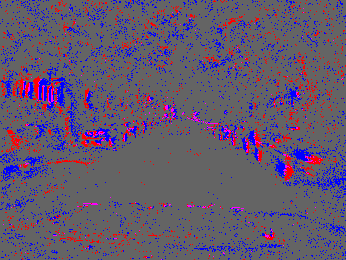}\\
  \includegraphics[width=0.24\linewidth]{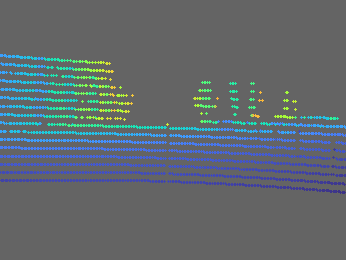}
  \includegraphics[width=0.24\linewidth]{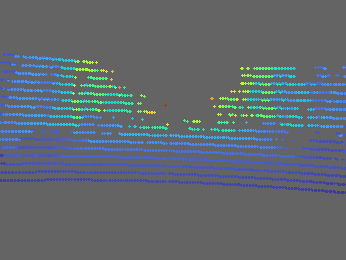}
  \includegraphics[width=0.24\linewidth]{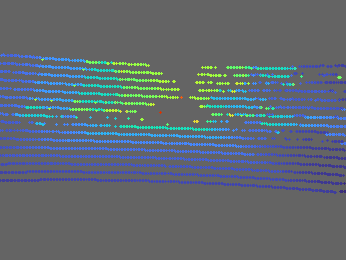}
  \includegraphics[width=0.24\linewidth]{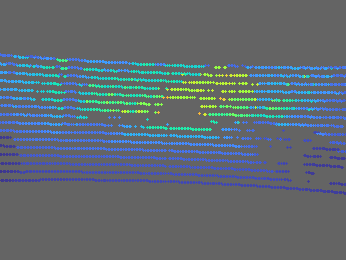}\\
  \includegraphics[width=0.24\linewidth]{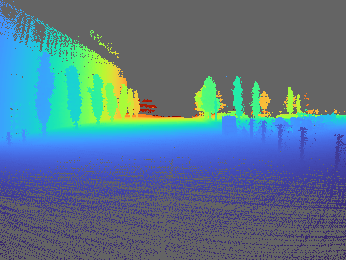}
  \includegraphics[width=0.24\linewidth]{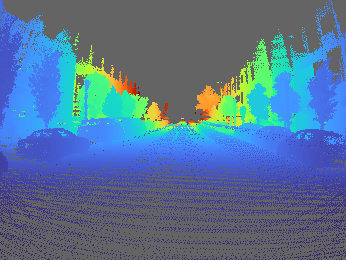}
  \includegraphics[width=0.24\linewidth]{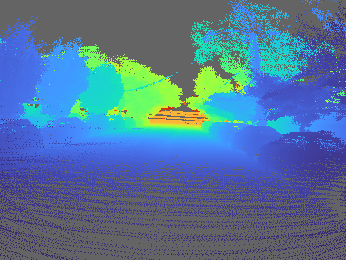}
  \includegraphics[width=0.24\linewidth]{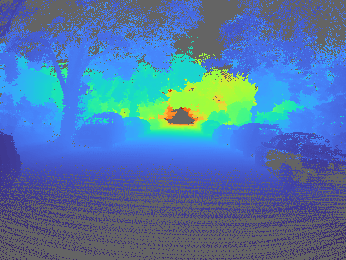}\\
  \includegraphics[width=0.24\linewidth]{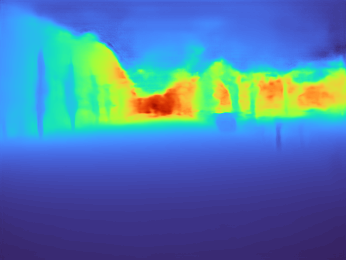}
  \includegraphics[width=0.24\linewidth]{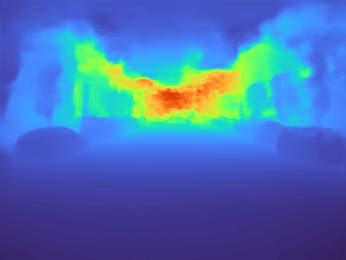}
  \includegraphics[width=0.24\linewidth]{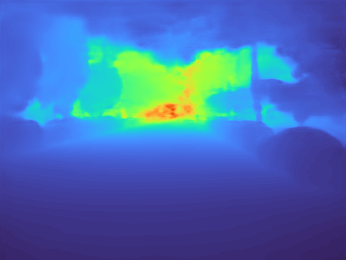}
  \includegraphics[width=0.24\linewidth]{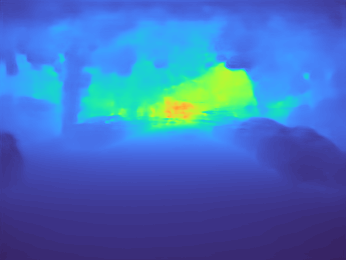}\\
  \includegraphics[width=0.24\linewidth]{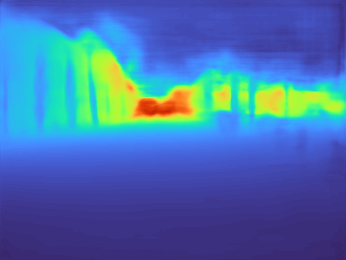}
  \includegraphics[width=0.24\linewidth]{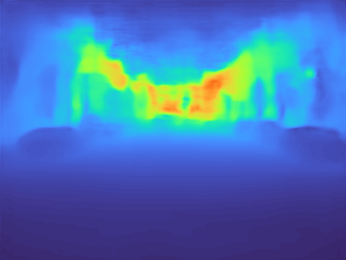}
  \includegraphics[width=0.24\linewidth]{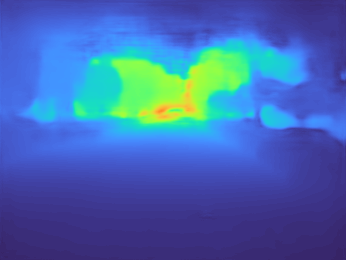}
  \includegraphics[width=0.24\linewidth]{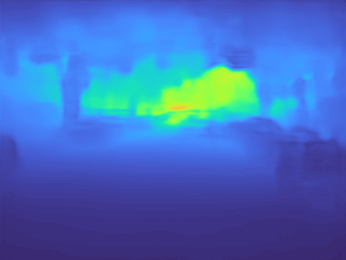}
  \cprotect\caption{Additional results on the MVSEC dataset. Sequences shown, from left to right: \verb|outdoor_day_1|; \verb|outdoor_night_1|; \verb|outdoor_night_2|; \verb|outdoor_night_3|. From top to bottom: reference image of the scene; events; LiDAR projection (with size of points increased for a better visibility); ground truth; our results (DELTA\textsubscript{MV}, DELTA\textsubscript{SL\(\rightarrow\)MV}).}\label{fig:cmp_mvsec_additional}
\end{figure*}

Additional qualitative results on the MVSEC dataset are given in \cref{fig:cmp_mvsec_additional}, showing the quality of the results for both day and night scenes despite the sparse and low-resolution input event and LiDAR data.

\section{Additional Visual Results on the M3ED Dataset}
\begin{figure*}
  \centering
  \includegraphics[width=0.4325\linewidth]{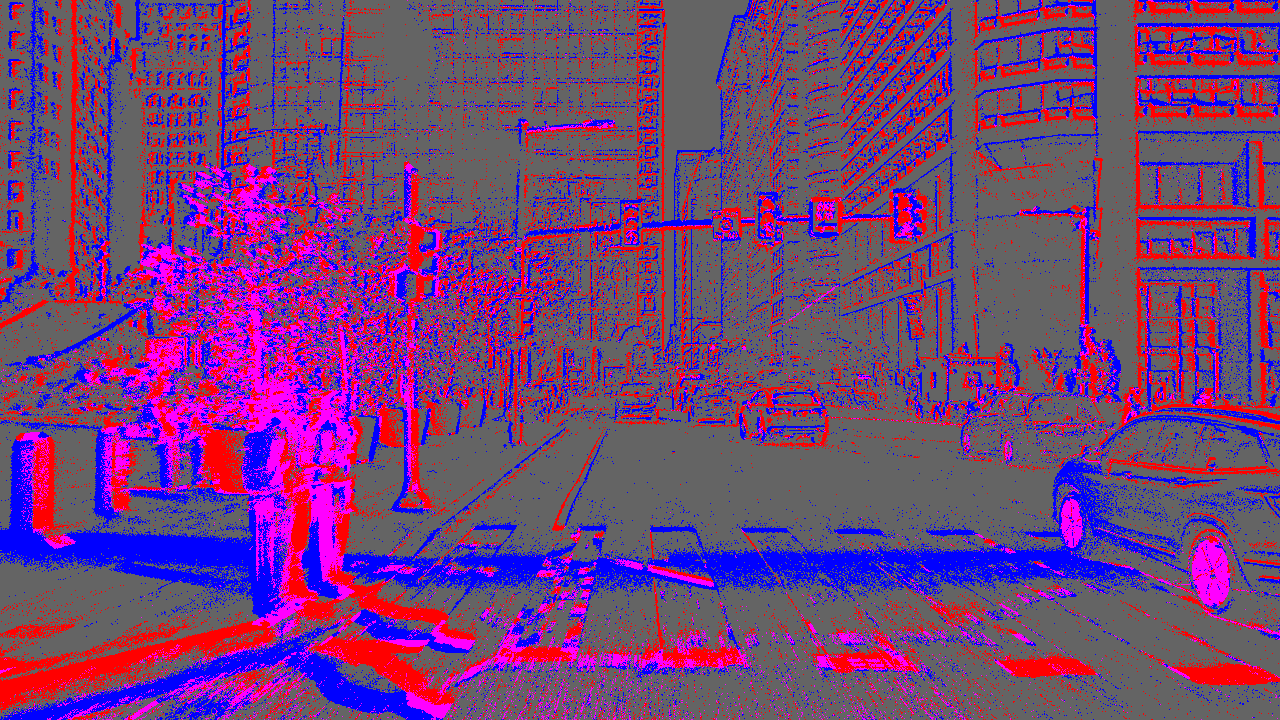}
  \includegraphics[width=0.4325\linewidth]{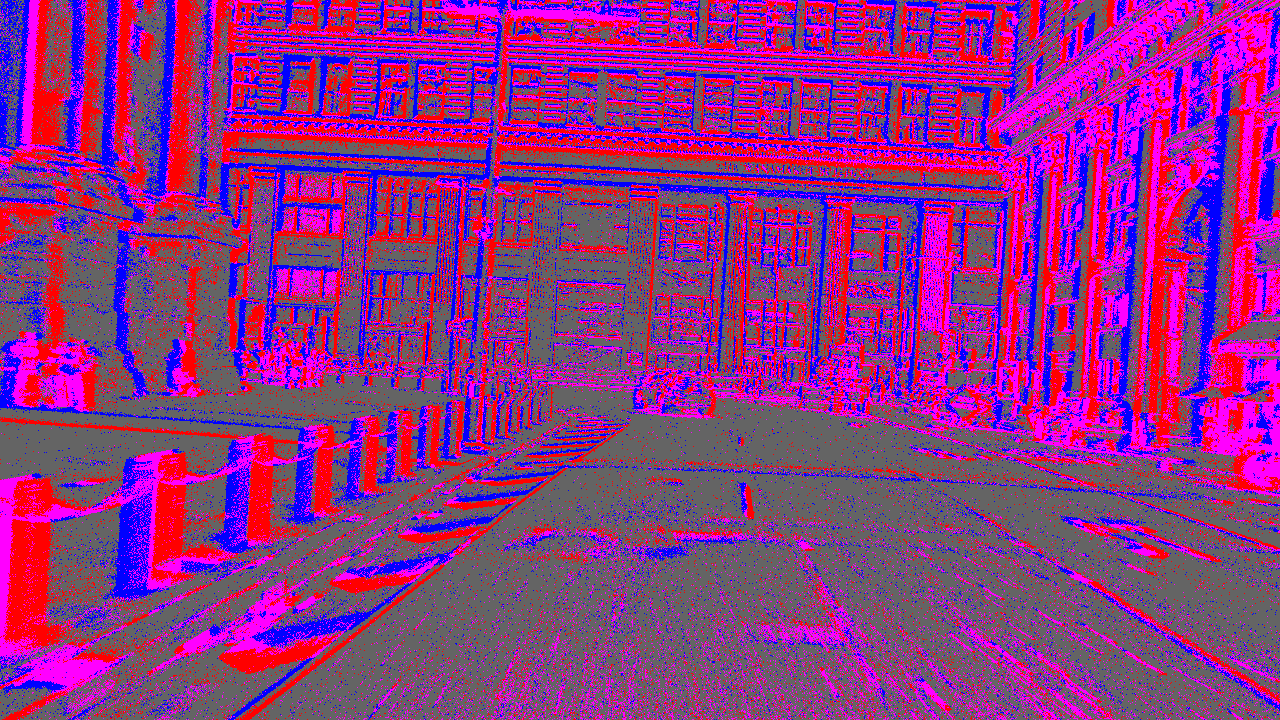}\\
  \includegraphics[width=0.4325\linewidth]{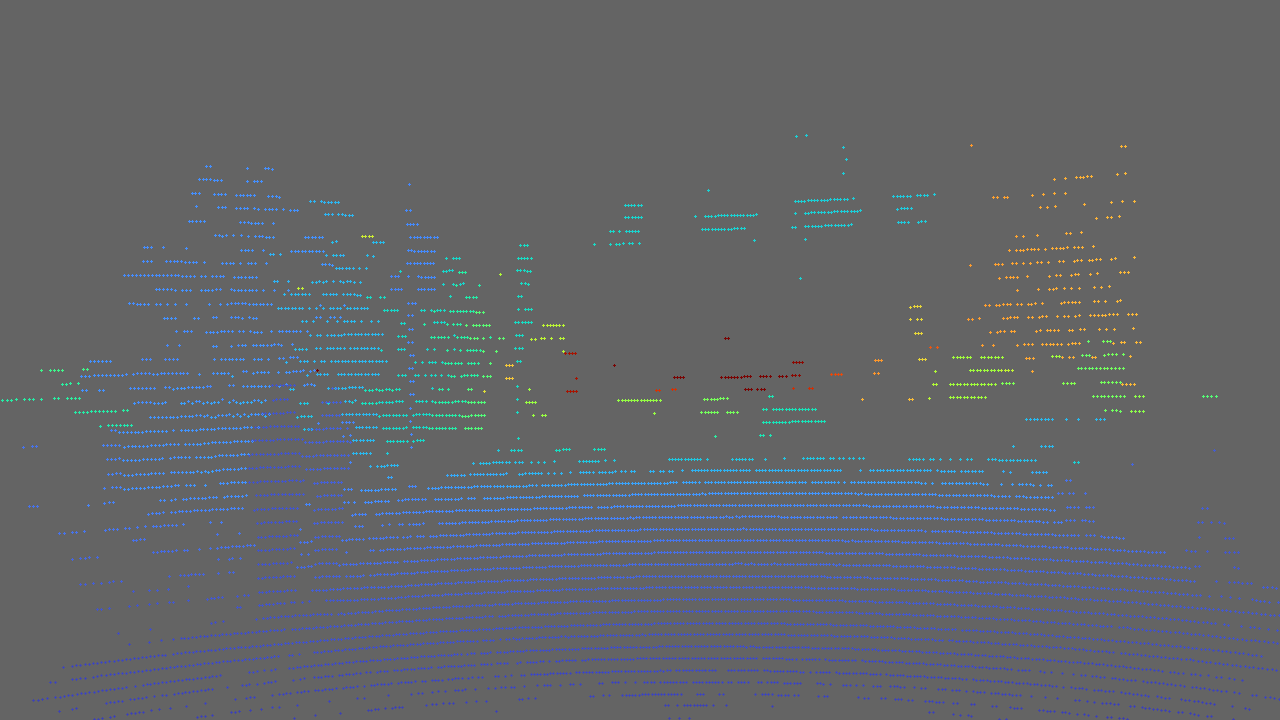}
  \includegraphics[width=0.4325\linewidth]{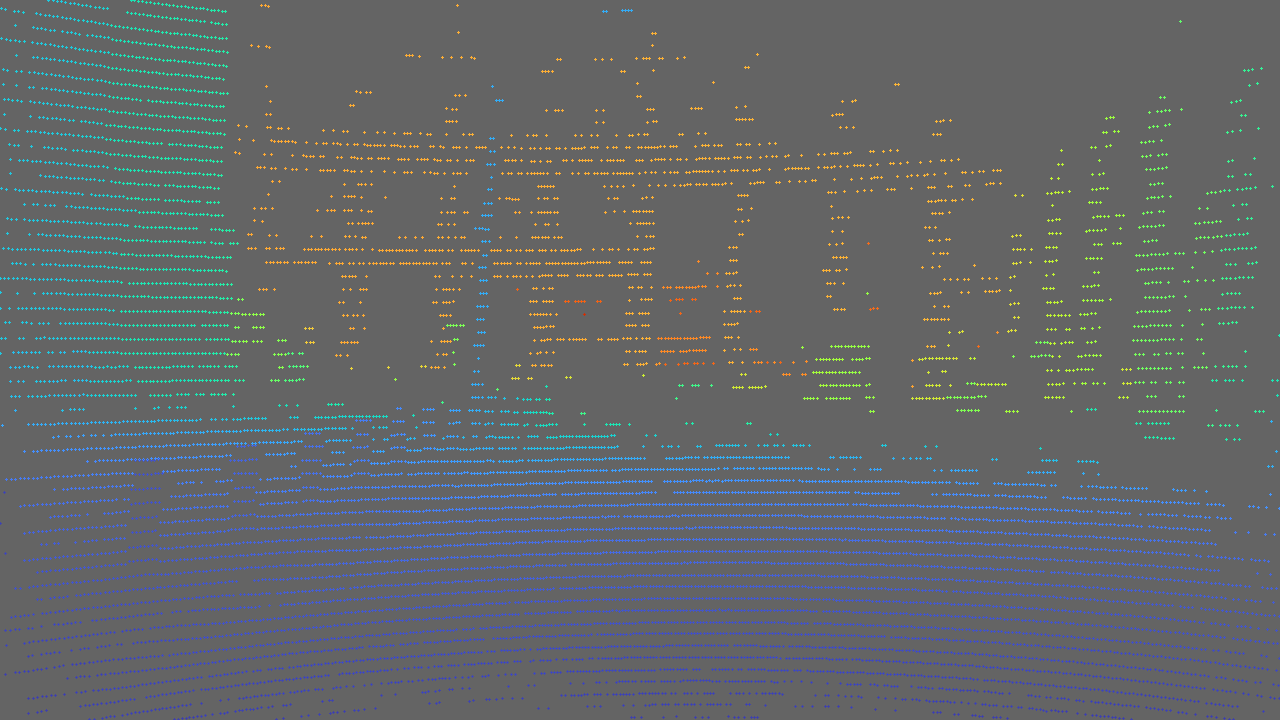}\\
  \includegraphics[width=0.4325\linewidth]{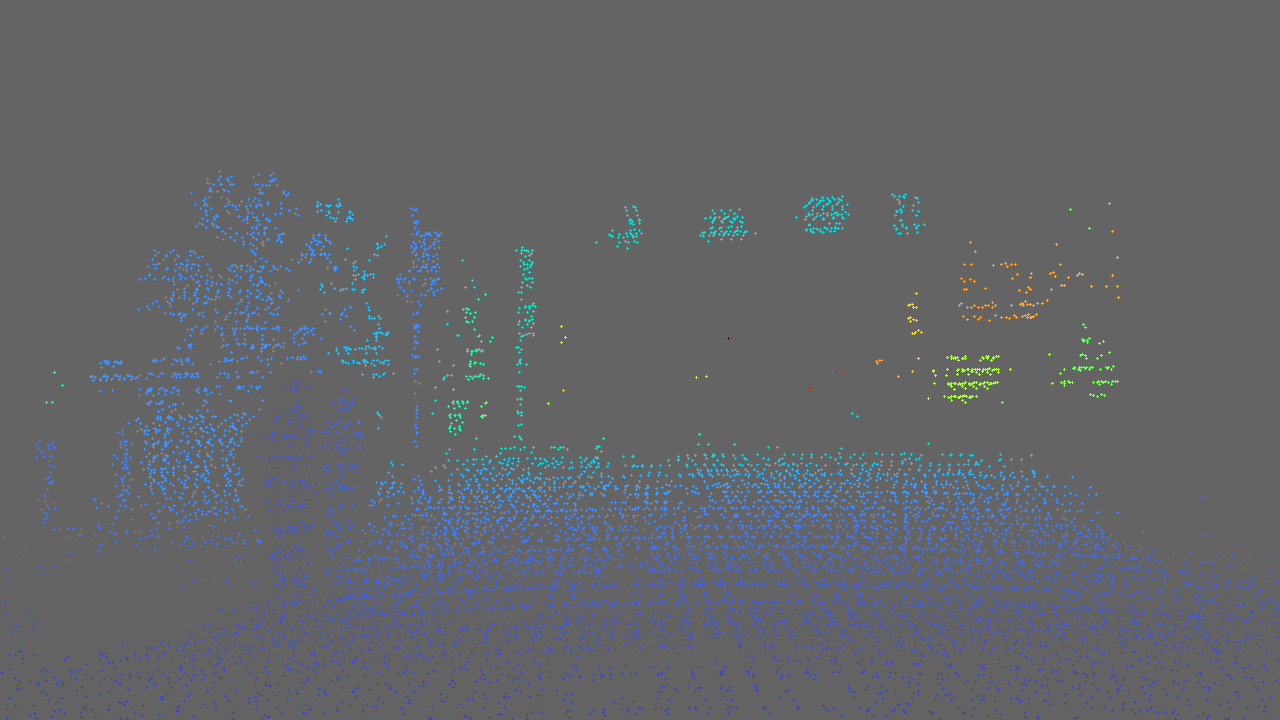}
  \includegraphics[width=0.4325\linewidth]{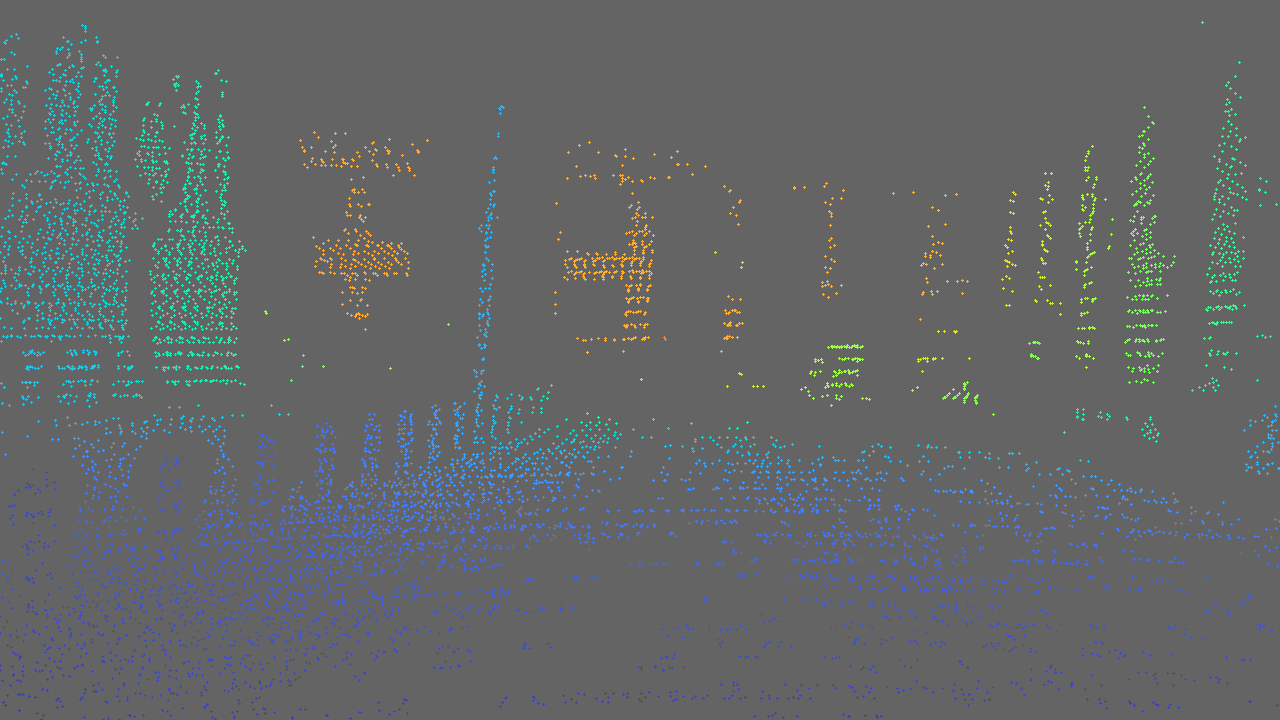}\\
  \includegraphics[width=0.4325\linewidth]{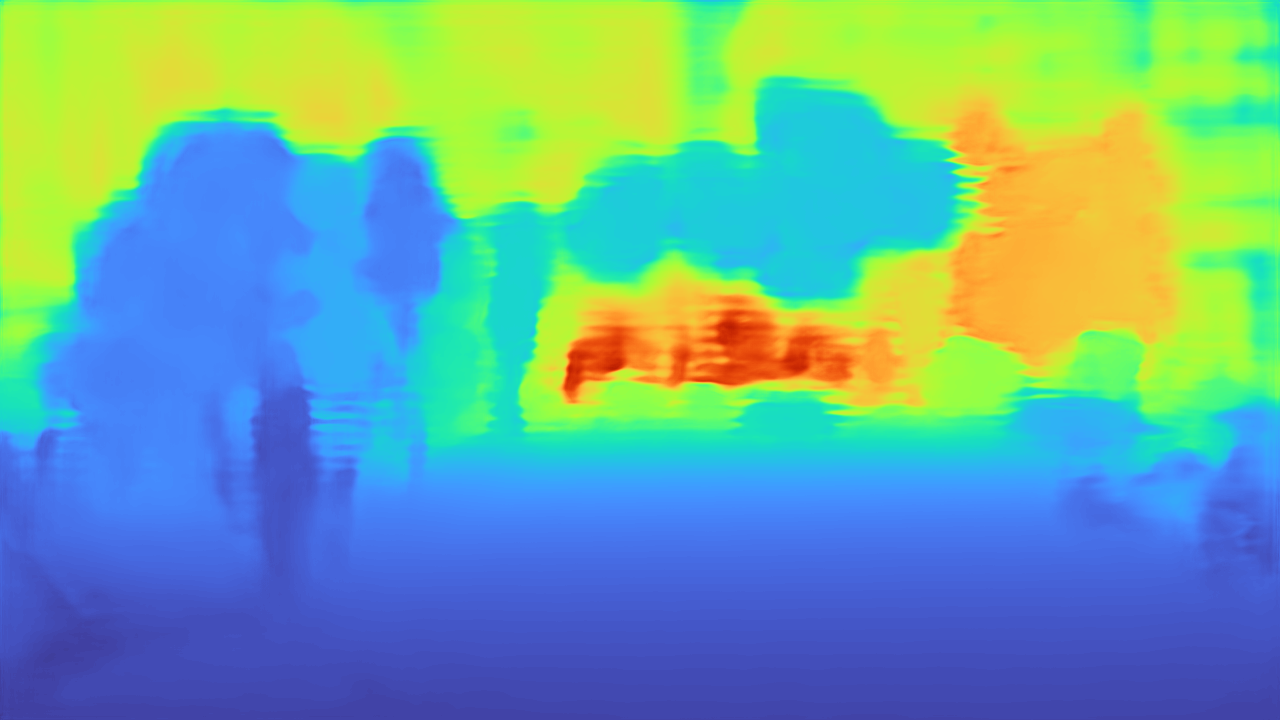}
  \includegraphics[width=0.4325\linewidth]{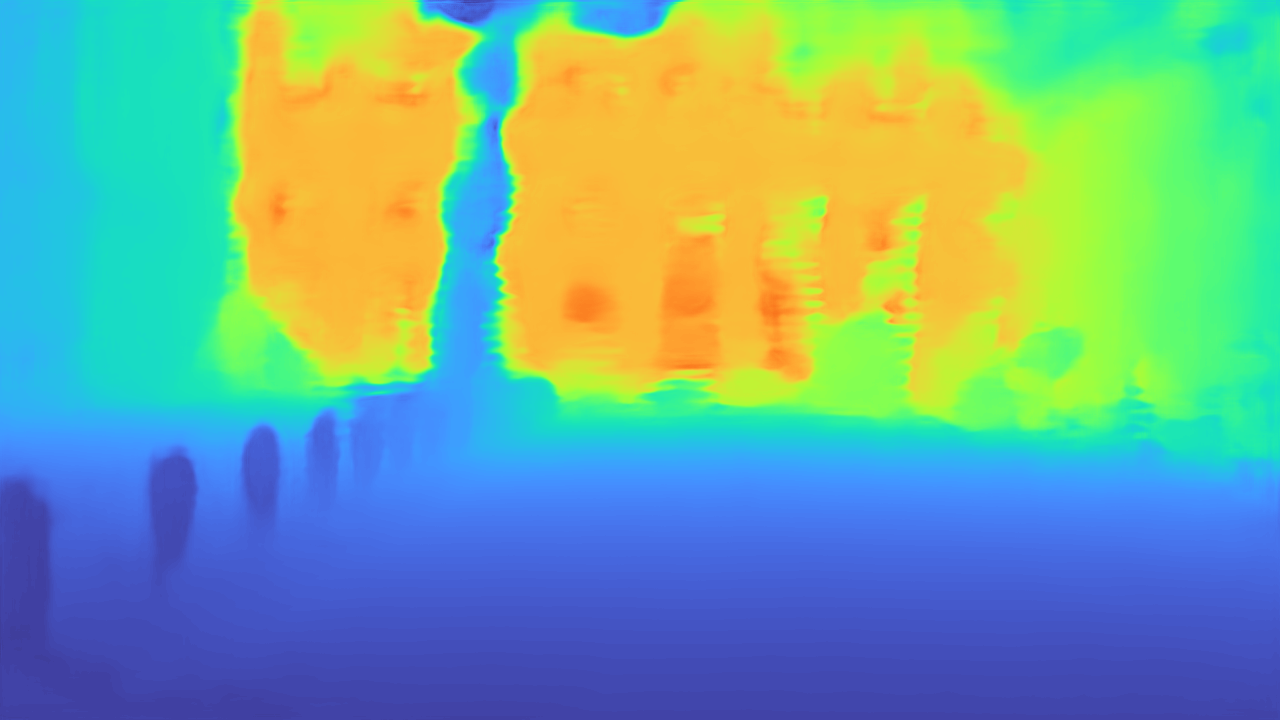}\\
  \includegraphics[width=0.4325\linewidth]{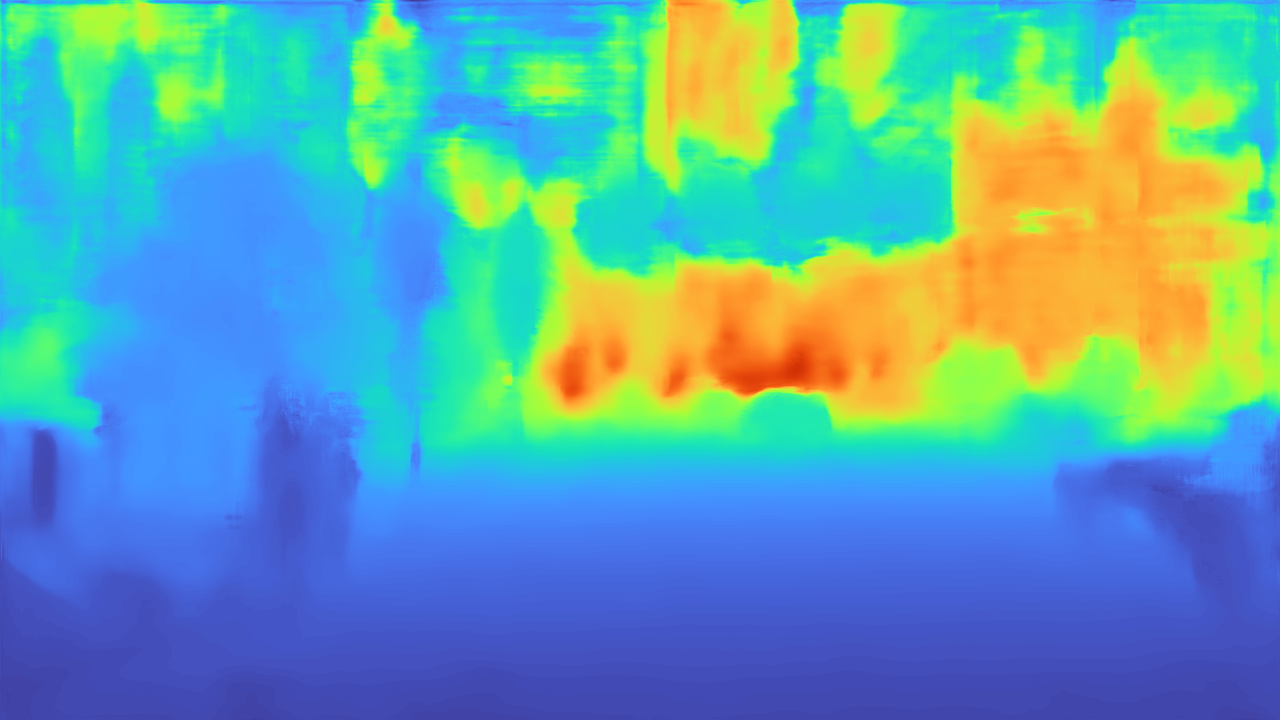}
  \includegraphics[width=0.4325\linewidth]{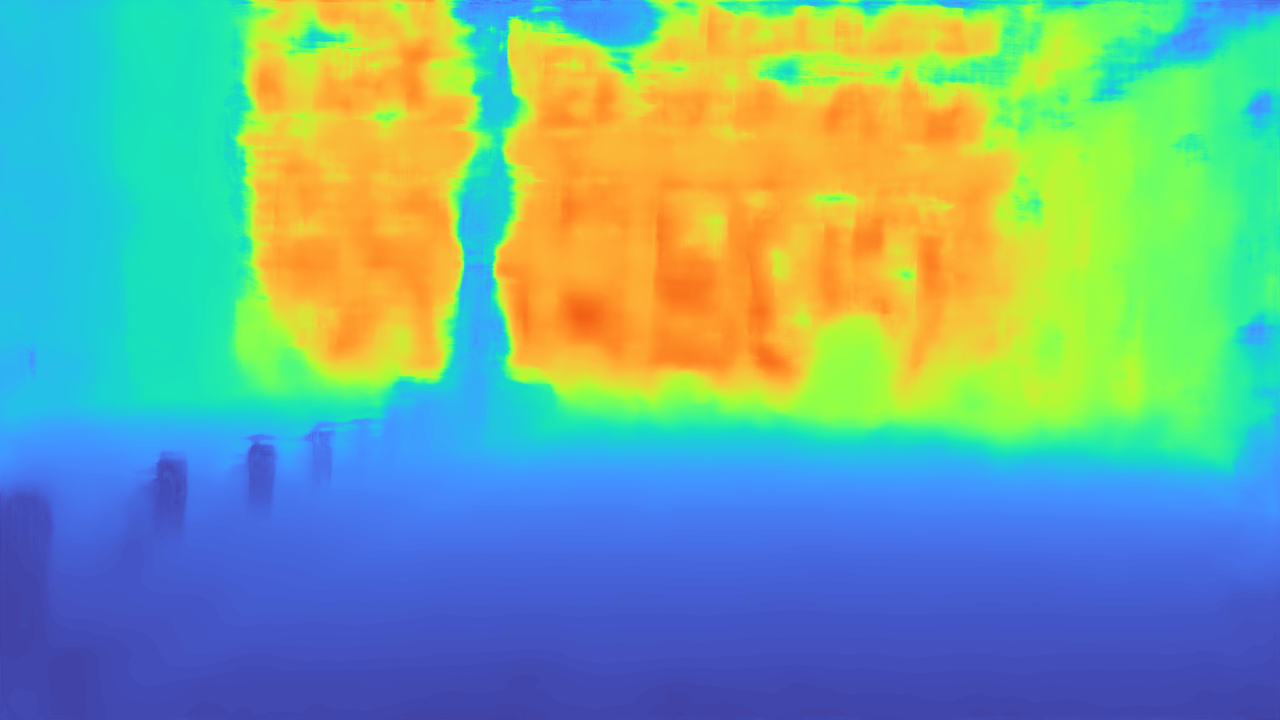}
  \cprotect\caption{Additional results on the M3ED dataset, for the \verb|city_hall_day| sequence. From top to bottom: events, LiDAR projection, ground truth, our results (DELTA\textsubscript{M3}, DELTA\textsubscript{SL\(\rightarrow\)M3}).}\label{fig:cmp_m3ed_additional_day}
\end{figure*}

\begin{figure*}
  \centering
  \includegraphics[width=0.4325\linewidth]{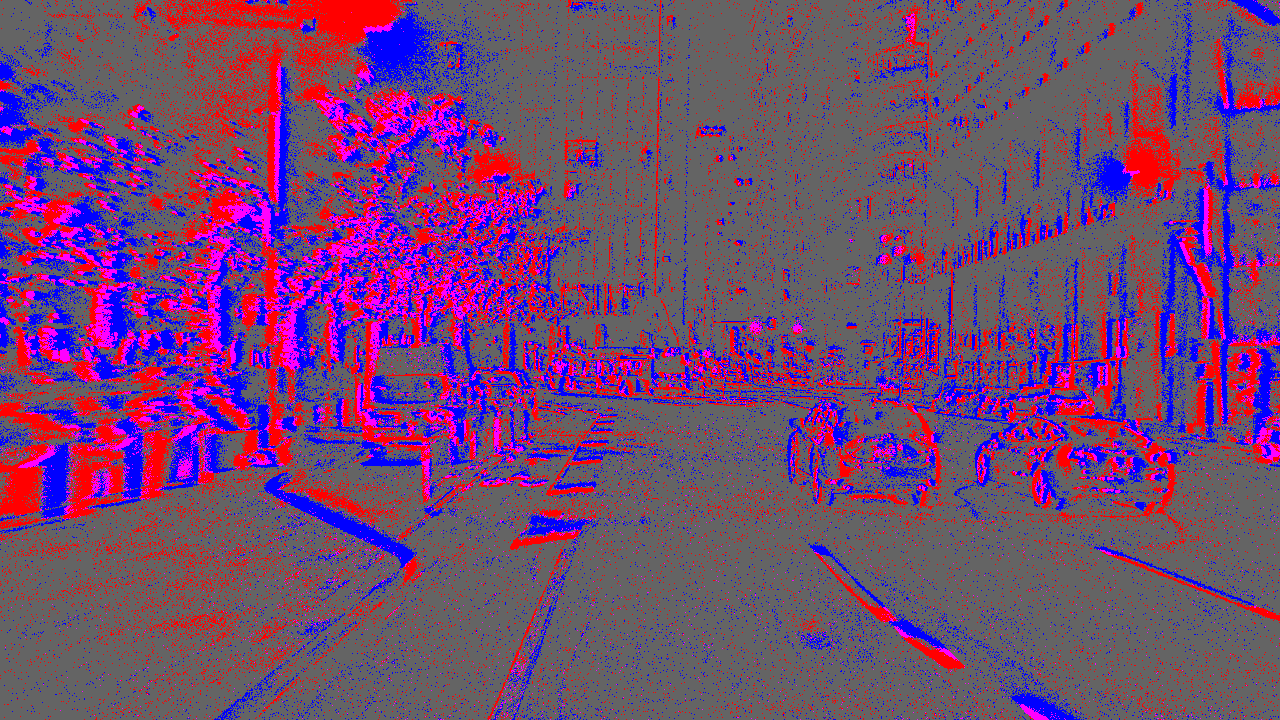}
  \includegraphics[width=0.4325\linewidth]{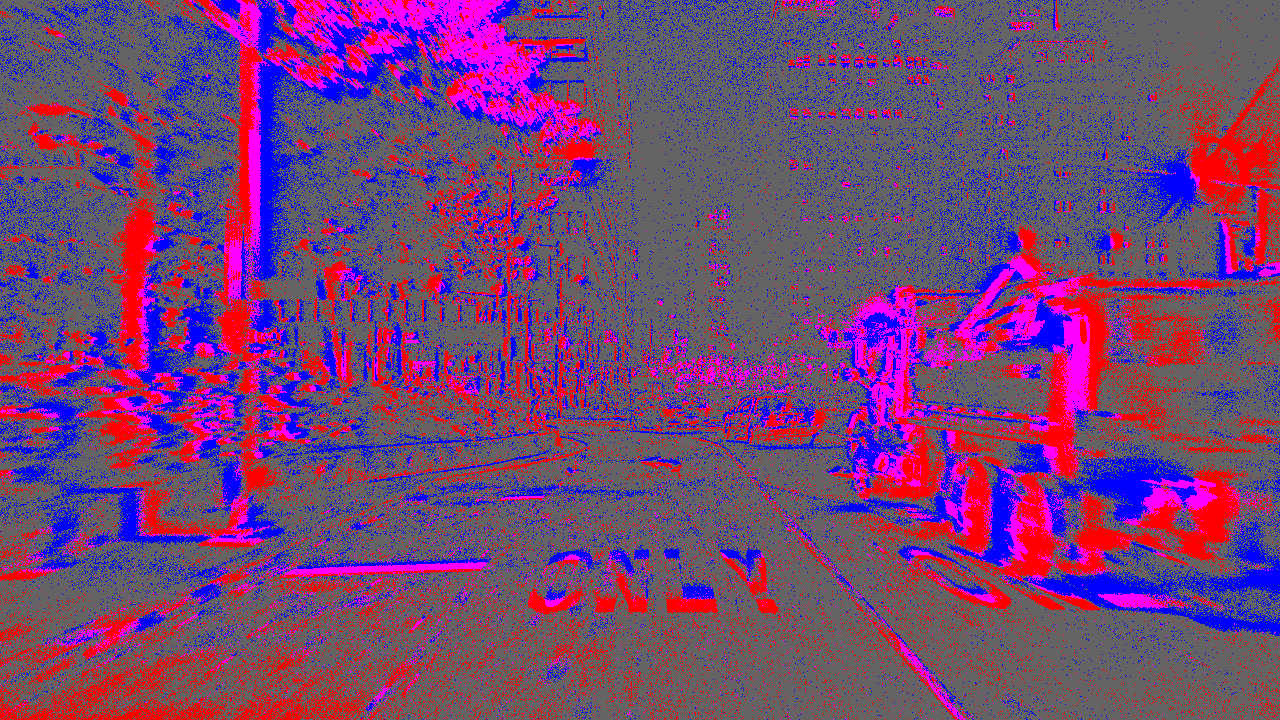}\\
  \includegraphics[width=0.4325\linewidth]{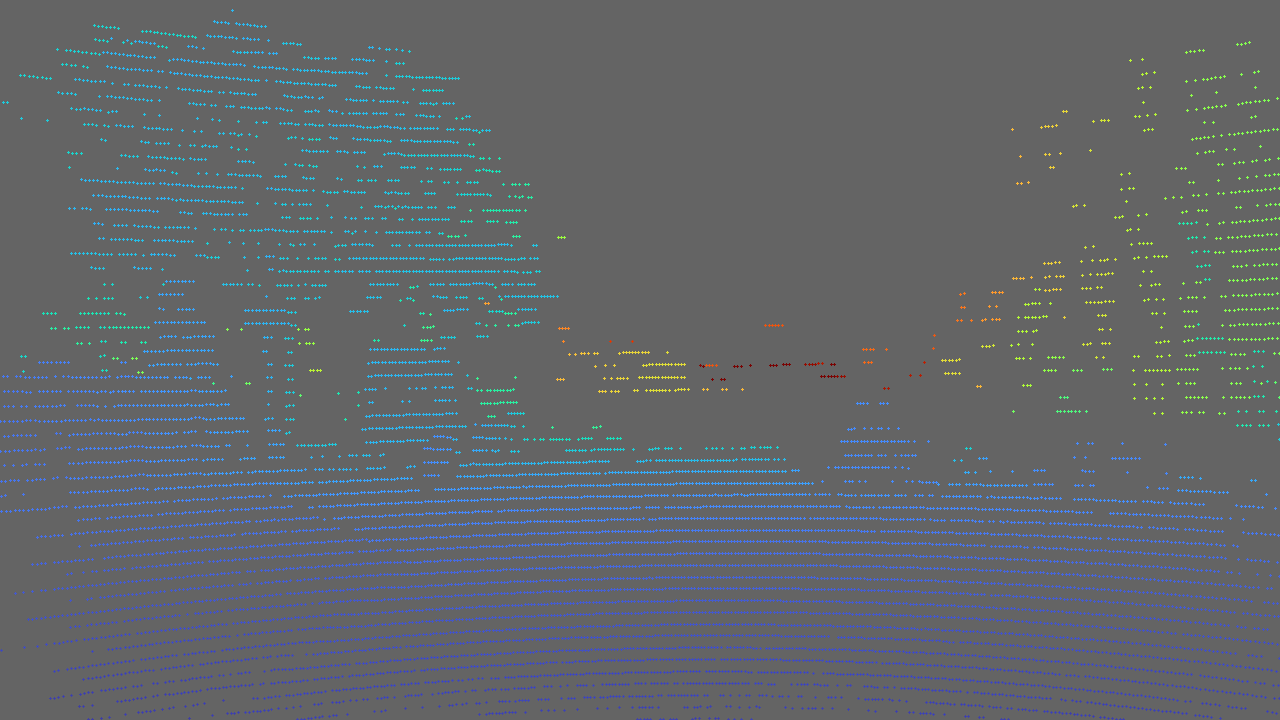}
  \includegraphics[width=0.4325\linewidth]{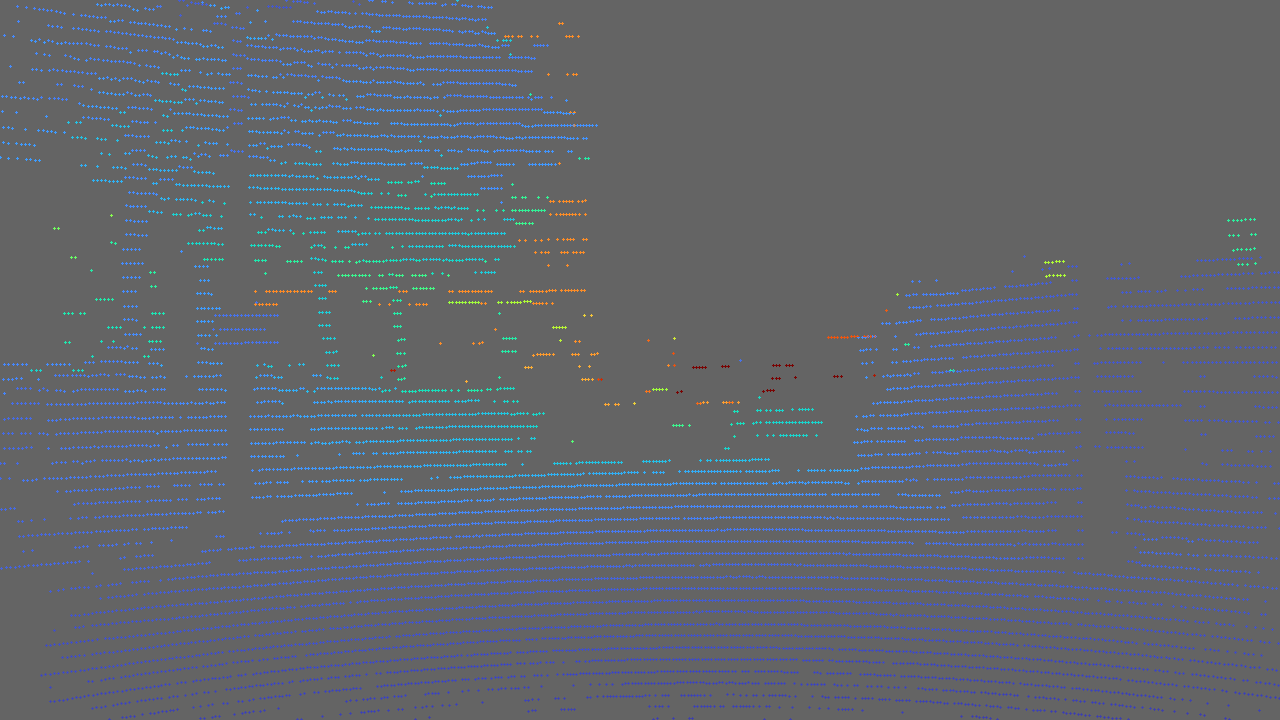}\\
  \includegraphics[width=0.4325\linewidth]{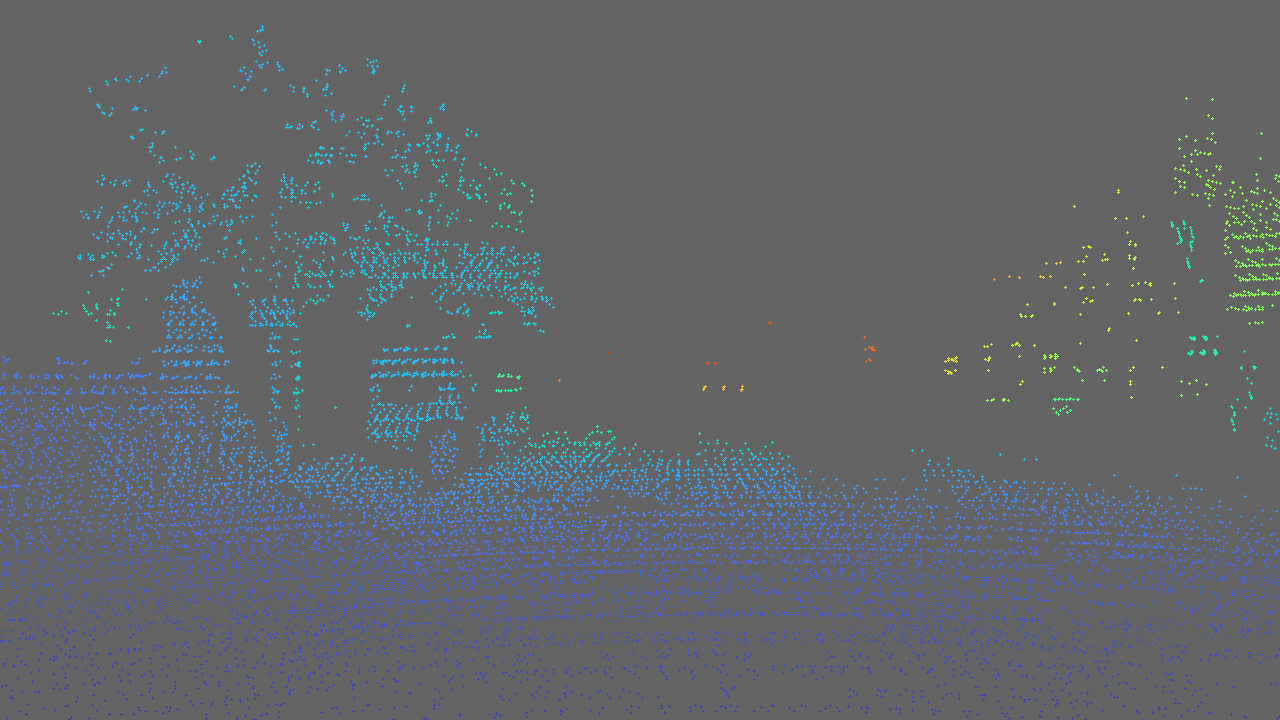}
  \includegraphics[width=0.4325\linewidth]{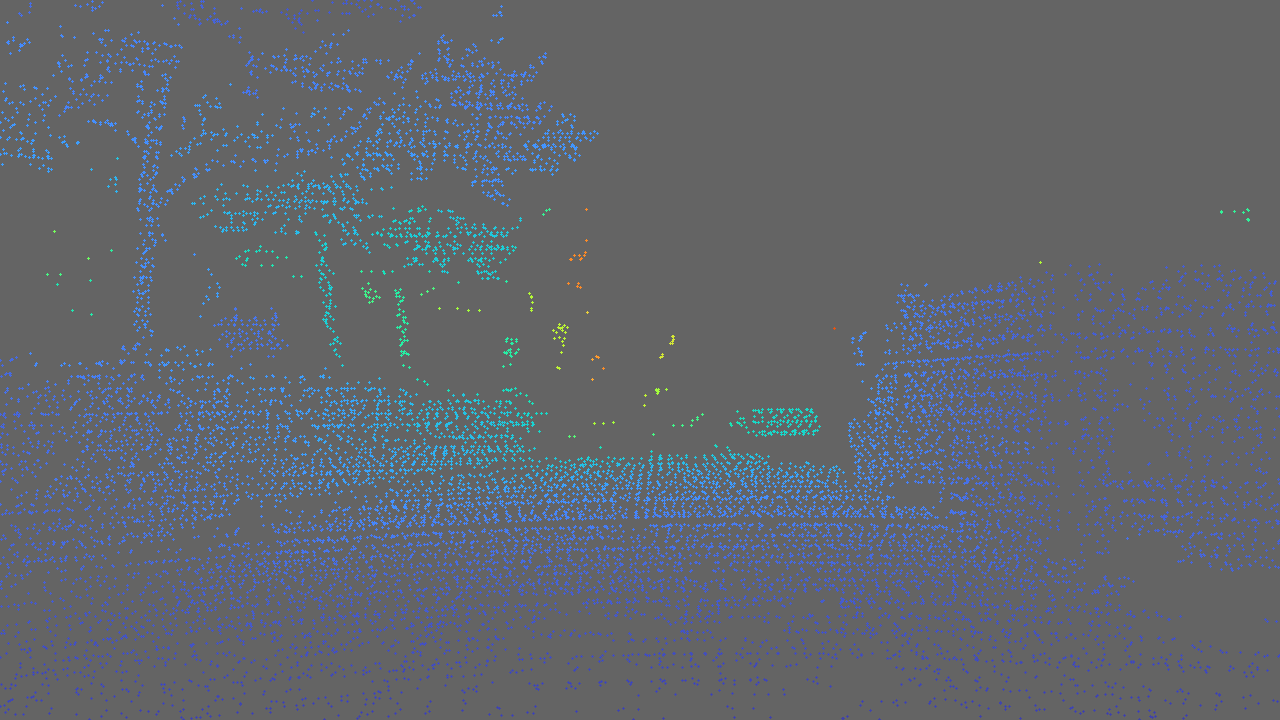}\\
  \includegraphics[width=0.4325\linewidth]{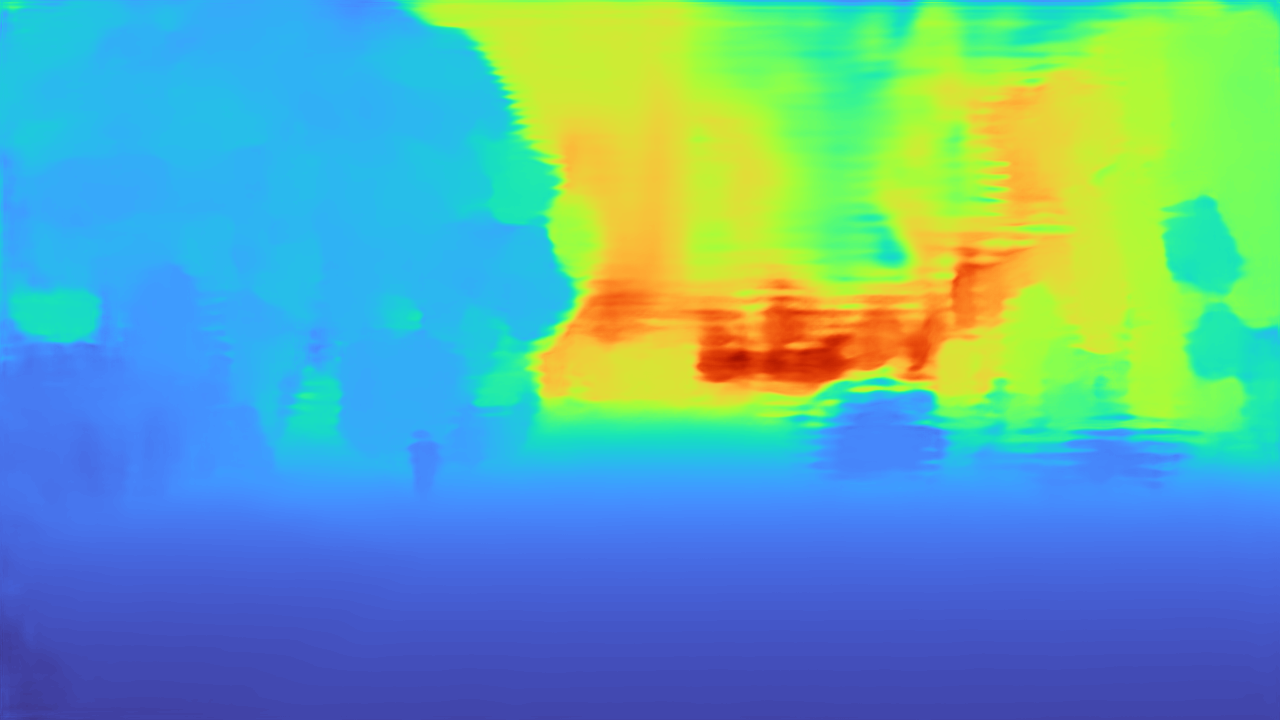}
  \includegraphics[width=0.4325\linewidth]{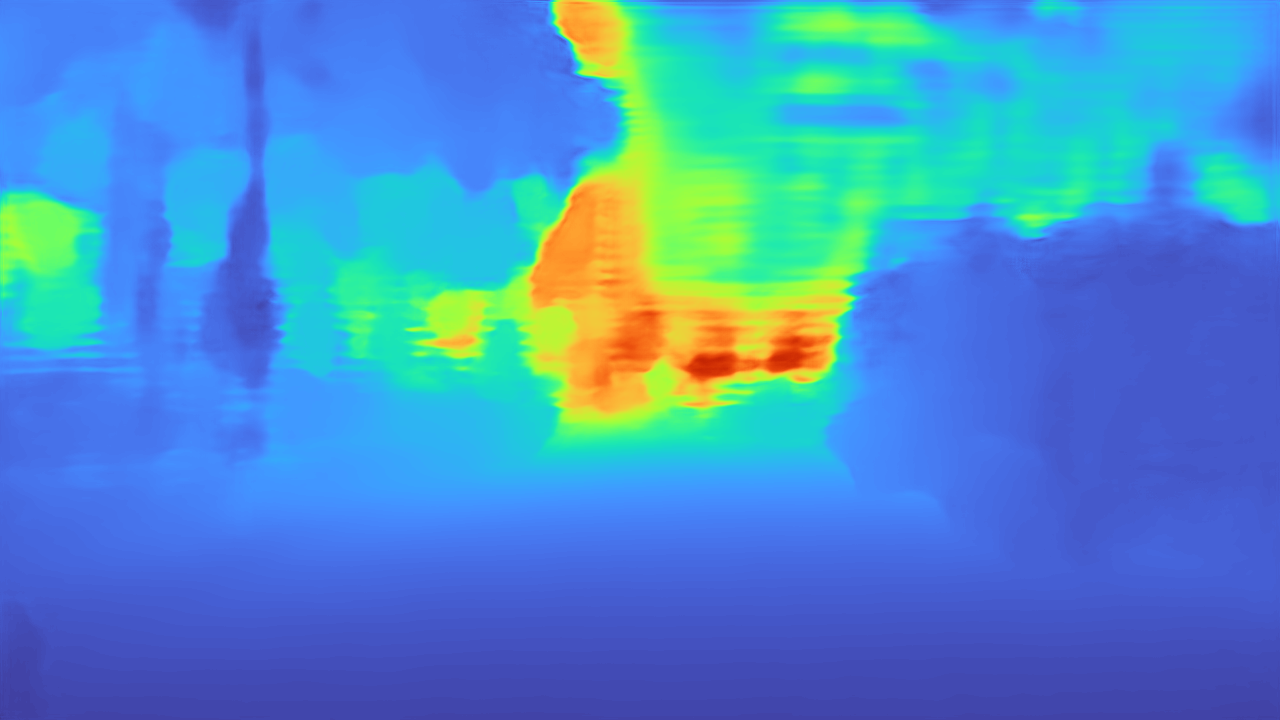}\\
  \includegraphics[width=0.4325\linewidth]{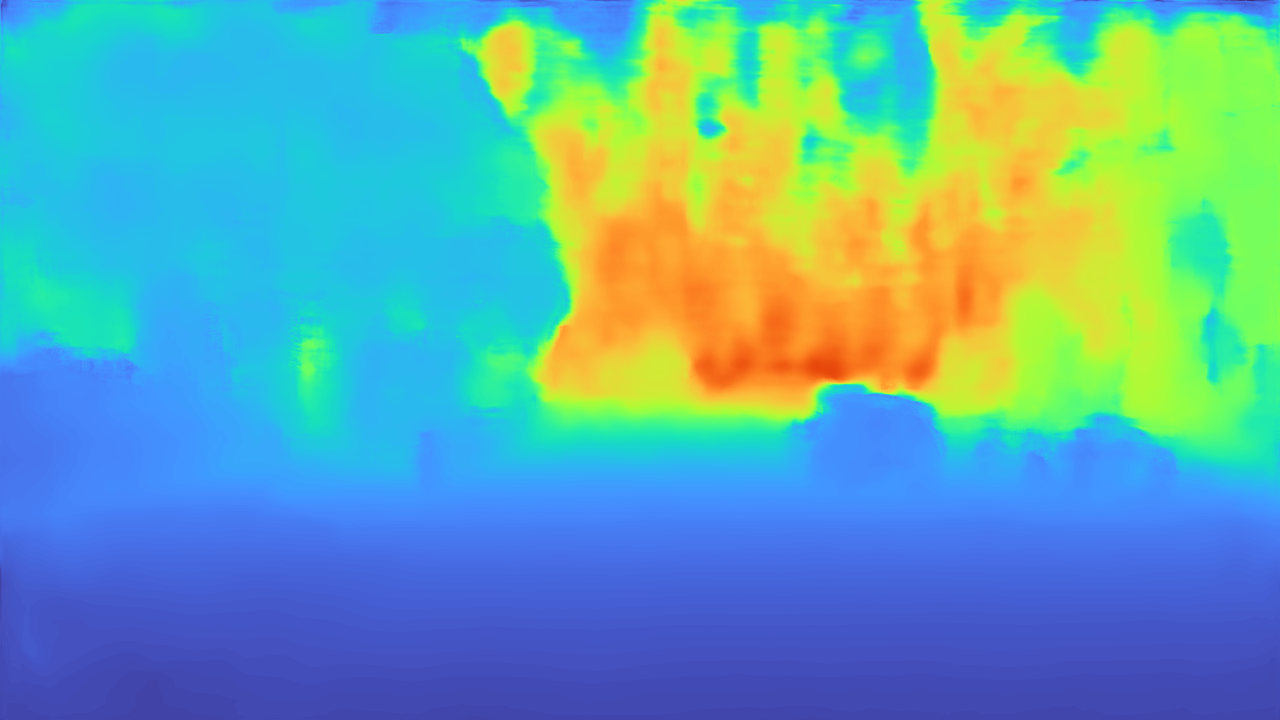}
  \includegraphics[width=0.4325\linewidth]{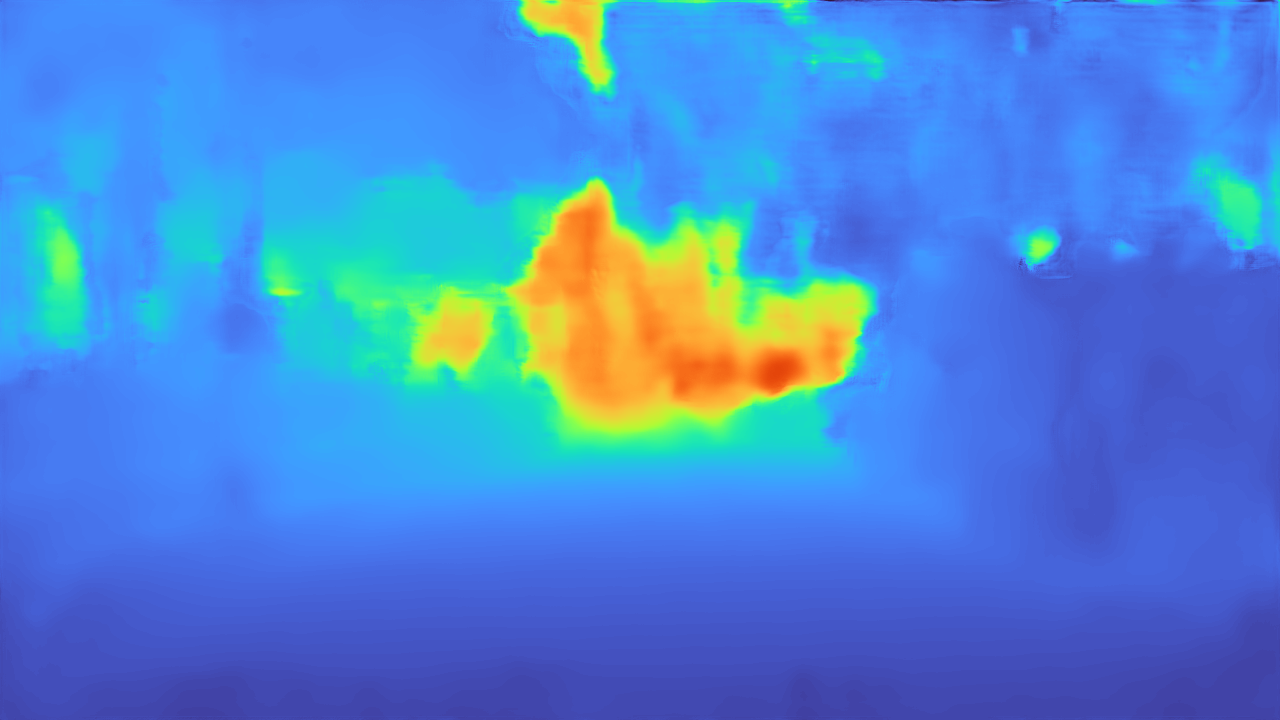}
  \cprotect\caption{Additional results on the M3ED dataset, for the \verb|city_hall_night| sequence. From top to bottom: events, LiDAR projection, ground truth, our results (DELTA\textsubscript{M3}, DELTA\textsubscript{SL\(\rightarrow\)M3}).}\label{fig:cmp_m3ed_additional_night}
\end{figure*}

Additional qualitative results on the M3ED dataset are given in \cref{fig:cmp_m3ed_additional_day,fig:cmp_m3ed_additional_night}, where the sparsity of the ground truth depth maps (especially compared to the density of the LiDAR data) and the blob-like appearance of the objects in the predictions can be better observed.

\begin{figure*}
  \centering
  \setlength\tabcolsep{1pt}
  \renewcommand{\arraystretch}{0.5}
\begin{tabular}{@{}cccccc@{}}
    \hphantom{Bonjourbonjourbon} & \hphantom{Bonjourbonjourbon} & \hphantom{Bonjourbonjourbon} & \hphantom{Bonjourbonjourbon} & \hphantom{Bonjourbonjourbon} & \hphantom{Bonjourbonjourbon} \\
    \multicolumn{3}{c}{\footnotesize LiDAR projection} & \multicolumn{3}{c}{\footnotesize Event volume} \\
    \multicolumn{3}{c}{\includegraphics[width=0.32\linewidth]{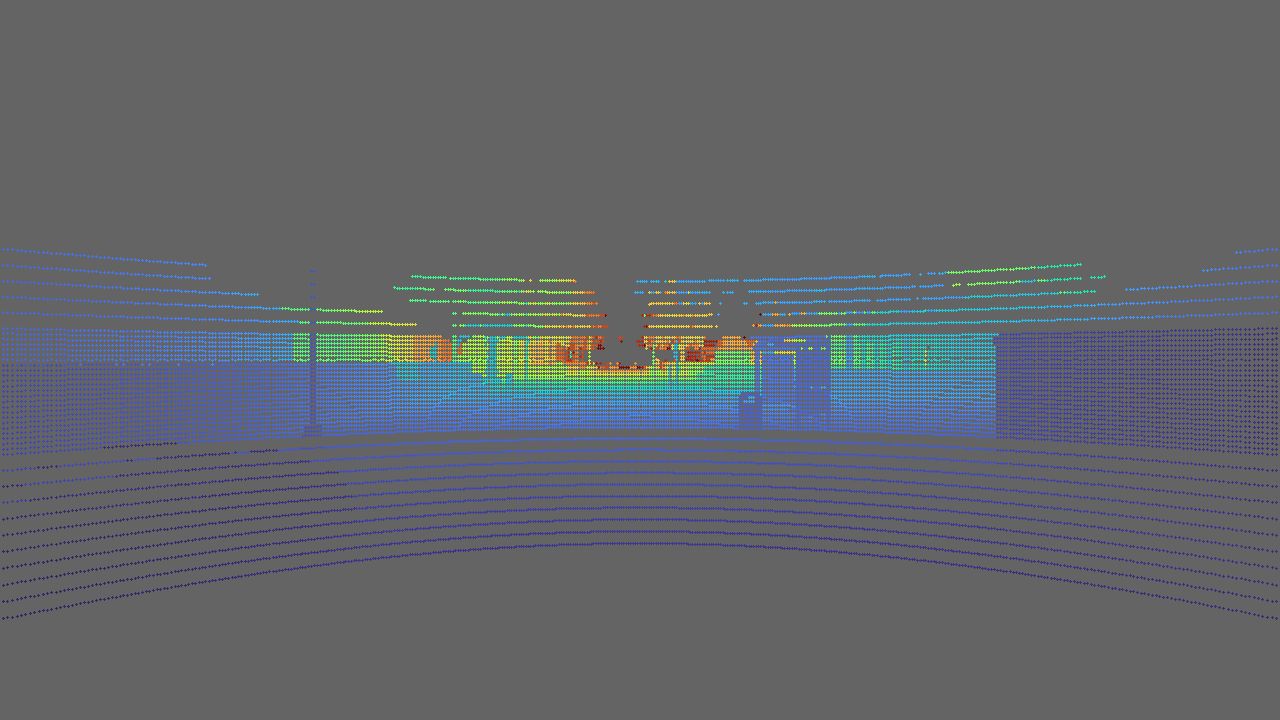}} &
    \multicolumn{3}{c}{\includegraphics[width=0.32\linewidth]{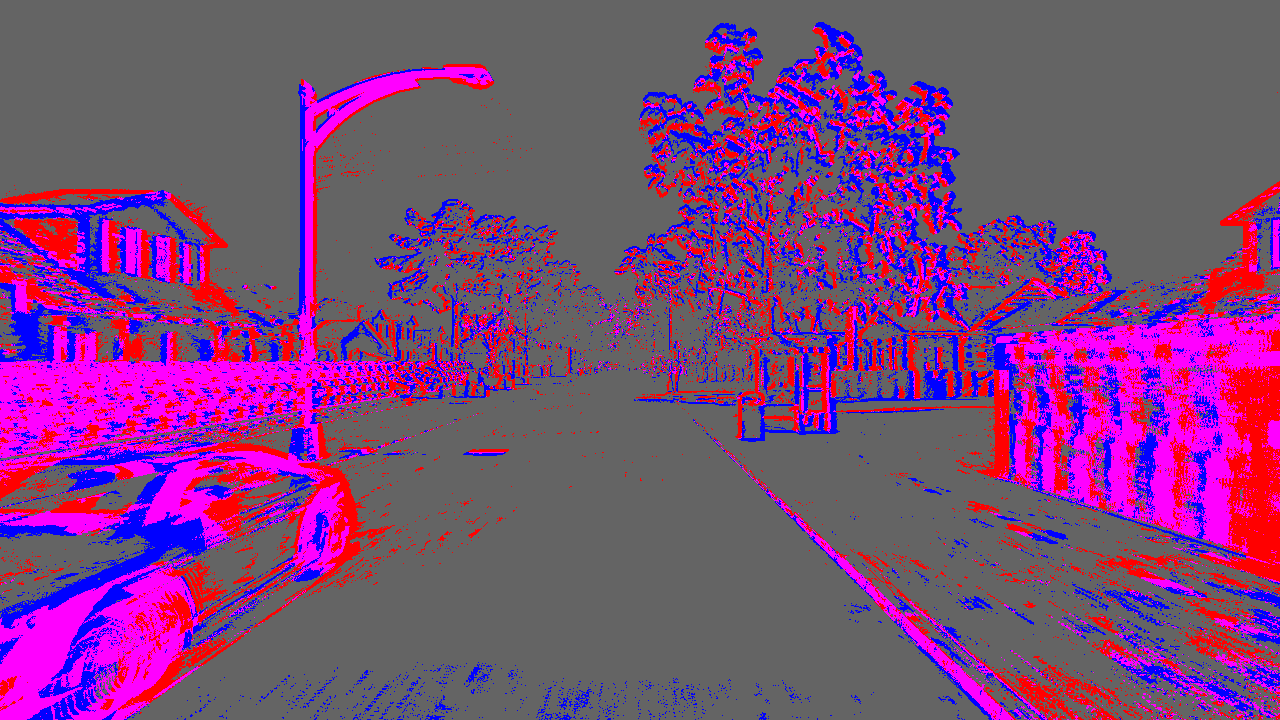}} \\
    \multicolumn{2}{c}{\footnotesize Ground truth} & \multicolumn{2}{c}{\footnotesize ALED\textsubscript{SL}~\cite{Brebion2023LearningTE} result} & \multicolumn{2}{c}{\footnotesize DELTA\textsubscript{SL} result} \\
    \multicolumn{2}{c}{\includegraphics[width=0.32\linewidth]{figures/sled_cmp/data_and_gt/gtbf_001662.png}} &
    \multicolumn{2}{c}{\includegraphics[width=0.32\linewidth]{figures/sled_cmp/aled/predbf_001662.png}} &
    \multicolumn{2}{c}{\includegraphics[width=0.32\linewidth]{figures/sled_cmp/delta/predbf_001662.png}} \\
    & & \multicolumn{2}{c}{\footnotesize ALED\textsubscript{SL}~\cite{Brebion2023LearningTE} error map} & \multicolumn{2}{c}{\footnotesize DELTA\textsubscript{SL} error map} \\
    & &
    \multicolumn{2}{c}{\includegraphics[width=0.32\linewidth]{figures/sled_cmp/aled/errorbf_001662.png}} &
    \multicolumn{2}{c}{\includegraphics[width=0.32\linewidth]{figures/sled_cmp/delta/errorbf_001662.png}} \\
  \end{tabular}

  \vspace{2.99mm}
  \hrule
  \vspace{2.99mm}

  \begin{tabular}{@{}cccccc@{}}
    \hphantom{Bonjourbonjourbon} & \hphantom{Bonjourbonjourbon} & \hphantom{Bonjourbonjourbon} & \hphantom{Bonjourbonjourbon} & \hphantom{Bonjourbonjourbon} & \hphantom{Bonjourbonjourbon} \\
    \multicolumn{3}{c}{\footnotesize LiDAR projection} & \multicolumn{3}{c}{\footnotesize Event volume} \\
    \multicolumn{3}{c}{\includegraphics[width=0.32\linewidth]{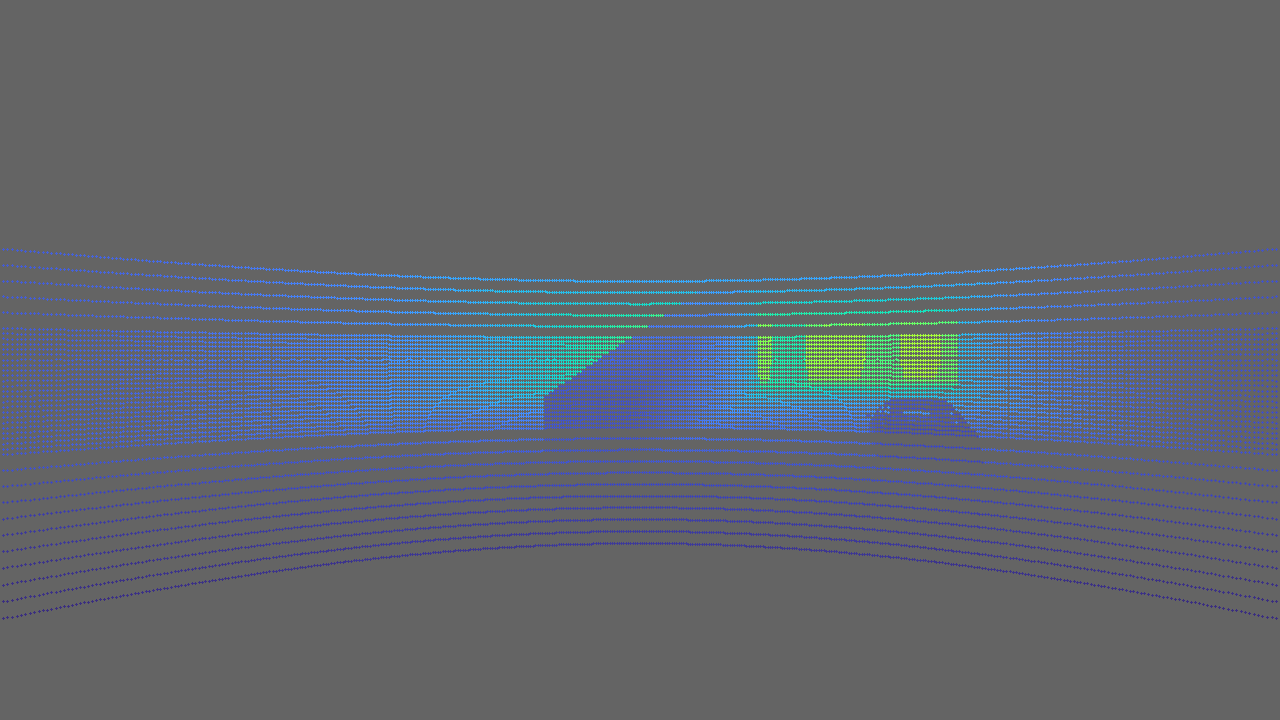}} &
    \multicolumn{3}{c}{\includegraphics[width=0.32\linewidth]{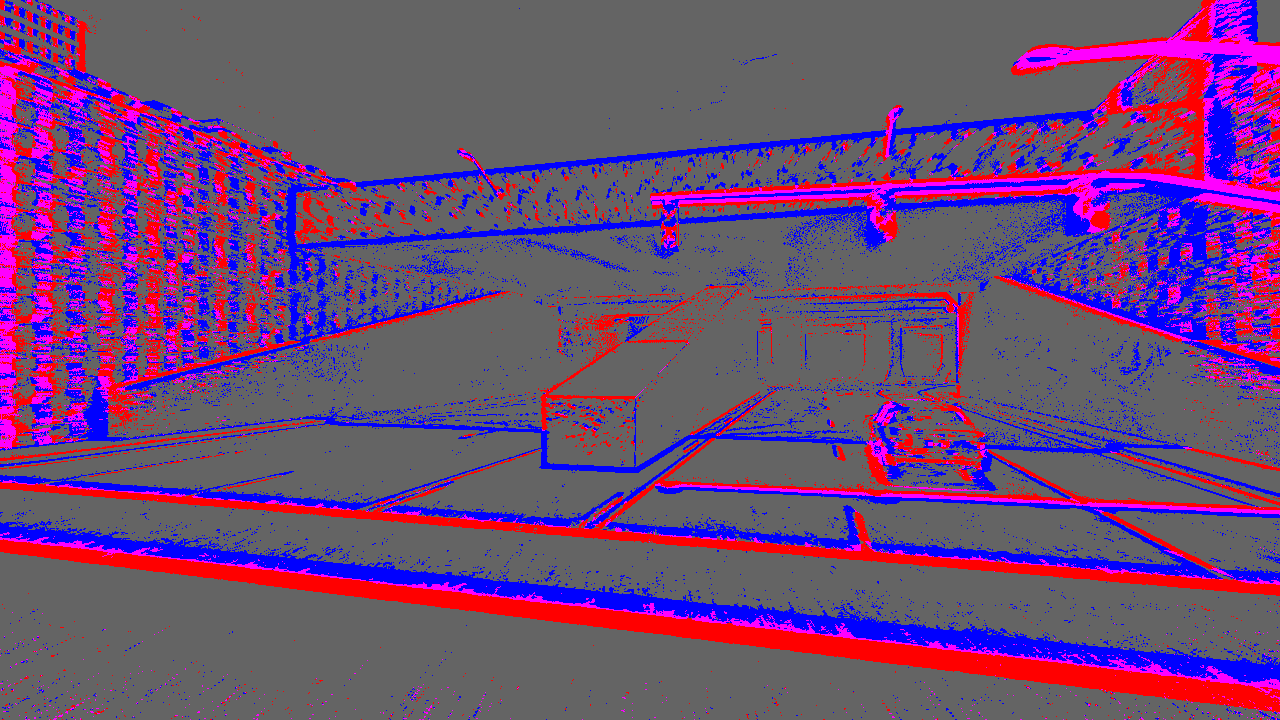}} \\
    \multicolumn{2}{c}{\footnotesize Ground truth} & \multicolumn{2}{c}{\footnotesize ALED\textsubscript{SL}~\cite{Brebion2023LearningTE} result} & \multicolumn{2}{c}{\footnotesize DELTA\textsubscript{SL} result} \\
    \multicolumn{2}{c}{\includegraphics[width=0.32\linewidth]{figures/sled_cmp/data_and_gt/gtbf_007812.png}} &
    \multicolumn{2}{c}{\includegraphics[width=0.32\linewidth]{figures/sled_cmp/aled/predbf_007812.png}} &
    \multicolumn{2}{c}{\includegraphics[width=0.32\linewidth]{figures/sled_cmp/delta/predbf_007812.png}} \\
    & & \multicolumn{2}{c}{\footnotesize ALED\textsubscript{SL}~\cite{Brebion2023LearningTE} error map} & \multicolumn{2}{c}{\footnotesize DELTA\textsubscript{SL} error map} \\
    & &
    \multicolumn{2}{c}{\includegraphics[width=0.32\linewidth]{figures/sled_cmp/aled/errorbf_007812.png}} &
    \multicolumn{2}{c}{\includegraphics[width=0.32\linewidth]{figures/sled_cmp/delta/errorbf_007812.png}} \\
  \end{tabular}
  \cprotect\caption{Comparison on the \verb|Town01_08| (top) and \verb|Town03_19| (bottom) sequences of SLED (enlarged version of \cref{fig:cmp_sled}).}\label{fig:cmp_sled_enlarged}
\end{figure*}

\begin{figure*}
  \centering
  \setlength\tabcolsep{1pt}
  \renewcommand{\arraystretch}{0.5}
  \begin{tabular}{@{}cccc@{}}
    \raisebox{1.4cm}[0pt][0pt]{\rotatebox[origin=c]{90}{Events}} &
    \includegraphics[width=0.2275\linewidth]{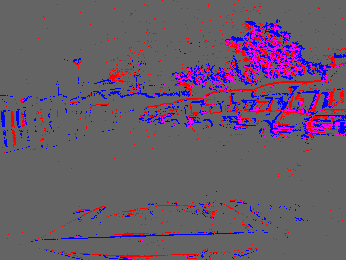} &
    \includegraphics[width=0.2275\linewidth]{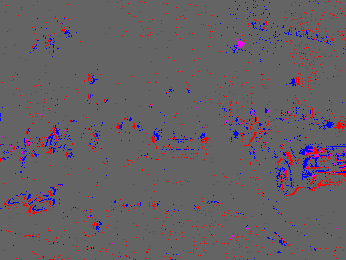} &
    \includegraphics[width=0.2275\linewidth]{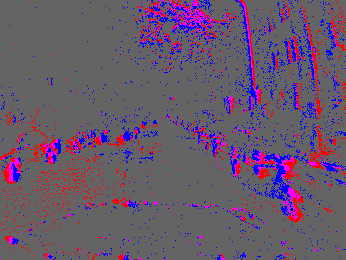} \\
    \raisebox{1.4cm}[0pt][0pt]{\rotatebox[origin=c]{90}{LiDAR proj.}} &
    \includegraphics[width=0.2275\linewidth]{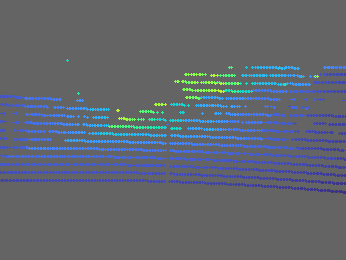} &
    \includegraphics[width=0.2275\linewidth]{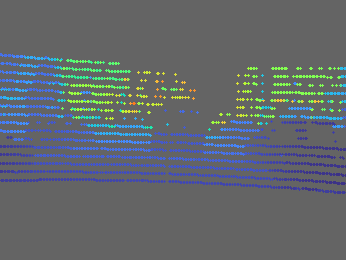} &
    \includegraphics[width=0.2275\linewidth]{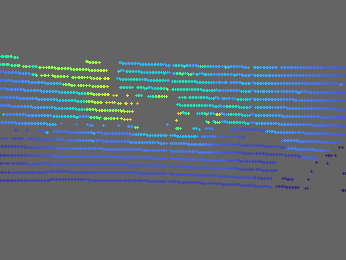} \\
    \raisebox{1.4cm}[0pt][0pt]{\rotatebox[origin=c]{90}{Ground truth}} &
    \includegraphics[width=0.2275\linewidth]{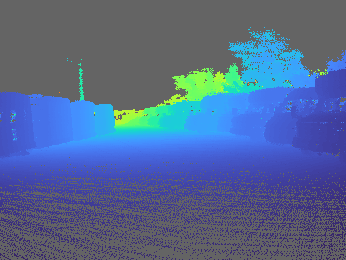} &
    \includegraphics[width=0.2275\linewidth]{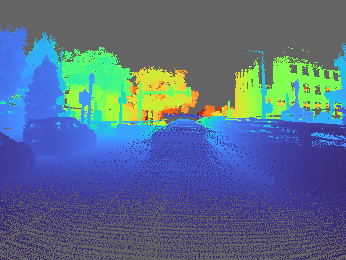} &
    \includegraphics[width=0.2275\linewidth]{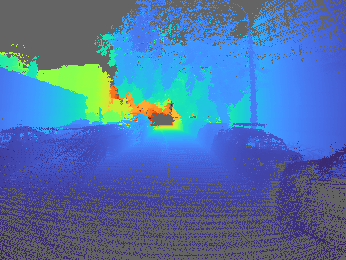} \\
    \raisebox{1.4cm}[0pt][0pt]{\rotatebox[origin=c]{90}{Cui \etal~\cite{Cui2022DenseDE}}} &
    \includegraphics[width=0.2275\linewidth]{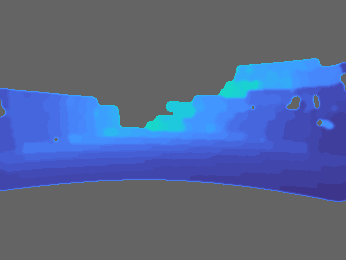} &
    \includegraphics[width=0.2275\linewidth]{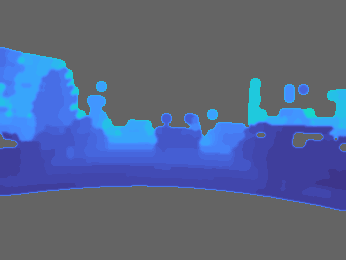} &
    \includegraphics[width=0.2275\linewidth]{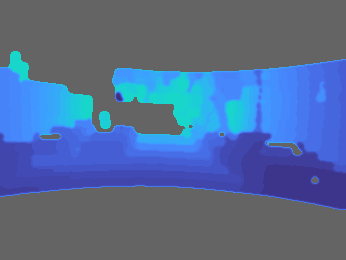} \\
    \raisebox{1.4cm}[0pt][0pt]{\rotatebox[origin=c]{90}{ALED\textsubscript{SL\(\rightarrow\)MV}~\cite{Brebion2023LearningTE}}} &
    \includegraphics[width=0.2275\linewidth]{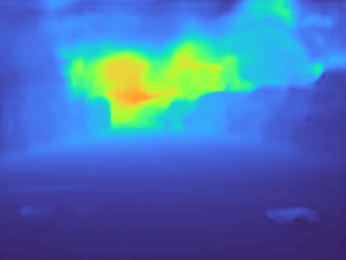} &
    \includegraphics[width=0.2275\linewidth]{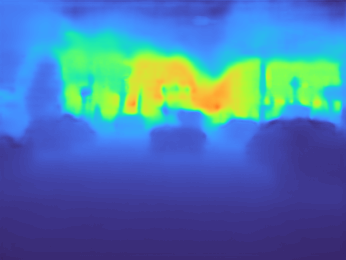} &
    \includegraphics[width=0.2275\linewidth]{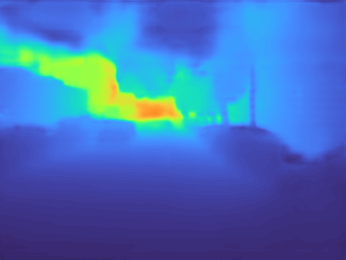} \\
\end{tabular}
  \cprotect\caption{Comparison on the \verb|outdoor_day_1|, \verb|outdoor_night_1|, and \verb|outdoor_night_2| sequences of MVSEC (enlarged version of \cref{fig:cmp_mvsec}).}\label{fig:cmp_mvsec_enlarged}
\end{figure*}

\begin{figure*}
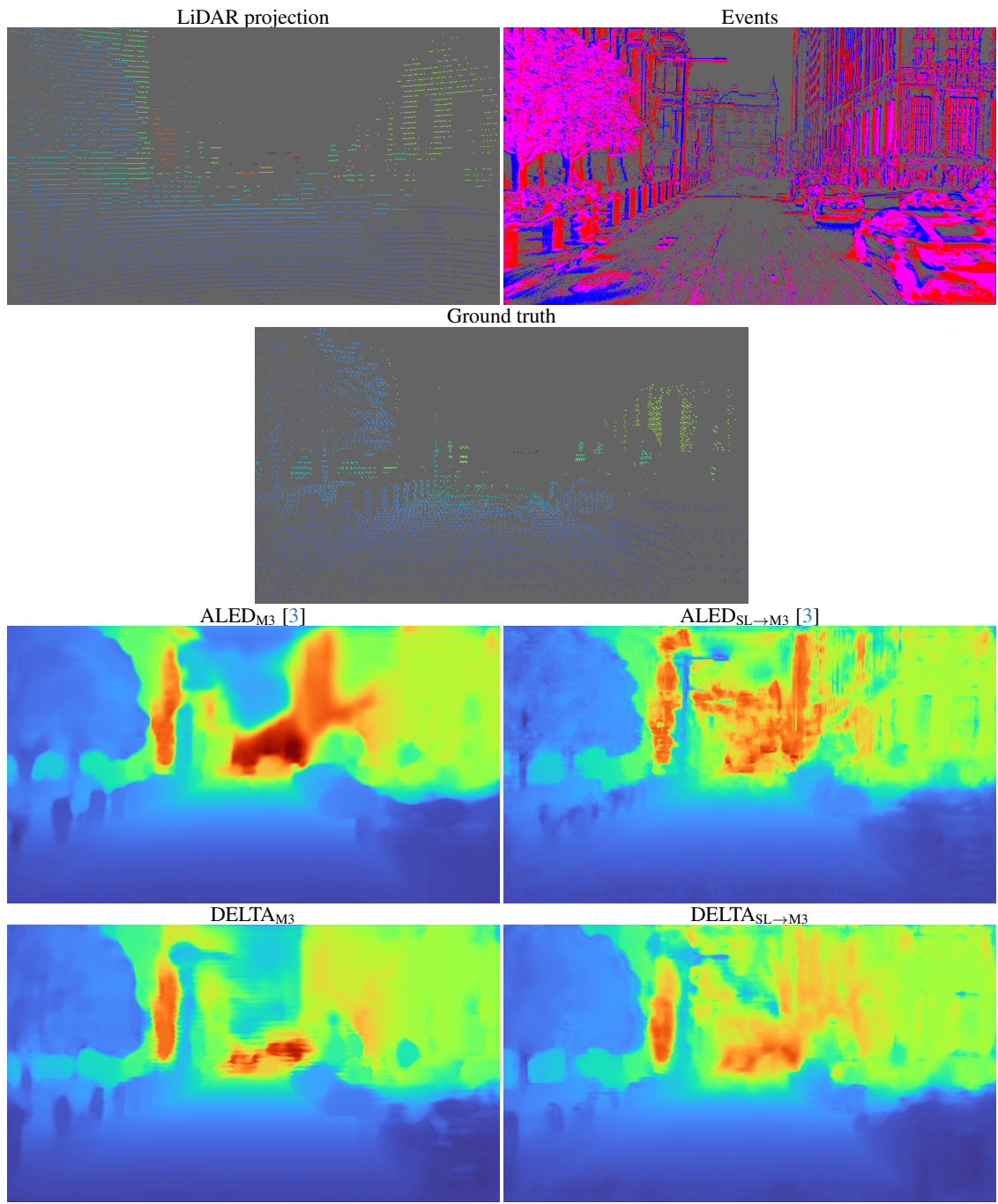

  \centering
  \setlength\tabcolsep{1pt}
  \renewcommand{\arraystretch}{0.5}
  \begin{tabular}{@{}cc@{}}
    \small LiDAR projection & \small Events \\
    \includegraphics[width=0.425\linewidth]{figures/m3ed_cmp/data_and_gt/lidar_001036_biggerpoints.png} &
    \includegraphics[width=0.425\linewidth]{figures/m3ed_cmp/data_and_gt/evts_001036.png} \\
    \multicolumn{2}{c}{\small Ground truth} \\
    \multicolumn{2}{c}{\includegraphics[width=0.425\linewidth]{figures/m3ed_cmp/data_and_gt/gtbf_001036_biggerpoints.png}} \\
    \small ALED\textsubscript{M3}~\cite{Brebion2023LearningTE} & \small ALED\textsubscript{SL\(\rightarrow\)M3}~\cite{Brebion2023LearningTE} \\
    \includegraphics[width=0.425\linewidth]{figures/m3ed_cmp/aled_m3/predbf_001036.png} &
    \includegraphics[width=0.425\linewidth]{figures/m3ed_cmp/aled_sl_m3/predbf_001036.png} \\
    \small DELTA\textsubscript{M3} & \small DELTA\textsubscript{SL\(\rightarrow\)M3} \\
    \includegraphics[width=0.425\linewidth]{figures/m3ed_cmp/delta_m3/predbf_001036.png} &
    \includegraphics[width=0.425\linewidth]{figures/m3ed_cmp/delta_sl_m3/predbf_001036.png}
  \end{tabular}
  \cprotect\caption{Comparison on the \verb|city_hall_day| sequence of M3ED (enlarged version of \cref{fig:cmp_m3ed}).}\label{fig:cmp_m3ed_enlarged}
\end{figure*}

\begin{figure*}
  \centering
  \resizebox{0.8\linewidth}{!}{
    %
  }

  \caption{The alternative architecture without the LiDAR input, DELTA\textsuperscript{NL}.}\label{fig:network_alt_nl}
\end{figure*}

\begin{figure*}
  \centering
  \resizebox{0.8\linewidth}{!}{
    %
  }

  \caption{The alternative architecture without the event input, DELTA\textsuperscript{NE}.}\label{fig:network_alt_ne}
\end{figure*}

\begin{figure*}
  \centering
  \setlength\tabcolsep{1pt}
  \renewcommand{\arraystretch}{0.5}
  \begin{tabular}{@{}ccc@{}}
    Ground truth & DELTA\textsubscript{SL} result & DELTA\rlap{\textsubscript{SL}}\textsuperscript{NEH} result \\
    \includegraphics[width=0.33\linewidth]{figures/sled_cmp/data_and_gt/gtbf_001662.png} &
    \includegraphics[width=0.33\linewidth]{figures/sled_cmp/delta/predbf_001662.png} &
    \includegraphics[width=0.33\linewidth]{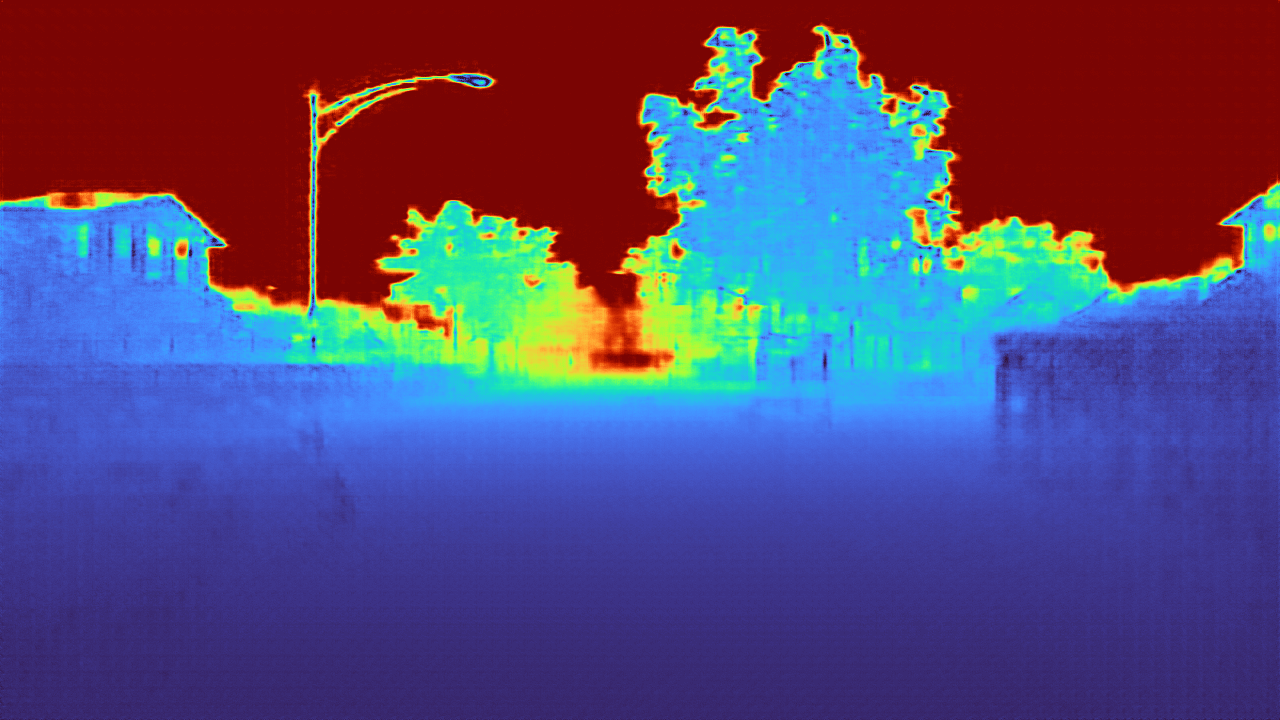} \\
    \includegraphics[width=0.33\linewidth]{figures/sled_cmp/data_and_gt/gtbf_007812.png} &
    \includegraphics[width=0.33\linewidth]{figures/sled_cmp/delta/predbf_007812.png} &
    \includegraphics[width=0.33\linewidth]{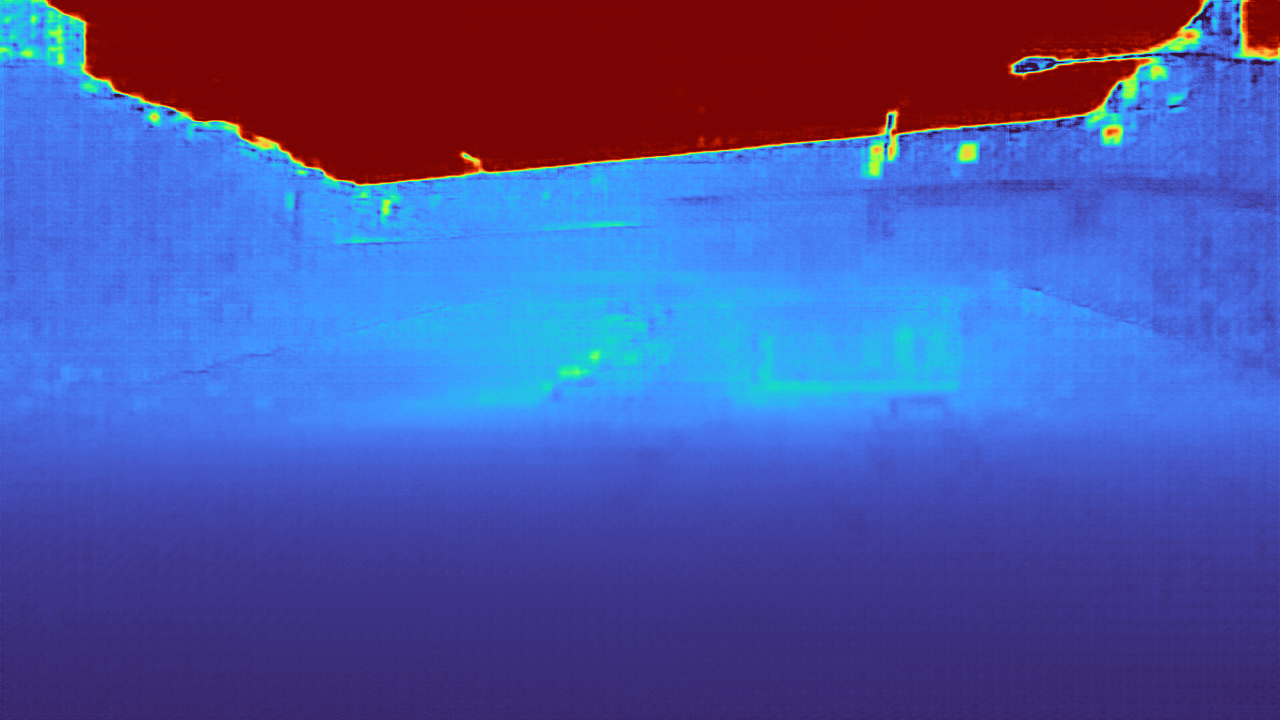} \\
  \end{tabular}
  \cprotect\caption{Results on the \verb|Town01_08| (top) and \verb|Town03_19| (bottom) sequences of SLED, for DELTA\textsubscript{SL} and DELTA\rlap{\textsubscript{SL}}\textsuperscript{NEH}. Zoom on the numerical version may be required to better see the individual patches and artifacts for DELTA\rlap{\textsubscript{SL}}\textsuperscript{NEH}.}\label{fig:cmp_neh}
\end{figure*}

\end{document}